%% file: main.tex
\documentclass[letterpaper,journal]{IEEEtran}
\usepackage[T1]{fontenc}
\usepackage{newtxtext}
\usepackage{amsmath,amssymb,amsfonts}
\usepackage{graphicx,booktabs,array,tabularx,multirow}
\usepackage{ragged2e}
\usepackage[table]{xcolor}
\usepackage{cite,url}
\usepackage{tikz}
\usetikzlibrary{calc,backgrounds}
\usepackage[hidelinks]{hyperref}
\graphicspath{{figures/}}
\newcommand{\R}{\mathbb{R}}
\newcommand{\supp}{\mathcal{S}}

\newcolumntype{Y}{>{\RaggedRight\arraybackslash}X}
\newcolumntype{P}[1]{>{\RaggedRight\arraybackslash}p{#1}}
\begin{document}
\bstctlcite{IEEEexample:BSTcontrol}
\title{Sparse Attention for Visual Generation: Toward Algorithm--System Co-Design}
\author{%
  Majun Zhang\textsuperscript{1,*},
  Yuan Guo\textsuperscript{2,*},
  Decheng Wu\textsuperscript{3,*},
  Jiayang Xu\textsuperscript{1},
  Changhao Pan\textsuperscript{1},
  Fan Zhuo\textsuperscript{1},
  Xiaoda Yang\textsuperscript{1},
  Chenyuhao Wen\textsuperscript{1},
  and Zhou Zhao\textsuperscript{1,\ensuremath{\dagger}}%
  \thanks{\textsuperscript{1}Zhejiang University}%
  \thanks{\textsuperscript{2}University of California, Los Angeles.}%
  \thanks{\textsuperscript{3}The University of Hong Kong.}%
  \thanks{\textsuperscript{*}Majun Zhang, Yuan Guo, and Decheng Wu
  contributed equally to this work.}%
  \thanks{\textsuperscript{\ensuremath{\dagger}}Corresponding author:
  Zhou Zhao.}%
}
\markboth{Sparse Attention for Visual Generation}{Toward Algorithm--System Co-Design}
\maketitle
\begin{abstract}
Rapid advances in image and video generation have enabled high-fidelity synthesis at higher
resolutions and longer durations, making the quadratic cost of dense self-attention a major
bottleneck as visual token sequences grow.
Sparse attention seeks to preserve generation quality at lower computational cost by retaining
informative query--key interactions and omitting less consequential ones, with end-to-end gains
depending on the cost of constructing sparse support and the efficiency with which the system
maps the resulting workload to hardware.
We survey sparse attention for visual generation through a \(4 + 1 + 3\) analytical framework
that links algorithmic design to system implementation.
The framework examines how static and dynamic methods use source information and the evidence
derived from it to decide which interactions to retain and at what granularity, while accounting
for the roles of reuse in reducing repeated computation and correction in compensating for
omitted contributions.
We relate these algorithmic choices to their implementation on hardware through the framework's
three system dimensions of data organization, execution, and dataflow.
We summarize common evaluation metrics and discuss open challenges and research directions,
providing researchers and practitioners with a reference for the joint design of sparse
attention algorithms and systems.
\end{abstract}
\input{sections/01_introduction}
\input{sections/01_background_preliminaries}
\input{sections/02_framework}

\input{sections/03_unified_review}

\input{sections/05_evaluation}

\input{sections/06_challenges}

\input{sections/07_conclusion}
\bibliographystyle{IEEEtran}
\bibliography{references}
\end{document}

%% file: sections/01_introduction.tex
\section{Introduction}
\label{sec:introduction}

Visual generation has advanced rapidly, with recent models delivering increasingly capable image and video synthesis. Qwen-Image \cite{wu2025qwenimage} and BAGEL \cite{deng2025bagel} demonstrate strong text rendering, precise image editing, and unified multimodal generation, while HunyuanVideo \cite{kong2024hunyuanvideo}, Wan \cite{wanteam2025wan}, and Seedance \cite{gao2025seedance} achieve high visual fidelity, temporal coherence, and prompt adherence in video generation. These advances are expanding the value of visual generation across creative production, personalized media, and interactive simulation.

Despite these advances, computational efficiency remains a major obstacle to the practical use of visual generation. Wan2.1 \cite{wanteam2025wan} takes approximately 56 minutes on a single A100 GPU to generate a five-second 720p video. Higher resolutions and longer durations are more demanding still, as the expanding visual-token sequence causes the quadratic pairwise computation of dense attention to rise sharply across transformer layers and denoising steps. Autoregressive visual models face a related challenge when current tokens attend to an increasingly long generation history. Such latency restricts generation scale, slows iterative content creation, and hinders real-time interaction and widespread deployment.

These constraints have motivated several complementary approaches to efficient visual generation. Distillation shortens the sampling trajectory \cite{salimans2022progressive}, whereas cache reuse avoids recomputing similar intermediate features across denoising steps \cite{deepcache2023,teacache2024}. Quantization reduces numerical precision \cite{qdiffusion2023}, and pruning removes less influential parameters or layers \cite{diffpruning2023,tinyfusion2025}. These strategies reduce the number or cost of model evaluations, but do not directly select which query--key interactions attention evaluates. Sparse attention addresses this interaction structure by restricting computation to a compact support that retains dominant spatial, temporal, and semantic dependencies \cite{p190410509,p250201776,p250218137}. It is therefore particularly relevant to the long visual-token sequences encountered in high-resolution image synthesis, video generation, and autoregressive visual modeling.

\begin{figure*}[!t]
  \centering
  \includegraphics[width=\textwidth]{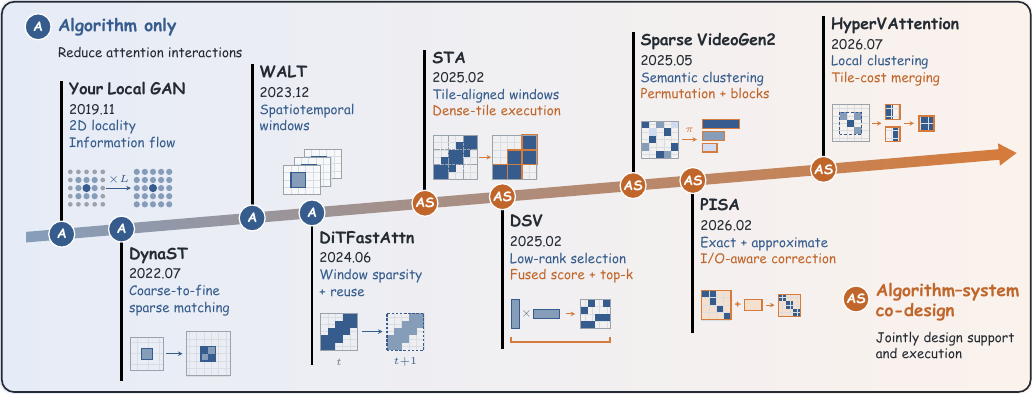}
  \caption{Development timeline of representative sparse-attention methods and systems, showing a shift from algorithmic sparsity design toward tighter algorithm--system co-design for more structured and efficient sparse execution.}
  \label{fig:timeline}
\end{figure*}

Reducing the number of selected interactions, however, does not by itself determine the end-to-end acceleration. A sparse support can leave many hardware tiles partially occupied, while evidence prediction, selection, indexing, and feature rearrangement consume part of the computation saved. Recovering omitted contributions can introduce additional arithmetic and memory traffic. These costs connect the algorithm to the system that executes it. Sliding Tile Attention aligns shared neighborhoods with dense execution tiles; Sparse VideoGen2 arranges semantic clusters into variable-length blocks; and PISA changes its correction statistics to reduce the memory cost of the approximation \cite{p250204507,p250518875,p260201077}. These costs gradually broaden the objective from achieving high logical sparsity with minimal fidelity loss to jointly designing accurate sparse supports and efficient ways to construct, represent, and execute them. Meeting this objective may require a more regular sparse pattern or a construction procedure with less overhead. These requirements therefore affect the algorithm's choices about which interactions to retain and how to identify them. Figure~\ref{fig:timeline} traces this development through representative sparse-attention methods, from purely algorithmic approaches to algorithm--system co-design.

Recent surveys discuss sparse attention from broader perspectives. Reviews of efficient transformers and transformer sparsity cover sparse-attention algorithms and, in some cases, related hardware considerations~\cite{tay2022efficienttransformers,sparsitytransformers2024survey}. The survey in \cite{attentionmethods2025survey} treats sparse attention as one category within a broader account of efficient attention, while the diffusion-oriented surveys in \cite{attentiondiffusion2025survey,efficientvideodiffusion2026survey} place it within broader accounts of attention design or video acceleration. As summarized in Table~\ref{tab:related-survey-comparison}, these surveys do not provide a dedicated treatment of sparse attention for visual generation that jointly organizes its algorithmic methods, system implementations, and their interactions. A focused review is therefore needed to analyze this literature through a unified algorithm--system framework.

\input{tables/related_survey_comparison}

To this end, we introduce a $4+1+3$ analytical framework. Four core dimensions characterize support construction: \emph{source}, \emph{evidence construction}, \emph{granularity}, and \emph{support specification}. Evidence construction derives decision information from the available source. At the specified granularity, support specification determines retained connectivity from this evidence and structural requirements. An additional dimension, \emph{auxiliary mechanisms}, covers reuse and correction. Three system dimensions---\emph{data organization}, \emph{execution}, and \emph{dataflow}---explain how these decisions become executable work. We examine mechanisms that implement sparse connectivity or address costs arising from its construction and evaluation.

Static support enables advance preparation of metadata and layouts. Dynamic support introduces runtime estimation and adaptation, while both regimes must account for sparse tile utilization, operand access, and workload imbalance. These execution costs can feed back into support granularity, grouping, and approximation choices. The main contributions are as follows.

\begin{itemize}
  \item \textbf{A unified analytical framework.}
  We introduce a $4+1+3$ framework that connects support construction, auxiliary mechanisms, and system implementation. The framework provides a common basis for comparing the mechanisms and design choices of different methods.

  \item \textbf{A systematic review from sparse algorithms to co-design.}
  We analyze sparse-attention methods for visual generation in detail, from their algorithmic mechanisms to their execution designs. We explain how algorithmic choices affect execution cost and how system requirements in turn shape those choices.

  \item \textbf{Evaluation and future directions.}
  We relate sparse-attention design to the requirements of visual-generation tasks and summarize benchmarks and metrics for generation quality and acceleration efficiency. We identify open challenges in support reliability, error accumulation, hardware portability, and integration with complementary acceleration techniques.
\end{itemize}

%% file: tables/related_survey_comparison.tex
\begin{table}[t]
  \caption{Comparison with related surveys that discuss sparse attention. A check mark indicates a primary topic under Topic Focus, systematic coverage of the corresponding methods under Method Coverage, or a framework linking algorithmic and system designs.}
  \label{tab:related-survey-comparison}
  \centering
  \begingroup
  \newcolumntype{Z}{>{\centering\arraybackslash}X}
  \newcommand{\surveyyes}{\ensuremath{\checkmark}}
  \newcommand{\surveyno}{\ensuremath{\times}}
  \newcommand{\surveyours}{\raisebox{-0.05ex}{\scalebox{1.16}{\ensuremath{\boldsymbol{\checkmark}}}}}
  \setlength{\tabcolsep}{1.4pt}
  \renewcommand{\arraystretch}{1.16}
  \fontsize{6.6}{7.8}\selectfont
  \begin{tabularx}{\columnwidth}{@{}>{\RaggedRight\arraybackslash}p{2.25cm}*{5}{Z}@{}}
    \rowcolor{black}
    \color{white}\bfseries Survey &
    \multicolumn{2}{c}{\color{white}\bfseries Topic Focus} &
    \multicolumn{2}{c}{\color{white}\bfseries Method Coverage} &
    \color{white}\bfseries\shortstack{Alg.--Sys.\\Analysis} \\
    \rowcolor{black!14}
    & \bfseries\shortstack{Sparse\\Attention}
    & \bfseries\shortstack{Visual\\Generation}
    & \bfseries\shortstack{Algorithmic\\Methods}
    & \bfseries\shortstack{System\\Methods}
    & \bfseries\shortstack{Unified\\Framework} \\
    \midrule
    \emph{Efficient Transformers}~\cite{tay2022efficienttransformers}
      & \surveyno & \surveyno & \surveyyes & \surveyno & \surveyno \\
    \emph{Sparsity in Transformers}~\cite{sparsitytransformers2024survey}
      & \surveyno & \surveyno & \surveyyes & \surveyyes & \surveyno \\
    \emph{Attention in Diffusion Models}~\cite{attentiondiffusion2025survey}
      & \surveyno & \surveyyes & \surveyyes & \surveyno & \surveyno \\
    \emph{Efficient Attention Methods}~\cite{attentionmethods2025survey}
      & \surveyno & \surveyno & \surveyyes & \surveyyes & \surveyno \\
    \emph{Efficient Video Diffusion Models}~\cite{efficientvideodiffusion2026survey}
      & \surveyno & \surveyyes & \surveyyes & \surveyno & \surveyno \\
    \rowcolor{black!28}
    \textbf{Ours}
      & \surveyours & \surveyours & \surveyours & \surveyours & \surveyours \\
    \bottomrule
  \end{tabularx}
  \endgroup
\end{table}

%% file: sections/01_background_preliminaries.tex
\section{Acceleration Methods for Visual Generation}
\label{sec:background}
\label{sec:related-methods}

\begin{figure*}[!t]
  \centering
  \includegraphics[width=\textwidth]{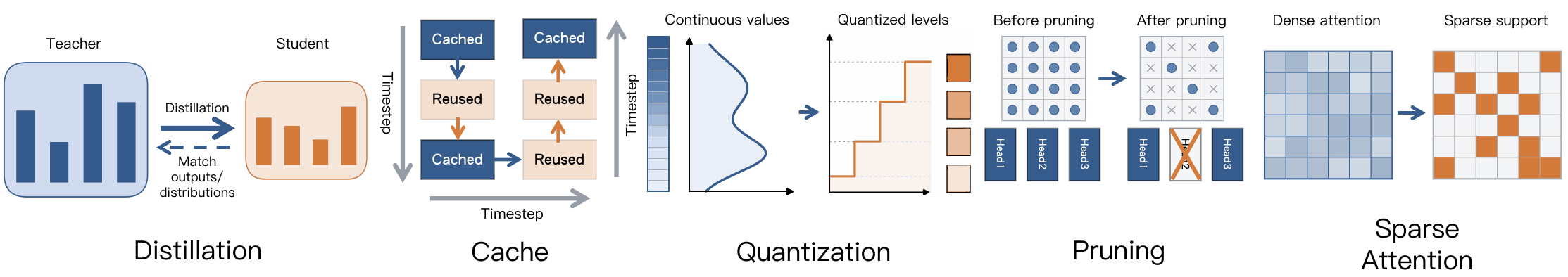}
  \caption{Five approaches to accelerating visual generation. Distillation trains a smaller student model, caching reuses intermediate results across timesteps, quantization lowers numerical precision, pruning removes selected model components, and sparse attention computes attention over selected query–key pairs.}
  \label{fig:technique-comparison}
\end{figure*}

Visual generation models incur substantial inference costs because they use large neural networks to synthesize high-dimensional image or video content, often through repeated denoising steps or sequential predictions. Acceleration methods address this cost at different levels. Distillation uses teacher supervision to reduce sampling steps, model size, or both, while cache reuse avoids recomputing intermediate results. Quantization lowers numerical precision, and pruning removes selected parameters or computational units. Figure~\ref{fig:technique-comparison} illustrates the intervention points of these four approaches alongside sparse attention.

\subsection{Distillation}
Distillation transfers a teacher's generative capabilities to a student with fewer sampling steps, a smaller network, or both. For model compression, BK-SDM \cite{kim2024bksdm} removes U-Net blocks and uses feature distillation to transfer the teacher's representations to the compact student. For sampling acceleration, Progressive Distillation \cite{salimans2022progressive} trains one student update to match two consecutive teacher updates and halves the sampling steps at each distillation stage. PDD \cite{shaul2026pdd} supervises a parallel decoder with interval mean velocities estimated from the teacher. The student predicts these velocities jointly and advances several sampling intervals with one network evaluation. At the distribution level, DMD \cite{yin2024dmd} combines score-based distribution matching with regression to teacher-generated outputs to train a one-step generator. TMD \cite{nie2026tmd} applies distribution matching distillation to a video student with a lightweight conditional flow head. This head performs several refinement updates using shared backbone features, so each refinement avoids another full backbone evaluation.

\subsection{Cache Reuse}
Cache reuse avoids repeated computation by storing intermediate results for use in subsequent generation steps. DeepCache \cite{deepcache2023} reuses high-level U-Net features across denoising steps and recomputes only the shallow network path between cache refreshes. For transformer backbones, $\Delta$-DiT \cite{deltadit2024} caches the input--output residuals of block groups, targeting rear blocks early in sampling and front blocks later. PAB \cite{pab2024} reuses spatial, temporal, and cross-attention outputs over different intervals according to their stability across denoising steps. TeaCache \cite{teacache2024} estimates output changes from rescaled differences between timestep-modulated inputs and refreshes cached transformer residuals when the accumulated estimate exceeds a threshold. Extending reuse to forecasting, TaylorSeer \cite{taylorseer2025} uses cached features and finite differences to approximate temporal derivatives and predict later features through Taylor expansion.

\subsection{Quantization}
Quantization represents model weights or activations with fewer bits to reduce storage and data movement and enable low-bit computation. PTQ4DM \cite{ptq4dm2023} calibrates 8-bit quantization using denoising samples from multiple timesteps to capture changing activation distributions. Q-Diffusion \cite{qdiffusion2023} combines timestep-aware calibration with separate quantization of features concatenated at U-Net skip connections, enabling 4-bit weights with 8-bit activations. For diffusion transformers, ViDiT-Q \cite{viditq2024} combines dynamic per-token activation quantization, channel balancing through scaling and rotation, and mixed precision based on layer and timestep sensitivity. SVDQuant \cite{svdquant2025} shifts activation outliers into weights and uses SVD to absorb them into a 16-bit low-rank branch; the residual branch uses 4-bit weights and activations. Its Nunchaku engine fuses low-rank and low-bit kernels to reduce additional memory traffic.

\subsection{Pruning}
Pruning reduces inference cost by removing selected parameters or computational units, such as channels, attention heads, and layers. Diff-Pruning \cite{diffpruning2023} aggregates gradients from selected diffusion timesteps to compute Taylor-based channel importance, then removes low-scoring channels and fine-tunes the reduced network. TinyFusion \cite{tinyfusion2025} targets transformer depth: it jointly optimizes differentiable layer-mask sampling and weight updates to select a shallow model with strong recoverability after fine-tuning. MosaicDiff \cite{mosaicdiff2025} uses the noise schedule to divide denoising into stages with different pruning budgets. For each stage, it uses calibration data and second-order scores to prune attention heads and MLP neurons. It also adjusts the retained weights to compensate for pruning errors without fine-tuning.

\section{Sparse Attention Preliminaries}
\label{sec:foundations}
\label{sec:sparse-attention-preliminaries}

Although the preceding techniques reduce inference cost along different dimensions, they generally leave the dense pairwise interactions inside each executed attention layer intact. Sparse attention addresses this remaining bottleneck by replacing all-to-all communication with a selected query--key support. It retains the complete token sequence and generation trajectory while avoiding redundant pairwise computation, making attention more scalable with image resolution, video length, and autoregressive context.

\subsection{Sparse Attention Formulation}
Attention computes a weighted sum of value vectors, with weights determined by query--key compatibility. For a single head, let $Q\in\R^{N_q\times d}$, $K\in\R^{N_k\times d}$, and $V\in\R^{N_k\times d_v}$ denote the query, key, and value matrices. Here, $N_q$ and $N_k$ count queries and keys, $d$ is the query/key dimension, and $d_v$ is the value dimension. Self-attention derives these matrices from the same input sequence, whereas cross-attention derives queries and key--value pairs from different sequences.

Scaled dot-product attention computes the score matrix $S$, normalizes each row to obtain attention weights $P$, and aggregates the values into the output $Y$:
\begin{equation}
 S=\frac{QK^{\mathsf T}}{\sqrt d}+B,\qquad
 P=\operatorname{softmax}(S),\qquad Y=PV,
 \label{eq:dense-attention}
\end{equation}
where $B$ contains additive biases and the attention mask. Let $\mathcal A_i$ denote the key positions that query $i$ is allowed to attend to. Setting $B_{ij}=-\infty$ for $j\notin\mathcal A_i$ assigns zero attention weight to these positions. For unrestricted attention, $\mathcal A_i=\{1,\ldots,N_k\}$; for causal self-attention over a sequence, $\mathcal A_i=\{1,\ldots,i\}$.

For each query $i$, sparse attention retains a nonempty subset $\supp_i\subseteq\mathcal A_i$ of admissible key positions. The support $\supp$ collects the retained query--key connections, and the binary mask $M$ records their locations:
\begin{equation}
 \supp=\{(i,j):j\in\supp_i\},\qquad
 M_{ij}=\mathbf{1}[j\in\supp_i].
 \label{eq:support-operator}
\end{equation}
Given this support, sparse attention evaluates the retained query--key scores, applies softmax over the selected keys, and aggregates their values:
\begin{equation}
 \widetilde P_{ij}=
 \begin{cases}
 \displaystyle\frac{\exp(S_{ij})}
 {\sum_{k\in\supp_i}\exp(S_{ik})},&j\in\supp_i,\\[1mm]
 0,&j\notin\supp_i,
 \end{cases}
 \qquad \widetilde Y=\widetilde P V.
 \label{eq:sparse-attention}
\end{equation}

Omitting connections removes their contributions to value aggregation and rescales the retained attention weights through renormalization, which can introduce deviations from the dense output. Correction mechanisms~\cite{p251119835,p260201077} can approximate the omitted contributions and compensate for these deviations to help preserve generation fidelity.

\subsection{Sparsity and Acceleration}
Let $M_{\mathrm{keep}}=\sum_i|\supp_i|$ count retained query--key connections and $M_{\mathrm{adm}}=\sum_i|\mathcal A_i|$ count all admissible connections. The logical density and sparsity are
\begin{equation}
 \rho_{\mathrm{log}}=\frac{M_{\mathrm{keep}}}{M_{\mathrm{adm}}},
 \qquad \sigma_{\mathrm{log}}=1-\rho_{\mathrm{log}}.
 \label{eq:logical-density}
\end{equation}
For a given support, the score and value products require $\mathcal O(M_{\mathrm{keep}}(d+d_v))$ arithmetic when only retained pairs are evaluated, compared with $\mathcal O(M_{\mathrm{adm}}(d+d_v))$ for all admissible pairs.

Many GPU kernels evaluate attention through dense matrix tiles. For a given attention-product stage, let $\mathcal B_{\mathrm{exec}}$ denote the executed query--key tiles, with query and key extents $B_q^{(a)}$ and $B_k^{(b)}$ including padding. The number of pair positions covered by these tiles is
\begin{equation}
 W_{\mathrm{tile}}=
 \sum_{(a,b)\in\mathcal B_{\mathrm{exec}}}
 B_q^{(a)}B_k^{(b)}.
 \label{eq:work}
\end{equation}
Partially active tiles and boundary padding can make $W_{\mathrm{tile}}$ exceed $M_{\mathrm{keep}}$, as dense tile operations also compute positions outside the retained support~\cite{p250204507}. The $QK^{\mathsf T}$ and $PV$ products can use different tilings, so this count applies separately to each stage. Supports with the same logical density can therefore require different amounts of arithmetic.

Support construction and output correction add computation, while index generation, tensor rearrangement, and key/value gathers add data-handling costs. The resulting attention latency depends on both arithmetic work and memory access.

To relate attention acceleration to generation latency, let $\alpha$ be the fraction of baseline generation time spent in the attention kernels being replaced. Let $r_{\mathrm{attn}}$ be the ratio of their baseline to sparse execution times, summed over all calls, and let $\omega$ be the additional latency outside these kernels, normalized by baseline generation time. With the remaining computation unchanged, the end-to-end speedup is
\begin{equation}
 G_{\mathrm{e2e}}=
 \frac{1}{(1-\alpha)+\alpha/r_{\mathrm{attn}}+\omega}.
 \label{eq:amdahl}
\end{equation}
End-to-end acceleration requires $\alpha(1-1/r_{\mathrm{attn}})>\omega$: the time saved in the attention kernels must exceed the additional overhead.

%% file: sections/02_framework.tex
\section{A 4+1+3 Framework}
\label{sec:framework}
\begin{figure*}[!t]
  \centering
  \includegraphics[width=\textwidth]{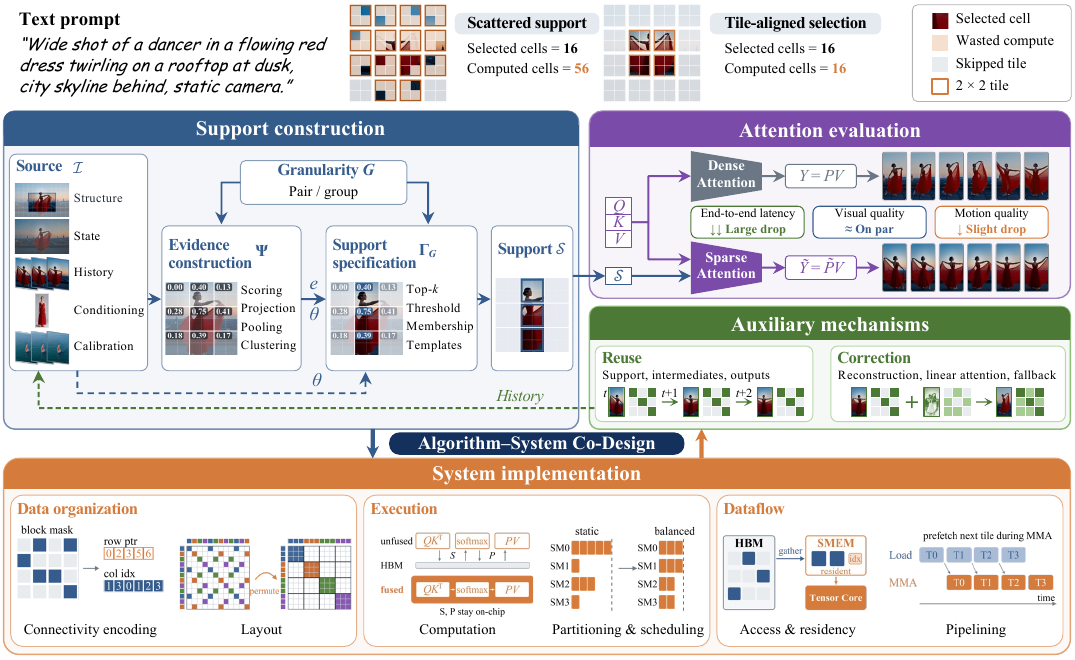}
  \caption{The $4+1+3$ framework for sparse attention. The source, evidence construction, granularity, and support specification determine which query–key pairs are retained. Reuse avoids repeating work across denoising steps, and correction compensates for contributions from omitted pairs. Data organization, execution, and dataflow determine how the sparse attention workload runs on hardware. The tile example shows how the shape of the support affects computation. Both patterns retain 16 query–key pairs, but with $2 \times 2$ tiles, the scattered pattern requires computing 56 entries and the tile-aligned pattern requires only 16.}
  \label{fig:framework-overview}
\end{figure*}

To systematically analyze how sparse attention constructs its support and translates sparsity into practical acceleration, we propose a $4+1+3$ framework for its algorithmic design and system implementation. On the algorithmic side, four dimensions---\emph{source}, \emph{evidence construction}, \emph{granularity}, and \emph{support specification}---characterize the construction of sparse support $\supp$. An additional dimension, \emph{auxiliary mechanisms}, covers reuse and correction that reduce repeated computation and compensate for approximation error. On the system side, three dimensions---\emph{data organization}, \emph{execution}, and \emph{dataflow}---describe data representation, computational implementation, and data delivery. Figure~\ref{fig:framework-overview} illustrates how these dimensions interact in a visual generation example. Table~\ref{tab:sparse_support_construction_taxonomy} summarizes these dimensions and their representative mechanisms.

\subsection{Support Construction}
Support construction is characterized by four components: the \emph{source} $\mathcal I$, \emph{evidence construction} $\Psi$, \emph{granularity} $G$, and \emph{support specification} $\Gamma$. Evidence construction derives decision information from the source, and support specification combines this evidence with configured structural requirements at the specified granularity. The resulting support template specifies the retained query--key connectivity $\supp$, equivalently represented by the binary mask $M$ in Eq.~\eqref{eq:support-operator}. The source comprises information such as structural context, current state, history, conditioning, and offline calibration traces. Evidence construction and support specification are expressed as
\begin{align}
 e &= \Psi(\mathcal I;G,\xi),\nonumber\\
 \supp &= \Gamma_G(e;\theta,\xi),
 \label{eq:construction}
\end{align}
where $e$ denotes decision evidence and $\theta$ specifies structural settings, decision policies, candidate template sets, and required connections. Structural settings originate from source information, while the policies determine budget and template parameters from fixed configurations or evidence $e$. The auxiliary variable $\xi$ collects content-independent random draws used by evidence construction or support specification; deterministic operations are independent of $\xi$. The subscript $G$ specifies the unit of connectivity decisions, while $\Gamma_G$ returns a support template with $\varnothing\ne\supp_i\subseteq\mathcal A_i$. When evidence construction is unnecessary, $e=\varnothing$ and $\Gamma_G$ instantiates the configured structural pattern.

\paragraph{Source}
Source specifies the information available for evidence construction and support specification. Structural context includes connectivity priors, such as spatial locality in two-dimensional images and spatiotemporal locality in videos. It also includes layer/head identifiers and the prescribed noise schedule, which specify model organization and the generation protocol. Current state encompasses the current timestep and noise level, together with sample-dependent quantities such as $Q$, $K$, $V$, attention scores, and intermediate statistics maintained by online softmax. These quantities can inform stage-specific support configurations and content-dependent importance estimates. Conditioning contributes externally supplied semantic, visual, or geometric cues, such as text prompts, reference frames, and camera poses. History carries information from earlier computations, such as features, statistics, supports, or outputs, so subsequent decisions can exploit previously observed behavior. Calibration traces record offline observations of attention behavior, generation fidelity, or execution cost and provide empirical evidence for configuring templates, budgets, or evidence models. A method may combine these sources to relate structural expectations to sample-specific evidence.

\paragraph{Evidence Construction}
Evidence construction $\Psi$ transforms source information into decision evidence $e$, such as relevance scores, approximation-error or cost estimates, and group assignments. Statistical scoring uses explicit formulas to derive similarity estimates, error indicators, or softmax-related bounds. Projection forms compact feature representations in which compatibility can be estimated at lower cost. Sampling estimates attention behavior from a subset of tokens or blocks. Pooling aggregates features or scores within groups to produce coarser evidence. These operations can be composed. For example, $Q$ and $K$ can first be projected into a lower-dimensional space and then pooled within token groups to estimate coarse attention scores.

Clustering summarizes feature similarity through group assignments, for example using $k$-means centroids. Hashing, such as LSH, uses bucket assignments as a proxy for similarity. Scoring networks learn mappings from input features to importance scores or routing evidence.

\paragraph{Granularity}
Granularity specifies whether a decision retains an individual query--key pair or a group of connections, such as a block, cluster, frame, or chunk. Task structure and hardware efficiency influence this choice. Each selected group contributes its admissible query--key connections to $\supp$; partitioning subsequently assigns their computation to hardware tiles. Given source and granularity, evidence construction and support specification define the central decision mechanism.

\paragraph{Support Specification}
Support specification $\Gamma_G$ determines the final support template from evidence $e$ and configuration $\theta$. Ranked selection orders candidates by relevance and retains a prefix using top-$k$ or top-$p$: top-$k$ specifies the retained count, while top-$p$ specifies the cumulative attention mass. Threshold rules retain candidates whose evidence meets a cutoff. The count, mass, and threshold parameters may be predefined, scheduled, or adapted using evidence. Membership rules use cluster, bucket, or route assignments to determine connections. Randomized selection uses $\xi$ for pivot sampling, window shifts, or score perturbations to diversify connectivity and broaden coverage of potentially relevant tokens.

Mandatory links retain designated connections, such as global anchors, sink tokens, self links, or links to a conditioning modality. They impose $\supp_{\mathrm{req}}\subseteq\supp$, where the required set $\supp_{\mathrm{req}}$ is specified by $\theta$. These mechanisms can be composed: top-$k$ connections, random additions, and mandatory anchors can jointly form one support template.

Structural templates instantiate specified patterns, such as windows, radial neighborhoods, stripes, or blocks. Their parameters may be fixed, scheduled, or updated from online source information and evidence.

Template search selects among candidate patterns or parameter configurations using error, fidelity, or cost evidence, then instantiates the selected template. Search can operate online or use offline calibration evidence to determine a configuration reused at deployment.

Figure~\ref{fig:sparse-support-patterns} illustrates representative connectivity patterns that emerge from the combined choices across these four dimensions.

\input{tables/framework_dimensions}

\subsection{Auxiliary Mechanisms}
Reuse and correction are optional enhancements to sparse attention. They augment support construction and attention evaluation by reducing repeated computation or compensating for sparsification error, and can be used separately or together.

\paragraph{Reuse}
Reuse retains support, construction intermediates, or attention outputs for subsequent use. Support reuse carries forward masks or instantiated templates for attention evaluated on current tensors. Construction intermediates include scores and normalization summaries used in evidence computation, together with feature summaries and group assignments used to guide refresh decisions or organize token partitions. Reusing these intermediates reduces repeated construction work while allowing support to be updated. For example, cached $k$-means centroids can initialize a new clustering step, whereas cached assignments preserve the preceding partition. Attention-output reuse retains the features obtained by weighting and aggregating $V$ and supplies stored or extrapolated outputs to replace selected query evaluations. Output caching also applies to dense attention and can complement sparsification. Reuse can occur between computational stages or across generation steps, with periodic or state-dependent refresh. Results retained from earlier evaluations provide a history source.

\paragraph{Correction}
Correction compensates for the loss of omitted keys' value contributions and the change in retained weights caused by sparse normalization. Output correction adds measured or reconstructed dense--sparse residuals to computed sparse outputs. Sharing these residuals combines correction with reuse~\cite{p240608552}, while low-rank reconstruction estimates or filters the output discrepancy. Normalization rescaling adjusts retained weights using estimates of full-context attention mass. Contribution recovery supplies omitted context through cached contribution estimates or approximations built from $Q/K/V$ features. Pooled or linear attention supplies approximate context, while Taylor methods estimate omitted value and normalization terms~\cite{p260201077}. Global approximation branches can cover the entire context, while block compensation targets regions excluded from exact computation. Dense fallback restores exact attention for selected queries or blocks when approximation error or task sensitivity warrants it.

\subsection{System Implementation}
The runtime benefit of sparse attention depends on more than the number of retained interactions. Indexing and data rearrangement, low hardware utilization, and memory traffic can offset the computational savings from sparsification. Realizing these savings therefore requires system designs that match sparse computation to the hardware. We characterize these designs through three dimensions: \emph{data organization}, \emph{execution}, and \emph{dataflow}. Data organization arranges sparse metadata and tensor operands in representations and layouts suited to hardware access and matrix computation. Execution optimizes operators and organizes parallel work to use hardware compute resources efficiently. Dataflow coordinates data access, transfer, and residency to supply computation while reducing memory and communication costs. These dimensions interact throughout support construction, attention evaluation, and auxiliary computation. Their implementation costs can also feed back into support granularity, selection rules, and correction design.

\paragraph{Data Organization}
Sparse connectivity can form contiguous runs, fragmented selections, or content-dependent groups. Representing and arranging these structures inefficiently increases metadata, traversal, and operand-access costs, and can introduce padded or redundant work in subsequent attention computation. Data organization addresses these problems through connectivity encoding and layout, matching sparse metadata and operand organization to hardware access patterns and compute units.

\emph{Connectivity encoding} represents the retained connections for efficient traversal. List encoding enumerates selected keys or blocks without scanning inactive positions. Interval encoding compresses consecutive selections into ranges, reducing metadata when support contains long runs. Bitmap encoding packs membership flags and supports compact tests or enumeration of active blocks. Implicit indexing derives connections from structural rules, avoiding explicit storage when geometry or a fixed pattern determines the selected positions. Segment offsets locate variable-length lists or groups, while reverse indices expose the connectivity needed for key-to-query traversal during backward computation.

\emph{Layout} addresses the fragmentation and conflicts created when sparse selections are scattered in the original tensor order. Permutation places related tokens or blocks near one another. Packing assembles selected or grouped operands into contiguous regions so that computation can operate on compact subproblems. Stride remapping exposes structured neighborhoods through address views when existing storage permits it, reducing explicit rearrangement. Bank mapping distributes operands across memory banks to align sparse accesses with parallel workers and limit conflicts. The appropriate transformation depends on the support structure and when its membership becomes available.

\paragraph{Execution}
Sparse algorithms introduce selection and correction operators, irregular computation shapes, and unequal amounts of work. We describe the corresponding execution choices through computation, partitioning, and scheduling.

\begin{figure*}[!t]
  \centering

\input{figures/review_forest}

  \caption{Structure of the framework-guided review with representative papers for each topic.}
  \label{fig:review-forest}
\end{figure*}
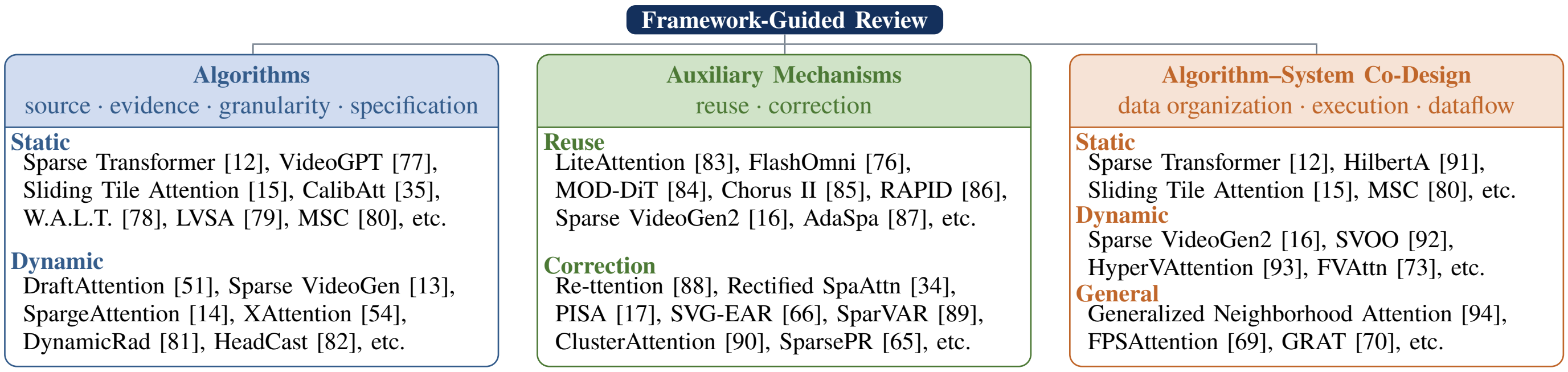

\emph{Computation} addresses the extra operations, irregular input shapes, and intermediate traffic introduced by sparse support construction and evaluation. Operator optimization changes how the required computation is expressed through operators, including operator selection, substitution, and decomposition, so that sparse workloads can use efficient computational primitives. Kernel optimization improves the device implementation of these operators through kernel rewriting, specialized computation loops, and fusion, reducing unnecessary work, launches, and intermediate memory traffic. Mixed-precision computation assigns different precisions to operands, multiplication, and accumulation to reduce execution cost while controlling numerical error.

\emph{Partitioning} addresses underfilled hardware tiles and small or irregular sparse subproblems. Tiling determines the query and key ranges processed by a task, connecting logical support to hardware-efficient compute units. Query grouping combines queries with overlapping selected operands to improve reuse and tile occupancy. Batching combines compatible sparse subproblems, increasing useful work per launch. These choices determine task geometry; when they also change the retained connections, they constitute feedback into algorithm design.

\emph{Scheduling} addresses unequal execution times after tasks have been formed. Cost-based dispatch uses sparse workload estimates to order or assign tasks. Persistent scheduling lets resident workers acquire new tasks as earlier ones finish, reducing idle time. Task migration redistributes work across devices when the current sparse support creates imbalance. Scheduling changes task ownership or order, while partitioning determines the units being scheduled.

\paragraph{Dataflow}
Selective connectivity introduces index-dependent retrieval, irregular transfers, and additional metadata or intermediate states. Dataflow addresses the resulting latency, traffic, and memory pressure through access, pipelining, and residency.

\emph{Access} optimizes retrieval of the operands and metadata required by sparse computation. Cooperative gather shares index decoding and address generation across threads to accelerate irregular retrieval. Coalesced access coordinates neighboring requests so that selected operands or correction summaries use fewer memory transactions. Selective communication applies the same principle across devices, exchanging data required by the support rather than complete remote shards.

\emph{Pipelining} addresses stalls between sparse data preparation, transfer, and computation. Prefetching requests upcoming metadata or operands before consumption. Asynchronous transfer lets delivery proceed alongside independent work, and double buffering separates the data currently being consumed from the next data being loaded. Warp specialization assigns sparse preparation and loading to producer workers and arithmetic to consumers, allowing the stages to progress concurrently. Their usefulness depends on whether the support provides enough advance knowledge and independent work for overlap.

\emph{Residency} addresses the memory traffic and capacity costs of sparse scores, indices, masks, and historical operands. On-chip retention keeps short-lived intermediates close to their consumers, avoiding unnecessary writes and reloads. Offloading moves large or infrequently used states to host memory and restores the required data before use. These storage decisions implement the needs of support construction, backward computation, and algorithmic reuse; they can also constrain the size of the sparse working set.

Tables~\ref{tab:framework-examples-algorithm} and~\ref{tab:framework-examples-system} summarize representative methods.

%% file: tables/framework_dimensions.tex
\newcommand{\supportband}[1]{%
  \rowcolor{black!12}
  \multicolumn{2}{@{}l@{}}{%
    \hspace{1.1mm}\color{black}\bfseries\strut #1\strut\hspace{1mm}}\\[-0.1mm]
}
\newcommand{\supporttaxonomycaption}{%
The $4+1+3$ design space for sparse attention in visual generation. The four support construction dimensions describe information sources, how evidence is derived, the unit of selection, and the rules specifying which connections are retained. The auxiliary dimension covers reuse and correction. The three system dimensions describe how sparse data is organized, how workloads are executed, and how data moves through the system. The framework provides a basis for analyzing sparse attention designs. The table draws on representative methods from the literature.%
}
\newcommand{\supporttaxonomytable}{%
\begingroup
\setlength{\tabcolsep}{3.6pt}
\renewcommand{\arraystretch}{1.10}
\arrayrulecolor{black}
\hyphenpenalty=10000
\exhyphenpenalty=10000
\emergencystretch=.55em
\fontsize{8}{9.4}\selectfont
\begin{tabularx}{\linewidth}{@{}>{\RaggedRight\arraybackslash}p{4.25cm}>{\RaggedRight\arraybackslash}X@{}}
\rowcolor{black}
\multicolumn{1}{@{}c}{\color{white}\bfseries Component} &
\multicolumn{1}{c@{}}{\color{white}\bfseries Choices / Mechanisms} \\

\supportband{1. Source}
\textbf{Structural Context} & Priors~\cite{p250204507}; Layer/head IDs~\cite{p260305503}; \\
\textbf{Current State} &
Hidden states~\cite{p251018692}; $Q/K/V$~\cite{p260201077}; Scores~\cite{p250218137}; Softmax state~\cite{p260616317}; \\
\textbf{History} & Past features~\cite{p250618226}; Past statistics~\cite{p250518875}; Past supports~\cite{p260514513}; Past outputs~\cite{p240608552} \\
\textbf{Conditioning} & Text~\cite{p260201801}; Image~\cite{p201207288}; Video~\cite{p260613872}; Audio~\cite{p221209478}; Supplied motion~\cite{p251002617}; Supplied pose~\cite{p260522718}; Supplied rays~\cite{p260207854}; \\
\textbf{Calibration Traces} & Offline samples~\cite{p250204507}; $Q/K$ traces~\cite{p260530325}; Attention traces~\cite{p260305503}; Fidelity measurements~\cite{p260623019}; Cost measurements~\cite{p250322796} \\

\supportband{2. Evidence Construction}
\textbf{Statistical Scoring} & Similarity score~\cite{p250514708}; 
Error estimate~\cite{p250603065}; Frequency estimate~\cite{p260305503}; Softmax bound~\cite{p260616317} \\
\textbf{Projection} & Low-rank projections~\cite{p250207590}; Random projections~\cite{p260201801} \\
\textbf{Sampling} & Strided downsampling~\cite{p250316428}; Random sampling~\cite{p250201776}; Representative sampling~\cite{p250810774} \\
\textbf{Pooling} & Mean aggregation~\cite{p250514708}; Max aggregation~\cite{p260530325}; Sum aggregation~\cite{p260305503} \\
\textbf{Clustering} & Batch $k$-means~\cite{p250518875}; Online $k$-means~\cite{p200305997} \\
\textbf{Hashing} & LSH~\cite{p260201801} \\
\textbf{Scoring Network} & MLP scorers~\cite{p260530325}; Gating heads~\cite{p251018692}; \\

\supportband{3. Granularity}
\textbf{Token Pair} & One query--key connection~\cite{p201207288} \\
\textbf{Token Group} & Blocks~\cite{p250901085}; Regions~\cite{p250514708}; Clusters~\cite{p250518875}; Buckets~\cite{p260201801}; Frames~\cite{p260207854}; Chunks~\cite{p260823565}; Memory segments~\cite{p260522718} \\

\supportband{4. Support Specification}
\textbf{Ranked Selection} & Top-$k$~\cite{p250514708}; Top-$p$~\cite{p250518875} \\
\textbf{Threshold} & Fixed $\tau$~\cite{p260305503}; Adaptive $\tau$~\cite{p260724027} \\
\textbf{Membership} & Cluster~\cite{p200305997}; Bucket~\cite{p260201801}; Route~\cite{p251018692}; Partition~\cite{p241200131} \\
\textbf{Randomized Selection} & Random pivots~\cite{p210513290}; Window shifts~\cite{p221209478}; Score perturbation~\cite{p260412219} \\
\textbf{Mandatory Links} & Global anchors~\cite{p250619852}; Sinks~\cite{p260823565}; Self links~\cite{p260207854}; Modality-wide links~\cite{p210513290} \\
\textbf{Structural Templates} & Window~\cite{p250204507}; Radial~\cite{p250619852}; Diagonal~\cite{p250603065}; Stripe~\cite{p250603065}; Block~\cite{p190410509}; Axial~\cite{p210414806}; Dilated~\cite{p190410509} \\
\textbf{Template Search} & Error~\cite{p250204507}; Fidelity~\cite{p260623019}; Execution cost~\cite{p250322796} \\

\supportband{5. Auxiliary Mechanisms}
\textbf{Reuse} & Support~\cite{p260514513}; Construction intermediates~\cite{p250518875}; Attention outputs~\cite{p240608552} \\
\textbf{Correction} & Residual sharing~\cite{p240608552}; Normalization rescaling~\cite{p251119835}; Low-rank reconstruction~\cite{p260818484}; Pooled attention~\cite{p260308982}; Linear attention~\cite{p250924006}; Taylor correction~\cite{p260201077}; Dense fallback~\cite{p250603065} \\

\supportband{6. Data Organization}
\textbf{Connectivity Encoding} & List encoding~\cite{p250901085}; Interval encoding~\cite{p260305503}; Bitmap encoding~\cite{p260623019}; Implicit indexing~\cite{p250204507} \\
\textbf{Layout} & Permutation~\cite{p250518875}; Packing~\cite{p251018692}; Stride remapping~\cite{p210513290}; Bank mapping~\cite{p260525798} \\
\supportband{7. Execution}
\textbf{Computation} & Operator optimization~\cite{p260525798}; Kernel optimization~\cite{p260201077}; Mixed-precision computation~\cite{p250604648} \\
\textbf{Partitioning} & Tiling~\cite{p250204507}; Query grouping~\cite{p250514687}; Batching~\cite{p260803335} \\
\textbf{Scheduling} & Cost-based dispatch~\cite{p250916518}; Persistent scheduling~\cite{p250916518}; Task migration~\cite{p260716190} \\
\supportband{8. Dataflow}
\textbf{Access} & Cooperative gather~\cite{p250916518}; Coalesced access~\cite{p260201077}; Selective communication~\cite{pdsaiclr2026} \\
\textbf{Pipelining} & Prefetching~\cite{p260823565}; Asynchronous transfer~\cite{p260530325}; Double buffering~\cite{p250616054}; Warp specialization~\cite{p250204507} \\
\textbf{Residency} & On-chip retention~\cite{p250925401}; Offloading~\cite{p260823565} \\\bottomrule
\end{tabularx}
\endgroup
}
\begin{table*}[!tp]
  \centering
  \supporttaxonomytable
  \caption{\supporttaxonomycaption}
  \label{tab:sparse_support_construction_taxonomy}
  \label{tab:dimensions}
\end{table*}

%% file: figures/review_forest.tex
\begingroup
\setlength{\parindent}{0pt}
\ifdefined\RFwidth\else\newlength{\RFwidth}\fi
\setlength{\RFwidth}{\linewidth}
\newcommand{\RFcite}[2]{\mbox{#1~\cite{#2}}}
\newcommand{\RFjump}[2]{\begingroup\hypersetup{hidelinks}\hyperref[#1]{#2}\endgroup}
\definecolor{RFink}{HTML}{14305C}
\definecolor{RFedge}{HTML}{6B7785}
\definecolor{RFalg}{HTML}{345C8B}\definecolor{RFalgwash}{HTML}{D1DCF0}
\definecolor{RFaux}{HTML}{4C7935}\definecolor{RFauxwash}{HTML}{D0E3C8}
\definecolor{RFsys}{HTML}{C0662A}\definecolor{RFsyswash}{HTML}{F6E3D6}
\colorlet{RFcur}{RFalg}\colorlet{RFcurwash}{RFalgwash}
\newcommand{\RFtheme}[1]{\colorlet{RFcur}{RF#1}\colorlet{RFcurwash}{RF#1wash}}
\fontsize{8}{9.2}\selectfont

\setlength{\fboxsep}{2pt}
\setlength{\fboxrule}{.4pt}
\newlength{\RFcardheight}
\setlength{\RFcardheight}{78pt}
\newcommand{\RFleaf}[4]{%
 \noindent{\color{RFcur}\textbf{\RFjump{#1}{#2}}}\par
 \noindent\hspace*{4pt}%
 \begin{tabular}[c]{@{}l@{}}#3\\#4\end{tabular}\par}
\newlength{\RFhdr}\setlength{\RFhdr}{24pt}
\newcommand{\RFcolumn}[4]{%
 \begin{minipage}[t]{.318\RFwidth}%
 \begin{tikzpicture}[baseline=(box.north), inner sep=0pt, outer sep=0pt]
 \node[anchor=north west] (box) at (0,0) {%
  \begin{minipage}[t]{\dimexpr.318\RFwidth-1pt\relax}%
   \raisebox{-1pt}[\height][\depth]{\parbox[c][\RFhdr][c]{\linewidth}{\centering{\bfseries\color{RFcur}\RFjump{#1}{#2}}\\[1pt]{\fontsize{8.4}{9.6}\selectfont\color{RFcur}#3\strut}}}\par
   \nointerlineskip
   \hspace*{2pt}\begin{minipage}[t][\RFcardheight][s]{\dimexpr\linewidth-4pt\relax}%
   \vspace{2pt}#4\vspace{2pt}%
   \end{minipage}\par\vspace{1pt}%
  \end{minipage}};
 \begin{scope}[on background layer]
  \begin{scope}
   \clip[rounded corners=4pt] (box.south west) rectangle (box.north east);
   \fill[white] (box.south west) rectangle (box.north east);
   \fill[RFcurwash] (box.north west) rectangle ([yshift=-\RFhdr]box.north east);
  \end{scope}
  \draw[RFcur, line width=.5pt, rounded corners=4pt] (box.south west) rectangle (box.north east);
  \draw[RFcur, line width=.5pt] ([yshift=-\RFhdr]box.north west) -- ([yshift=-\RFhdr]box.north east);
 \end{scope}
 \end{tikzpicture}%
 \end{minipage}}
\noindent\makebox[\RFwidth][c]{\begin{tikzpicture}[baseline=(r.base)]\node[fill=RFink, rounded corners=3pt, inner xsep=5pt, inner ysep=2pt, text=white, font=\bfseries] (r) {\RFjump{sec:review}{Framework-Guided Review}};\end{tikzpicture}}\par

\nointerlineskip
\noindent\makebox[\RFwidth][c]{\textcolor{RFedge}{\rule{.45pt}{3pt}}}\par
\nointerlineskip
\noindent\makebox[\RFwidth][c]{\textcolor{RFedge}{\rule{.682\RFwidth}{.45pt}}}\par
\nointerlineskip
\noindent\makebox[\RFwidth][c]{\makebox[.682\RFwidth][s]{\textcolor{RFedge}{\rule{.45pt}{3pt}}\hfill\textcolor{RFedge}{\rule{.45pt}{3pt}}\hfill\textcolor{RFedge}{\rule{.45pt}{3pt}}}}\par
\nointerlineskip
\noindent
\RFtheme{alg}\RFcolumn{sec:review-prior}{Algorithms}{source \textperiodcentered\ evidence \textperiodcentered\ granularity \textperiodcentered\ specification}{%
\RFleaf{sec:review-prior}{Static}{\RFcite{Sparse Transformer}{p190410509}, \RFcite{VideoGPT}{p210410157},}{%
\RFcite{Sliding Tile Attention}{p250204507}, \RFcite{CalibAtt}{p260305503},\\
\RFcite{W.A.L.T.}{p231206662}, \RFcite{LVSA}{p260531057}, \RFcite{MSC}{p241209828}, etc.}%
\vfill%
\RFleaf{sec:review-content}{Dynamic}{\RFcite{DraftAttention}{p250514708}, \RFcite{Sparse VideoGen}{p250201776},}{%
\RFcite{SpargeAttention}{p250218137}, \RFcite{XAttention}{p250316428},\\
\RFcite{DynamicRad}{p260420470}, \RFcite{HeadCast}{p260720125}, etc.}%
}%
\hfill
\RFtheme{aux}\RFcolumn{sec:review-auxiliary}{Auxiliary Mechanisms}{reuse \textperiodcentered\ correction}{%
\RFleaf{sec:review-reuse}{Reuse}{\RFcite{LiteAttention}{p251111062}, \RFcite{FlashOmni}{p250925401},}{%
\RFcite{MOD-DiT}{p260111641}, \RFcite{Chorus II}{p260625040}, \RFcite{RAPID}{prapidcvpr2026},\\
\RFcite{Sparse VideoGen2}{p250518875}, \RFcite{AdaSpa}{p250221079}, etc.}%
\vfill%
\RFleaf{sec:review-correction}{Correction}{\RFcite{Re-ttention}{p250522918}, \RFcite{Rectified SpaAttn}{p251119835},}{%
\RFcite{PISA}{p260201077}, \RFcite{SVG-EAR}{p260308982}, \RFcite{SparVAR}{p260204361},\\
\RFcite{ClusterAttention}{p260826965}, \RFcite{SparsePR}{p260818484}, etc.}%
}%
\hfill
\RFtheme{sys}\RFcolumn{sec:review-system-codesign}{Algorithm--System Co-Design}{data organization \textperiodcentered\ execution \textperiodcentered\ dataflow}{%
\RFleaf{sec:review-static-codesign}{Static}{\RFcite{Sparse Transformer}{p190410509}, \RFcite{HilbertA}{p250926538},}{%
\RFcite{Sliding Tile Attention}{p250204507}, \RFcite{MSC}{p241209828}, etc.}%
\RFleaf{sec:review-dynamic-codesign}{Dynamic}{\RFcite{Sparse VideoGen2}{p250518875}, \RFcite{SVOO}{p260318636},}{%
\RFcite{HyperVAttention}{p260703012}, \RFcite{FVAttn}{p260716190}, etc.}%
\vfill%
\RFleaf{sec:review-shared-sparse-constraints}{General}{\RFcite{Generalized Neighborhood Attention}{p250416922},}{%
\RFcite{FPSAttention}{p250604648}, \RFcite{GRAT}{p250514687}, etc.}%
}%
\par\endgroup

%% file: sections/03_unified_review.tex
\section{Sparse Attention: A Framework-Guided Review}
\label{sec:review}

We organize the algorithmic review into static and dynamic sparse attention according to whether support depends on runtime content. Using the $4+1+3$ framework, we analyze how evidence construction derives decision evidence from source information and how support specification combines this evidence with structural requirements to produce the final support mask at the chosen granularity. Reuse and correction complement these support decisions by reducing repeated computation and compensating for sparsification error. We then examine how the resulting algorithms interact with data organization, execution, and dataflow. The analysis traces a progression from reducing attention interactions while preserving generation fidelity to jointly designing support and execution, as estimation overhead, irregular computation, and data movement become limiting costs. Fig.~\ref{fig:review-forest} outlines the structure of the review.

\subsection{Static Sparse Attention Algorithms}

\label{sec:review-prior}
Static sparse attention uses support that is independent of runtime content. This support can be derived from structural priors, such as spatial locality and temporal continuity, or from recurring attention patterns identified across samples through offline analysis.

Within the framework, structural priors can directly configure support specification through templates and mandatory links, without a separate evidence-construction stage. Calibration instead uses offline traces as the source. Evidence construction derives quantities such as retained attention mass or approximation error from these traces, and support specification uses this evidence to select a reusable support configuration. Both routes allow support to be instantiated at deployment without content-dependent selection. Training may adapt the generator to the prescribed support or optimize its configuration before deployment.

\subsubsection{Architectural and Structural Patterns}
Architectural and structural patterns use structural context as the source for support construction, including locality priors, causal order, and modality or token roles. At deployment, these methods instantiate support without a separate evidence-construction stage ($e=\varnothing$). Support specification applies prescribed structural templates, optionally combined with mandatory links or content-independent random choices. Their principal differences lie in the priors adopted as source information and how support specification translates these priors into template geometry. These choices determine how local information and distant context remain accessible under sparse support.

Spatial locality supplies a source prior for image generation: nearby regions often require detailed interaction, while distant context remains necessary for global coherence. Image Transformer translates locality into one- and two-dimensional window templates~\cite{p180205751}. Sparse Transformer supplements local support with strided factors, allowing information to propagate beyond a single neighborhood~\cite{p190410509}. Your Local GAN examines these propagation paths and uses Manhattan-distance ordering to make the prescribed geometry respect two-dimensional proximity~\cite{p191112287}. Qihoo-T2X also specifies local and shifted window templates, with an auxiliary proxy-token branch supplying global context~\cite{p240904005}. Conditioning roles introduce an additional source prior. DALL\textperiodcentered E combines row, column, and convolutional templates with mandatory text support, whereas CogView combines local blocks and text support with random image pivots~\cite{p210212092,p210513290}. Under the shared source prior of two-dimensional spatial locality, differences within support specification concern the geometry of structural templates and their combination with mandatory conditioning links or randomized global coverage.

Spatial locality and temporal continuity provide related source priors for video generation. VideoGPT extends local and strided templates across spacetime~\cite{p210410157}. GODIVA instead specifies separate temporal, row, and column support, while Open-Sora separates within-frame spatial attention from temporal attention at corresponding positions~\cite{p210414806,p241220404}. These axial templates make some spacetime dependencies accessible through successive layers. N\"UWA uses joint three-dimensional neighborhoods, and W.A.L.T. alternates spatial and spatiotemporal windows, providing more direct access to nearby spacetime context~\cite{p211112417,p231206662}. Interleaved partitions provide another way to extend coverage beyond a local window. Open-Sora Plan alternates regularly spaced tokens and token groups, while OSP-Next applies complementary tokenwise and groupwise partitions along the spatial axes~\cite{p241200131,p260528691}. MSC further distinguishes local high-frequency detail from global low-frequency information. Its structural templates combine finite spatiotemporal windows in a high-resolution branch with spatially broader, temporally strided attention in a low-resolution branch~\cite{p241209828}. These methods translate shared spatial and temporal priors into different structural templates within support specification. Their differences concern axial factorization, joint neighborhoods, interleaved partitions, and combinations of local and strided templates across resolution branches.

\begin{figure*}[!t]
  \centering
  \resizebox{\textwidth}{!}{\input{figures/sparse_attention_pattern_atlas.tex}}
  \caption{Representative sparse-attention support patterns. Rows index queries and columns index keys; light cells are skipped query--key pairs. Blue cells mark the primary selected support. Dark gray cells mark mandatory or structural connections, such as global anchor rows and columns, self-blocks, and the second axis of a factorized pattern. Orange cells mark auxiliary connections, such as random additions, secondary routes, and long-range or multiscale links.}
  \label{fig:sparse-support-patterns}
\end{figure*}
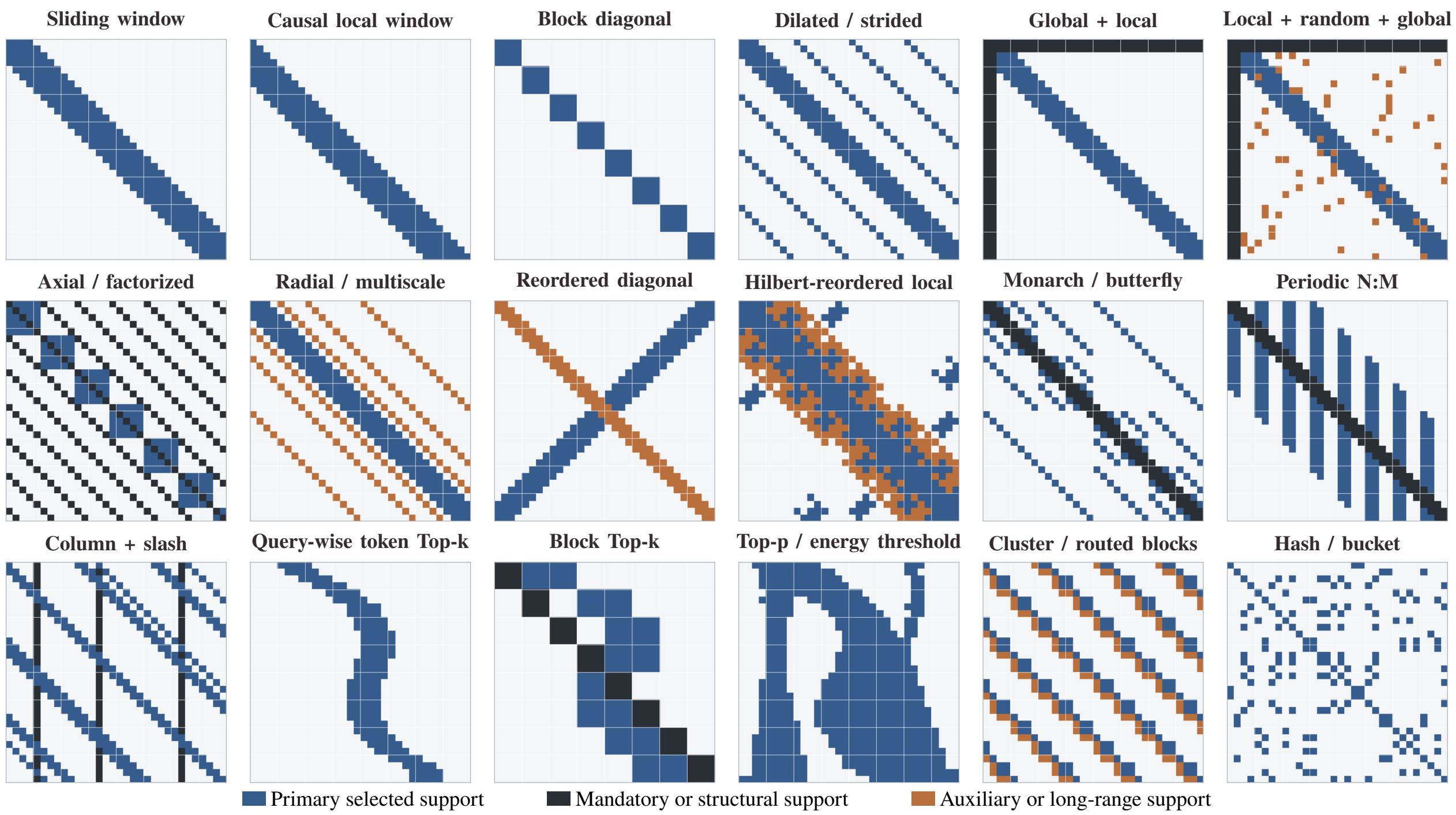

A further temporal prior is that attention relevance tends to decrease as frame separation grows. This source assumption motivates retaining distant context at lower density instead of imposing a strict temporal cutoff. Radial Attention implements it through spatial bands that narrow with temporal distance, while Sculpt4D increases the stride between retained blocks~\cite{p250619852,p260421592}. Both supplement the distance-dependent template with a mandatory first-frame anchor. LVSA uses a structural template in which a local temporal window preserves nearby frames and periodically spaced global frames preserve distant context~\cite{p260531057}. Its prescribed phase rotation broadens coverage across denoising steps, while window expansion compensates for boundaries and overlap with the global set. Frame-level granularity retains complete spatial support at selected times, in contrast to sparsifying within distant frame pairs. Radial Attention offers training-free inference and LoRA-based long-video adaptation, whereas LVSA applies its prescribed support without retraining. To encode temporal relevance decay, the structural templates within support specification differ in how distant context is thinned: distance-dependent band widths or block strides reduce spatial coverage within frame pairs, whereas windowed and periodic frame selection retains complete spatial support at selected times.

Specific generation tasks introduce source priors about which context must remain accessible and how it should be accessed. In reference-guided video and 4D generation, appearance consistency motivates persistent reference access. Tune-A-Video combines first-frame and previous-frame support, TalkingMachines retains initial, previous, and current chunks, and Helix4D pairs a local frame window with a first-frame anchor~\cite{p221211565,p250603099,p260526109}. Tune-A-Video adapts selected attention parameters, whereas TalkingMachines distills a bidirectional teacher into a sparse causal student. Avatar V extends the reference to a video and specifies asymmetric access, allowing generated tokens to read reference tokens while reference processing remains independent of the generated sequence~\cite{p260613872}. For incremental and streaming synthesis, causal availability identifies usable history, while spatial proximity or recency determines which part is retained. NUWA-Infinity follows generation order to retain neighboring generated patches, while Vorch-Streamer and the causal variant of JoyAI-Echo-1.5 combine a persistent prefix with recent chunks~\cite{p220709814,p260805663,p260823383}. Audio--video generation further requires temporally aligned context. MM-Diffusion combines local cross-modal windows with content-independent random shifts, while Hallo-Live accommodates anticipatory speech-driven motion through an asymmetric template exposing committed, synchronous, and provisional future audio~\cite{p221209478,p260423632}. Across these task-specific priors, structural templates within support specification differ in context extent and permitted access direction. Mandatory links preserve references or prefixes, while randomized selection broadens cross-modal window coverage.

\subsubsection{Calibrated Templates}
Calibrated templates use offline calibration traces as the source for support construction. These traces capture model behavior observed by running a generator on representative inputs before deployment. Evidence construction derives measures of importance, fidelity, or cost from these observations, and support specification uses the resulting evidence to determine reusable template configurations. At deployment, these configurations instantiate static support independently of runtime content. With calibration traces as their common source, methods differ in how evidence construction evaluates model behavior and how support specification translates that evidence into template geometry, parameter settings, or block membership.

Calibration traces share an offline origin, but their coverage depends on how source inputs are selected. LiteVAR obtains attention maps from a subset of its training data~\cite{p241117178}. PAROAttention uses the first two prompts in the CogVideo example set for its calibration experiments~\cite{p250616054}. ScalingAttention instead assembles prompts covering semantic dimensions such as camera motion and object interactions when constructing its sparse topology~\cite{p260623019}. These choices affect which model behaviors are represented in the source and distinguish collecting traces from an available sample set from explicitly seeking coverage across generation scenarios.

Statistical scoring can express how well candidate support preserves dense computation. DiTFastAttn measures relative absolute error at the model output when evaluating compression choices that include window attention with residual sharing. Efficient-vDiT measures mean squared error in the final hidden states after sparsifying a layer~\cite{p240608552,p250206155}. Both evaluate the discrepancy after subsequent model layers have processed the modified attention output. Sliding Tile Attention and Sparse-vDiT use the same error-scoring principle, measuring mean squared error directly between dense and sparse attention-head outputs~\cite{p250204507,p250603065}. DiTFastAttnV2 uses relative squared error at the head output~\cite{p250322796}. These measurements differ in the tensor being compared and the error metric. Statistical scoring also supplies selection-frequency evidence in CalibAtt, where averaging preliminary per-prompt masks yields block-retention frequencies for the final mask decision~\cite{p260305503}.

Pooling and sampling can be composed with statistical scoring to connect calibration traces to support decisions: pooled attention mass provides evidence for regional coverage or block membership, while sampled evaluations provide error or fidelity evidence for candidate masks. LiteVAR pools attention probabilities within candidate windows to estimate retained-mass ratios, whereas Compact Attention aggregates regional mass to measure the recall lost when a window boundary contracts~\cite{p241117178,p250812969}. PAROAttention forms block-sum evidence after token reordering, and CalibAtt and ScalingAttention aggregate attention scores over query--key blocks~\cite{p250616054,p260305503,p260623019}. For sampled evaluation, ScalingAttention selects query rows within strata to estimate how block-mask density affects fidelity relative to dense attention~\cite{p260623019}. Efficient-vDiT samples denoising timesteps and uses the largest observed output error to assess a candidate mask across the sampled generation states~\cite{p250206155}.

Template search within support specification determines support through a choice among candidate geometries and parameter settings. LiteVAR uses pooled attention-mass evidence to expand multidiagonal windows until a target retention ratio is reached, with mandatory support covering the first three scale segments~\cite{p241117178}. Compact Attention combines regional pooling with statistical scoring of the recall lost per unit of computation removed. This criterion guides the contraction of complementary spatial windows within frame-distance groups, subject to recall and contraction-cost thresholds, after which configurations are united across calibration prompts~\cite{p250812969}. Sliding Tile Attention selects among candidate three-dimensional window extents by minimizing averaged head-output error. SALAD's static ST-SWA configuration selects the smallest head-specific window satisfying a relative-error tolerance after arranging spatial or temporal neighbors contiguously~\cite{p250204507,p260116515}. Efficient-vDiT combines timestep sampling with final-hidden-state error scoring to search reference-frame budgets under an error tolerance. The selected template retains within-frame support and mandatory access to the designated global frames~\cite{p250206155}. Sparse-vDiT compares diagonal, multidiagonal, and stripe templates alongside head skipping using head-output error plus a retained-density penalty, keeping dense attention when no compressed mode meets its threshold~\cite{p250603065}. DiTFastAttnV2 couples head-relative-error evidence with latency estimates in integer optimization to assign head configurations jointly within its compression plan. Its arrow templates vary the visual window while preserving every text-related interaction as mandatory support~\cite{p250322796}. Across these methods, template search changes geometric extent, pattern family, or the joint assignment of configurations, while mandatory links constrain which context remains accessible.

Ranked selection and threshold rules select candidate units at the chosen granularity to form the calibrated support. Ranked selection uses top-$k$ or top-$p$ to retain candidates under a count or cumulative-mass budget. PAROAttention applies a threshold to pooled block attention sums in its calibrated tensor layout~\cite{p250616054}. CalibAtt first applies top-$p$ selection to pooled block scores so that the retained blocks meet an attention-mass target for each prompt. Statistical scoring then converts these preliminary masks into cross-prompt retention frequencies, and a threshold on those frequencies determines the deployed mask~\cite{p260305503}. ScalingAttention uses pooled block scores for top-$k$ selection at different support densities and unions the masks across prompts, recording the minimum density at which each block enters the union~\cite{p260623019}. An offline density cutoff is chosen from the sampled fidelity--density profiles and applied to the entry-density map to instantiate the mask. Ranking and thresholding directly select candidates that satisfy prescribed criteria, giving support specification flexibility in support membership beyond adjustments to a template's geometric extent.

Offline calibration can also configure a dynamic selector without fixing its eventual support. Such methods belong to the content-dependent regime: calibration establishes budgets, thresholds, or granularity parameters, while runtime evidence still determines retained connections. SVOO derives layer/head density budgets from the density required to retain a target attention mass across calibration inputs, while Focused Forcing uses head-ablation sensitivity to allocate frame budgets~\cite{p260318636,p260518346}. PASA measures changes in denoising velocity across calibration trajectories and uses their averages to distribute the attention budget across timesteps~\cite{p260412219}. For threshold calibration, HEART uses spectrally weighted denoising-velocity error to assign head-specific top-$p$ thresholds under a global sparsity budget, whereas VecAttention uses dynamic programming over offline sparsity--performance profiles to assign filtering ratios~\cite{p260514513,p260329494}. LiteAttention searches timestep-specific skip thresholds under relative-output-error bounds~\cite{p251111062}. Calibration also shapes granularity. Veda compares candidate three-dimensional tile shapes of fixed volume and selects the configuration minimizing attention-output error~\cite{p260530325}. DynamicRad performs offline Bayesian optimization over sparsity configurations using proxy attention error and a sparsity constraint~\cite{p260420470}. These uses of calibration determine rule configurations within support specification or grouping parameters associated with granularity.

\input{sections/05_qualitative_image}

\subsection{Dynamic Sparse Attention Algorithms}
\label{sec:review-content}
\label{sec:review-dynamic-algorithms}
Dynamic sparse attention uses support that depends on runtime content. Such support can adapt to semantic relationships, spatial correspondences, or historical dependencies that vary across samples and generation states. Its source may include current state, such as $Q/K/V$ features and attention statistics, history from earlier computations, or conditioning inputs such as text prompts, reference images, and camera poses.

Given source and granularity, dynamic methods differ primarily in how evidence construction derives decision evidence and how support specification uses it to determine support. For score-based selection, obtaining exact attention scores would require the dense computation being avoided. Evidence construction therefore estimates attention behavior at a fraction of this cost; geometric constraints and content-dependent group assignments can also provide decision evidence. Support specification converts scores, error estimates, or assignments into retained connections through ranking, thresholding, membership rules, or template search. Content-dependent evidence can also configure selection by adjusting budgets or thresholds, or by choosing a strategy or template configuration. The estimator that produces this evidence and the policy that configures selection can be explicit computations or learned modules. Training may also adapt the generator to the resulting sparse computation, either separately or jointly with these modules.

\subsubsection{Source}

Dynamic support requires information about the current sample, supplied by current state, generation history, or conditioning. Current $Q/K$ features are available before attention evaluation and provide a common source for adapting support to token affinities~\cite{p250514708,p250207590,p250518875}. Other current-state statistics become available only within the evaluation: SpargeAttention reads running softmax maxima, while DynSparse additionally uses accumulated normalization statistics to assess newly visited tiles~\cite{p250218137,p260616317}. History supplies information from earlier attention evaluations or generation steps. In autoregressive generation, ADSA uses previously generated value features to identify redundant context, whereas Focused Forcing combines current queries with historical keys to assess earlier frames~\cite{p250618226,p260518346}. Externally supplied conditioning can provide semantic or geometric evidence, such as poses in co-speech generation or camera and action states in world generation~\cite{p251002617,p260522718}. These sources differ in their availability and refresh frequency: supplied conditioning and stored history can guide a selection before attention starts, whereas online softmax statistics can only refine work that has already begun.

\subsubsection{Evidence Construction}
\label{sec:review-dynamic-evidence}
Pooling constructs regional relevance evidence by replacing nearby features with summaries, so support can be evaluated without first computing every token pair. DraftAttention averages $Q$ and $K$ within regions and computes a coarse attention map for subsequent ranking~\cite{p250514708}. LongCat-Video forms these summaries over three-dimensional video blocks, while ElasticDiT uses two-dimensional image blocks and shifts their boundaries across layers~\cite{p251022200,p260515684}. The reliability of this approximation depends on what a summary conceals. SpargeAttention supplements pooled scores with within-block cosine similarity to identify groups for which a mean is unreliable~\cite{p250218137}. DFSAttn preserves more internal variation by pooling smaller sub-blocks and aggregating their attention scores into the evidence used for block selection~\cite{p260523445}. Block Approximate Sparse Attention instead groups tokens by feature norm and supplements mean-based logits with a correction derived from diagonal second moments~\cite{p260519726}. These refinements separate the granularity of estimation from that of retention: richer evidence need not require equally fine retained units. The two sides of the interaction can also use different granularities: FG-Attn permits different query and key group sizes, while VecAttention pools queries but scores individual keys~\cite{p250916518,p260329494}. Pooled scoring is also used during generator training. VSA jointly learns the generator and the combination of coarse and sparse outputs around mean-pooled scores, while MotionCraft uses soft top-$k$ to propagate training gradients through pooled-score selection~\cite{p250513389,p260808553}. SpargeAttention2 retains pooled scoring and fine-tunes the sparse generator by minimizing velocity-prediction error against a frozen dense teacher~\cite{p260213515}.

\input{sections/05_qualitative_video}

Sampling estimates attention behavior from a subset of query--key observations instead of replacing every region by a summary. To score candidate blocks, XAttention samples strided antidiagonals that intersect vertical and diagonal concentrations~\cite{p250316428}. BLADE and the video-generation variant of PSA use maxima over sampled interactions as block-importance evidence~\cite{p250810774,p251204025}. A sampled maximum can expose a dominant interaction concealed by a mean-based summary, but only if the sample reaches it; sampling geometry therefore matters as well as sample count. Sampling can alternatively evaluate a complete candidate structure. Sparse VideoGen compares dense and candidate sparse outputs on sampled query rows using mean squared error, whereas RainFusion estimates the attention mass retained by its candidate patterns~\cite{p250201776,p250521036}. Distributed Sparse Attention evaluates template suitability on local subsequences before aggregating local preferences by majority vote~\cite{pdsaiclr2026}. These methods trade the coverage of their observations against the cost of deciding which structure to retain.

Statistical scoring can use quantities maintained by attention itself, or an observed attention distribution, to estimate contributions without a separate proxy-score computation. SpargeAttention uses the gap between a tile's maximum logit and the running softmax maximum to identify potentially negligible attention weights~\cite{p250218137}. DynSparse also accounts for normalization, comparing the tile maximum with accumulated log-sum-exp to bound the normalized weights of its entries~\cite{p260616317}. At the query level, Token Radius Attention uses the entropy of the attention distribution to estimate the number of keys needed to retain a target attention mass~\cite{p260802504}.

Whereas positional pooling and sampled block scoring start from predefined candidate units, hashing can define units from content. It constructs discrete similarity evidence without comparing every pair. Reformer uses random projections to obtain angular locality-sensitive hash codes, placing similar representations into shared buckets~\cite{p200104451}. Multiple hash rounds provide alternative groupings when a single boundary separates relevant tokens. FAST-AR uses locality-sensitive hashing to assign current visual queries and prompt keys to semantic buckets~\cite{p260201801}. Here, the output of evidence construction is a bucket assignment. Its use differs from a scalar relevance estimate: support specification can test shared membership directly, without first ordering all candidates by score.

Clustering exploits similarity among tokens to construct compact evidence for support decisions. Routing Transformer uses spherical $k$-means with shared query--key centroids updated during training~\cite{p200305997}. Sparse VideoGen2 instead clusters queries and keys separately and estimates attention between their centroids, accounting for key-cluster populations when estimating attention mass~\cite{p250518875}. The grouping must balance attention affinity, information lost within each group, and construction cost. A suitable key grouping depends on how queries attend to those keys, which independent clustering of $Q$ and $K$ does not capture. SVOO therefore jointly refines the two partitions, alternately grouping keys by their affinities to query centroids and queries by their affinities to key centroids~\cite{p260318636}. AdaCluster tailors the clustering metric to the roles of queries and keys. It groups queries by direction, while key clustering preserves both direction and magnitude and reclusters outliers to obtain more compact groups~\cite{p260418348}. To retain distinctions among individual keys, SCOPE clusters post-RoPE key features separately in their temporal, height, and width subspaces. For each query cluster, each key's assignments index subspace-centroid scores whose sum approximates that key's logit~\cite{p260812780}. To reduce clustering overhead, HyperVAttention restricts each token's centroid search to a local spatiotemporal neighborhood~\cite{p260703012}.

Projection shapes the space in which similarity is measured and can reduce scoring cost. Its directions may be content-independent, derived from current attention responses, or learned for a scoring objective. Reformer and FAST-AR compute hash codes from randomly projected feature vectors~\cite{p200104451,p260201801}. ClusterAttention derives projection directions from current features and recursively partitions queries and keys into prescribed group sizes along these directions. Its diagonalized implementation reuses a principal-component basis to avoid estimating a new projection at every split~\cite{p260826965}. To make projected distances reflect attention responses, SparsePR derives low-dimensional key representations from the second-moment matrix of sampled queries, then constructs query representations using the resulting key-group centroids. Clustering these projected features accounts for how tokens respond to the opposite feature set~\cite{p260818484}. Projections can also be learned for the scoring objective. DSV trains low-dimensional query and key projections against the original logits, encouraging their products to preserve the relative magnitudes of attention scores~\cite{p250207590}. SLA2 applies dimension-preserving linear projections to pooled $Q/K$ features before block scoring and trains these projections through a soft top-$k$ relaxation to improve its sparse--linear attention decomposition~\cite{p260212675}.

Scoring networks predict scores for support specification from input features and can be trained separately or jointly with the generator. Veda uses head-specific MLPs to score tiles from their mean, maximum, and minimum features. It first trains these scorers against dense-attention tile distributions with the generator frozen, then jointly fine-tunes the scorers and generator~\cite{p260530325}. Direct3D-S2 jointly learns sparse 3D convolutions and the generator, using attention to pooled convolutional descriptors to score spatial blocks~\cite{p250517412}. Learned scores can also describe ordering or group membership. Sparse Sinkhorn Attention trains a block-ordering network jointly with the model through differentiable Sinkhorn normalization~\cite{p200211296}. MoGA learns token-to-group scores alongside the generator, with a balancing loss that discourages assigning most tokens to only a few groups~\cite{p251018692}. DynamicRad separately trains an MLP on pooled text embeddings to predict motion intensity under optical-flow supervision. The predicted intensity determines a motion regime used to retrieve an offline-optimized sparsity configuration~\cite{p260420470}. Together with the pooled and projected estimators above, these methods expose distinct training choices: adapting the generator around a fixed estimator, fitting an estimator to dense-attention targets, or optimizing both jointly. The resulting evidence may be a scalar score, a contribution bound, a group assignment, or a template-level prediction; support specification determines how each form becomes retained connectivity.

\subsubsection{Support Specification}
\label{sec:review-dynamic-specification}
Ranked selection converts relevance evidence into support by ordering candidates and retaining a prefix under a count or cumulative-mass budget. Top-$k$ controls the number of retained tokens or groups~\cite{p250207590,p250514708,p250618226,p260518346,p250821058}, whereas top-$p$ selects enough candidates to reach a target attention mass~\cite{p250316428,p250916518,p250518875}. These criteria can be combined to preserve both a minimum amount of context and sufficient estimated mass. Jenga first retains a prescribed fraction of the highest-scoring KV blocks, then adds further blocks in rank order if their accumulated probability remains below the target~\cite{p250516864}. SpargeAttention2 takes the union of top-$k$ and top-$p$ masks, allowing the mass criterion to expand support for diffuse distributions while the count criterion enforces a minimum retained count when mass is concentrated on a few blocks~\cite{p260213515}. The budget parameters can themselves adapt to current evidence. BSA derives its cumulative-score target from score statistics, while SCOPE averages initial retained counts with weights proportional to query-cluster sizes, then uses this head-specific average as a minimum count to expand smaller selections~\cite{p250901085,p260812780}. USV uses attention entropy to distribute sparsity across layers, assigning more aggressive pruning to layers with concentrated attention~\cite{p251205754}. Beyond budget choice, methods differ in which candidates compete for retention. VMoBA ranks query--key-block pairs jointly within each head, allowing different queries to receive different amounts of support under a shared budget~\cite{p250623858}. Hierarchical selection instead restricts the candidate set before finer ranking. Log-linear Sparse Attention recursively selects coarse groups and applies top-$k$ to their descendants, while Light Forcing first retrieves relevant frames and then selects blocks within those frames~\cite{p251216615,p260204789}.

Threshold rules convert score or contribution estimates into retention decisions by comparing each candidate with a cutoff. The cutoff provides an acceptance criterion without requiring an explicit ordering of all candidates. VecAttention retains keys whose estimated logits lie within a prescribed margin of the row maximum~\cite{p260329494}. Sol-Attn instead places the cutoff a specified number of standard deviations above the mean block score, adapting the raw-score criterion to each query's score distribution~\cite{p260724027}. Thresholding can additionally determine whether an already-scored tile participates in value aggregation. SpargeAttention skips this aggregation when the tile maximum falls sufficiently below the running softmax maximum, while DynSparse uses accumulated normalization statistics to skip tiles whose estimated normalized weights fall below a cutoff~\cite{p250218137,p260616317}. A threshold can also refine a candidate set produced by another selection rule. DynaST propagates top-ranked matches across scales and neighboring positions, then retains only candidates with positive learned match scores~\cite{p220706124}.

Randomized selection broadens support coverage by varying the candidates considered or their order of retention. High-Resolution Sparse Attention initializes reference-image matches randomly, proposes additional candidates around current matches, and propagates candidates from neighboring queries. Each iteration retains the highest-scoring match from this search pool and adds it to the query's support, combining random exploration with score-based refinement~\cite{p201207288}. PASA introduces randomness at the ranking stage by adding a random bias to block scores before top-$k$ selection. Resampling the bias across layers and denoising steps gives moderately scored regions opportunities to receive exact attention, reducing their persistent exclusion from the selected support~\cite{p260412219}. Randomness thus acts on candidate discovery in the former and on ranking in the latter, while relevance evidence continues to guide retention.

\input{tables/framework_examples_algorithm}

When evidence takes the form of group assignments rather than scalar scores, membership rules determine attention access through shared membership. Reformer allows attention within hash buckets and combines support across multiple hashing rounds, reducing the chance that similar tokens remain separated by a particular partition~\cite{p200104451}. Routing Transformer connects queries and keys assigned to the same centroid, selecting a prescribed number of the nearest tokens per centroid to constrain group size~\cite{p200305997}. MoGA assigns each token to its highest-scoring learned group and performs full attention among that group's members~\cite{p251018692}. Group size is itself a support-level concern: Routing Transformer constrains it by selection, whereas MoGA encourages balanced assignments through its router training loss. FAST-AR applies membership at the frame level to specify cross-attention support. A prompt key is retained if it shares a bucket with at least one current-frame query. All queries in that frame then attend to the retained prompt KV subset, so membership determines which key columns remain accessible across the frame~\cite{p260201801}.

Support specification can exploit recurring attention geometry by adapting structural templates or choosing among candidate configurations through template search. Sparse VideoGen selects the spatial or temporal candidate with the lower estimated attention-output error, assigning each head a corresponding within-frame or cross-frame pattern~\cite{p250201776}. RainFusion tests whether a temporal pattern preserves sufficient estimated attention mass, then tests a spatial pattern if needed. When neither passes, it uses a checkerboard template that distributes retained KV tokens across the spatial grid~\cite{p250521036}. HeadCast compares restricted-context outputs with full-history outputs after warm-up and assigns heads to initial-sink, recent-block, or spatially local templates when their fidelity criteria are met. Remaining heads retain unrestricted spatial access within the temporal window, and the assignments govern subsequent attention~\cite{p260720125}. A method can also adapt the parameters of a prescribed geometry. Token Radius Attention estimates per-query token budgets from the current sample's attention entropy. It selects the smallest base radius that accommodates each budget and reduces the radius with temporal separation. This yields content-dependent support within a predefined neighborhood template~\cite{p260802504}. Dynamic configuration selection can operate at the prompt level. DynamicRad prepares candidate configurations offline and selects one at inference according to the motion regime predicted from the current prompt. The selected configuration sets the temporal decay, long-range allocation, and block-retention thresholds used to construct the support~\cite{p260420470}. A fixed template vocabulary can retain regular execution shapes despite content-dependent selection. Whether evidence is collected at each evaluation, during warm-up, or once per prompt determines how often its construction cost is paid.

Mandatory links constrain support specification to preserve designated context independently of its relevance ranking. The final support combines this required subset with content-dependent selections, allowing task requirements to determine which context remains accessible. Jenga and Light Interaction preserve access to text tokens while sparsifying visual context, maintaining semantic conditioning under visual pruning~\cite{p250516864,p260531158}. Temporal anchors impose different access requirements. RainFusion2.0 retains both all queries' access to first-frame keys and first-frame queries' access to all keys, creating bidirectional support around the initial frame~\cite{p251224086}. ViewRoPE instead always includes the current frame alongside selected historical frame blocks, preserving complete within-frame context~\cite{p260207854}. For longer streams, Wonder, WorldKV, ReWorld, and Sparse Forcing combine mandatory initial sink context with selected historical support, preserving an initial reference as the remaining context changes~\cite{p260726037,p260522718,p260823565,p260421221}. The mandatory subset can also vary across generation stages. In SparSTAR, it includes both text and image-reference blocks for the first clip, whereas only text blocks remain mandatory for subsequent clips~\cite{p260810519}. These rules compose: rankings can be augmented with mandatory anchors, propagation can be filtered by thresholds, and template choice can be combined with parameter adaptation. Their combinations determine both retained information and the structure presented to the execution system.

\subsection{Auxiliary Algorithmic Mechanisms}
\label{sec:review-auxiliary}
Reuse and correction complement both support regimes. Reuse avoids repeated evidence construction, support selection, or attention evaluation by carrying information across uses. Correction compensates for omitted value contributions and altered normalization.

\subsubsection{Reuse}
\label{sec:review-reuse}
Reusing support carries an existing mask into later attention evaluations, reducing repeated construction while outputs are computed from current tensors. Token Radius Attention reuses the radius masks obtained during dense warm-up throughout subsequent sparse steps, while LiteAttention retains tile-skip decisions in a persistent mask and reuses them to bypass previously excluded tiles as new skips are added~\cite{p260802504,p251111062}. The duration of mask reuse can be fixed or state-dependent. DFSAttn reuses each mask between periodic refreshes, whereas HEART retains the mask from its latest refresh until pooled $Q/K$ drift exceeds a threshold and triggers reconstruction~\cite{p260523445,p260514513}. Chorus II extends reuse across requests by caching historical image-to-video masks and applying a retrieved mask directly when the new request has sufficiently similar text and image conditioning~\cite{p260625040}.

Reusing construction intermediates reduces repeated estimation by retaining results for subsequent evidence construction or support decisions. RAPID retains warm-up block scores and reuses them for a later pruning step, constructing a sparser mask without rescoring the blocks~\cite{prapidcvpr2026}. MOD-DiT reuses stored attention-map entries at currently omitted positions and combines them with newly observed entries to update the map used for pattern prediction~\cite{p260111641}. AdaSpa reuses cached log-sum-exp values to normalize current scores during later block searches, avoiding a separate normalization pass~\cite{p250221079}. Reuse can also retain the group representations on which evidence is computed. Sparse VideoGen2 and AdaCluster reuse preceding-step centroids as the initialization for current clustering~\cite{p250518875,p260418348}. SVOO reuses the query--key group assignments between scheduled co-clustering runs, bypassing partition construction on intervening steps~\cite{p260318636}.

Output reuse avoids repeated attention computation by reusing or extrapolating previously computed outputs. LiteVAR reuses conditional attention outputs in the unconditional classifier-free-guidance branch, combining this sharing with multidiagonal window attention~\cite{p241117178}. For reuse across denoising steps, FlashOmni caches query-block outputs and reuses or forecasts them for selected queries. A pooled attention map jointly coordinates this output reuse and KV-block skipping for the recomputed queries~\cite{p250925401}. DiSC reuses preceding-step attention outputs for tokens with small latent changes and recomputes the remaining query rows using softmax-threshold sparse attention~\cite{p260525798}.

\subsubsection{Correction}
\label{sec:review-correction}
Restricting support removes value contributions from omitted keys and renormalizes retained weights over a smaller key set. Output correction estimates the resulting dense--sparse discrepancy and adjusts the computed sparse output. DiTFastAttn caches the difference between dense and window outputs at a reference step. Reusing this residual corrects subsequent window outputs~\cite{p240608552}. SparVAR upsamples a dense--sparse residual measured at an earlier resolution and adds it to finer-resolution sparse outputs~\cite{p260204361}. QuantSparse corrects quantized sparse outputs using a cached residual plus an estimate of its inter-step change, filtering that change through dominant singular components~\cite{p250923681}. SparsePR evaluates dense attention for a few current queries and fits an affine residual predictor with a low-rank linear term to their dense--sparse differences. Its predictions correct the remaining sparse outputs~\cite{p260818484}. Retained-weight normalization can also be corrected through rescaling. Re-ttention uses a cached sparse-to-dense softmax-denominator ratio to rescale retained weights~\cite{p250522918}. Rectified SpaAttn estimates retained attention mass from pooled full-context weights and uses it to rescale weights inside the sparse support~\cite{p251119835}.

Contribution recovery seeks to restore information from positions outside the exact support, thereby reducing the quality degradation introduced by sparsification. Methods differ in how they represent this omitted context and how they combine the resulting approximation with exact sparse attention. Re-ttention represents the omitted region by the residual between a dense reference output and its rescaled sparse counterpart, then adds this residual to later sparse outputs~\cite{p250522918}. SLA applies exact attention to critical blocks and linear attention to marginal blocks outside the exact support, adding a learned projection of the latter to the sparse output~\cite{p250924006}. Regional methods instead summarize the omitted positions with representative keys and values. SVG-EAR ranks query--key blocks by estimated centroid-compensation error per unit cost and computes the highest-ranked blocks exactly, while using cluster-mean keys and values to compensate the rest~\cite{p260308982}. ClusterAttention selects exact key clusters through centroid-based top-$k$ or attention-mass routing, then uses size-weighted cluster means to compensate the unselected clusters~\cite{p260826965}. Taylor-based recovery estimates the contribution of omitted blocks through local expansions. PISA and PASA use Taylor expansions to estimate the weighted-value and normalization terms of omitted blocks, then combine them with the exact terms of selected blocks before normalization~\cite{p260201077,p260412219}.

Dense fallback restores exact attention for queries or heads whose sparse or approximate alternatives fail an error or fidelity criterion. LIVEditor routes queries with high estimated zeroth-order approximation error to exact attention over their candidate KV set~\cite{p260504569}. Sparse-vDiT retains dense attention for a head when every compressed mode exceeds its calibrated loss threshold~\cite{p250603065}. HeadCast retains the baseline sliding-window attention for heads whose more restricted candidates fail its fidelity tests~\cite{p260720125}.

\subsection{Algorithm--System Co-Design}
\label{sec:review-system-codesign}
\label{sec:review-system}

Logical sparsity does not directly reduce hardware work: fragmented connections, partially occupied tiles, and support-construction overhead can absorb the savings. Static support is known before an attention call, so metadata, layouts, kernels, and batches can be prepared in advance. Dynamic support changes with the input, so its construction and the reorganization of the resulting work become part of the runtime workload. Both regimes must fit retained interactions to hardware tiles, retrieve selected operands efficiently, and keep metadata and correction state available. We review the three cases in turn.

\subsubsection{Static Support}
\label{sec:review-static-codesign}
Predetermined connectivity allows systems to move preparation out of repeated attention calls. The relevant opportunity is the availability of the pattern and its configuration sequence, including changes prescribed across layers, heads, or timesteps.

\paragraph{Connectivity Encoding}
Known sparse masks may still occupy substantial memory and require expensive traversal. Their structure can be inspected before inference to choose compact encodings and allocate metadata for the retained connections. CalibAtt exploits contiguous selected KV ranges through interval encoding and trims each layer's allocation to its maximum required row width~\cite{p260305503}. ScalingAttention uses bitmap encoding: aligned row bitmaps pack block membership, while bit-scan and bit-clear operations enumerate active blocks~\cite{p260623019}. Intervals exploit long runs, whereas bitmaps represent fragmented selections compactly. CalibAtt further merges nearby intervals to reduce storage, accepting computation of intervening blocks. Thus, advance preparation supports both metadata compression and a measured tradeoff between representation size and executed support.

\paragraph{Layout}
Predetermined sparse neighborhoods can be scattered in the original token order, preventing regular block computation. Because their positional mappings are known, permutation can be prepared and reused rather than rediscovered from each input. Sparse Transformer transposes strided connectivity into local block computation~\cite{p190410509}. HilbertA maintains a Hilbert-ordered sequence across layers, applies scheduled shifts through pointer offsets, and limits explicit reordering to initial permutation and final restoration~\cite{p250926538}. These choices reduce the recurring movement required to place sparse operands in an executable layout.

\paragraph{Computation}
When support is fixed before inference, its computation can be expressed through operators and kernels matched to the selected pattern. Sparse-vDiT replaces dense attention with mode-specific computation~\cite{p250603065}. Its diagonal pattern admits local-window evaluation; rearranging the multi-diagonal pattern exposes a similar window form; and vertical-stripe attention restricts score computation to selected key columns. The method prepares a sparse kernel for each mode, with a predefined fraction of attention blocks to compute. The resulting kernels compute the required block products rather than all query--key pairs.

\paragraph{Partitioning}
Sparse heads with the same pattern perform the same type of computation on different $Q/K/V$ tensors. Sparse-vDiT first assigns a sparse mode to each head through offline search, then groups same-mode heads within each layer for joint execution~\cite{p250603065}. The heads remain independent attention problems, but grouping makes their work available together to the corresponding pattern-specific kernel. This batching can improve GPU utilization by exposing parallel work across heads when individual sparse head computations are too small to fill the device.

\subsubsection{Dynamic Support}
\label{sec:review-dynamic-codesign}
Content-dependent support makes its preparation and execution structure part of the runtime workload. The system must repeatedly construct decision evidence, select connections, and update the organization of the resulting work.

\paragraph{Computation}
Kernel optimization reduces the computation and intermediate data movement required for online evidence construction and selection. Fusing dependent stages lets kernels reuse input data and consume intermediate results directly. For evidence construction, Light Interaction fuses query blocking, mean pooling, and block-major output into one preparation pass, and jointly prepares $K/V$ while computing pooled keys~\cite{p260531158}. HyperVAttention combines local-neighborhood processing, cluster assignment, and centroid accumulation in one clustering kernel~\cite{p260703012}. SVG-EAR optimizes its evidence-estimation kernel by expanding the squared compensation-error expression and accumulating the required statistics while logits are available, avoiding a separate routing-error pass over stored logits~\cite{p260308982}. To combine scoring with selection, DSV fuses low-rank matrix multiplication with incremental top-$k$: each score tile is merged into the query's retained candidates without storing a complete score matrix~\cite{p250207590}. Sol-Attn further fuses selection with attention evaluation: an outer loop scores a chunk of pooled keys and compares the scores with a precomputed threshold; an inner loop immediately loads the selected original KV blocks and updates online softmax~\cite{p260724027}. This chunk-local decision removes a separately stored global routing map and its later traversal.

\input{tables/framework_examples_system}

Mixed-precision computation can reduce the cost of the intermediate operations used to build dynamic support. Expensive evidence-estimation or selection stages can use lower precision, while attention evaluation retains its required numerical accuracy. For example, FAST-AR performs nearest-neighbor matching repeatedly across layers and generation steps to retrieve relevant keys. It supports low-bit quantized matching for this stage~\cite{p260201801}, using compact representations for the comparisons while attention evaluates the retrieved operands. Bit width trades matching cost against the accuracy of the selected support.

\paragraph{Layout}
Permutation and packing make operands contiguous when input-dependent sparsification leaves the selected data scattered in memory. Because the support can change between attention calls, the layout transformations are performed at runtime and updated with the new selections. SVG2 constructs separate permutations for query clusters and key clusters, applies the key permutation to both $K$ and $V$, and stores each cluster as a contiguous segment~\cite{p250518875}. MoGA places $Q/K/V$ in a common route-group order, concatenates the groups, and records cumulative lengths~\cite{p251018692}. LIVEditor gathers selected context KV blocks into compact tensors and reorders query blocks by sharpness before sending them to different computation branches~\cite{p260504569}. Inverse mappings or scatter restore the output order after evaluation.

\paragraph{Partitioning}
Query grouping improves reuse in dynamic attention by forming execution groups from the currently selected support. As support changes, the groups are updated to match the new workload. HyperVAttention evaluates query-cluster merges using both the hardware-rounded query extent and the union of selected key clusters~\cite{p260703012}.

\paragraph{Scheduling}
Cost-based dispatch and task migration adapt task assignments to the workload produced by the current support. Dynamic selection can change the distribution of work between attention calls, so these methods use updated workload information to revise assignments. FVAttn counts retained blocks after mask construction, exchanges head-load information, and chooses a bounded migration plan~\cite{p260716190}. Each rank sends and receives at most one head; its state moves by point-to-point transfer so that attention runs under revised ownership. DSV also assigns heads using sparse workloads in its hybrid context-parallel scheme~\cite{p250207590}. These assignments respond to realized computation costs rather than maintaining an equal head count despite unequal work.

\subsubsection{General Sparse Execution}
\label{sec:review-shared-sparse-constraints}
Sparse attention introduces indexing and data-management work alongside its reduced arithmetic. Compact metadata, effective task organization, and efficient data delivery are therefore needed to turn fewer interactions into faster execution. Across different support patterns, the underlying execution problems are shared.

\paragraph{Connectivity Encoding}
List encoding makes the retained connections directly available to sparse kernels, avoiding traversal of a dense connectivity matrix. BSA stores each query block's selected-KV count and block indices, which the kernel follows during evaluation~\cite{p250901085}. The list orientation can also match the direction of computation: key/value gradients require the queries connected to each key. Log-linear Sparse Attention starts from query-to-key top-$k$ indices, counts how many queries select each key, computes offsets by prefix sum, and writes per-key query lists~\cite{p251216615}. Its $dK/dV$ kernel then gathers these queries directly. The reverse lists are rebuilt when the selected connections change.

Implicit indexing avoids storing every connection when sparse neighborhoods can be generated from positional rules. Instead of reading an explicit list, the kernel computes the required operand locations from the current query coordinates. STA derives KV-tile locations from the query tile's window, while Generalized Neighborhood Attention converts multidimensional tile coordinates into one-dimensional load indices~\cite{p250204507,p250416922}.

\paragraph{Layout}
Stride remapping exposes sparse operands through a different tensor view, reducing the copying needed to prepare inputs for computation. CogView uses padding and modified tensor strides to expose block-window views in the existing buffer when the attended entries are already consecutive~\cite{p210513290}. Computation can then read the window operands without gathering a separate copy of every window.

Bank mapping organizes sparse operands across memory banks to reduce access conflicts during parallel computation. DiSC places a key column in the bank selected by its hash and assigns that column's nonzero score computations to the corresponding unit~\cite{p260525798}. It generates projection outputs in a prearranged order so that the following transpose places key columns and value rows in the required banks. This aligns operand locations with sparse task ownership.

\paragraph{Computation}
Operator optimization improves the computational form used to evaluate a sparse workload. Selecting or replacing operators can reduce redundant arithmetic and data movement introduced by an inefficient formulation. DiSC replaces dense score and aggregation products with elementwise sparse operators: sampled dense--dense multiplication computes selected $QK^\top$ entries, and sparse--dense multiplication aggregates their weighted values~\cite{p260525798}. Jenga optimizes the operators used in block selection by replacing \emph{einsum} contractions with cuBLAS-backed batched matrix multiplication and using vectorized indexing to avoid large explicit loops~\cite{p250516864}.

Kernel optimization improves how sparse computation is carried out on the device. Custom implementations can reduce unnecessary arithmetic, and fusion can avoid intermediate transfers between dependent stages. FlashOmni incorporates block-skip tests into the attention kernel: before loading a KV block, the kernel decodes its skip symbol and executes the $QK^\top$ and $PV$ products only for retained blocks~\cite{p250925401}. PISA addresses the additional passes introduced by correction with a fused kernel~\cite{p260201077}. It evaluates selected blocks exactly, scans block summaries for zeroth-order recovery, and adds the global first-order term within the same online-softmax computation. Exact and approximate contributions share normalization and accumulate directly into the output.

Mixed-precision computation lowers the cost of sparse execution by using lower precision in selected operations. Operand, multiplication, and accumulation precisions can be chosen separately to balance execution cost and numerical error. FPSAttention computes $Q/K$ scaling factors per sparse tile, uses FP8 operands, and returns BF16/FP16 outputs~\cite{p250604648}. PAROAttention applies INT8/INT4 computation to blockwise attention products and reorders tokens to make values within each quantization group more coherent~\cite{p250616054}. The sparse blocks determine the numerical groups processed by these implementations.

\paragraph{Partitioning}
Tiling reduces wasted computation from poorly aligned sparse regions by organizing work into hardware-efficient blocks. STA matches its sparse-attention tiles to the kernel's compute blocks, executing retained query--KV tile pairs and skipping excluded pairs~\cite{p250204507}. Generalized Neighborhood Attention uses simulated tile-access counts to choose tile shapes and neighborhood configurations~\cite{p250416922}. PSA separates logical pooling blocks from execution units, splitting large blocks or combining small ones into independently tuned hardware tiles~\cite{p251204025}.

Query grouping improves data reuse and device utilization by processing several queries against a shared set of KV operands. Generalized Neighborhood Attention uses a stride parameter to make queries within each group share the neighborhood of a designated query~\cite{p250416922}. GRAT groups contiguous queries and gives each group a common structured KV neighborhood, with group size matched to the GPU's execution configuration~\cite{p250514687}.

Batching combines compatible sparse subproblems to provide more parallel work when each subproblem is too small to use the device efficiently. SPADE groups heads sharing a blocking scheme into a single kernel launch, increasing the work available to that launch, and co-schedules heads with identical sparse patterns~\cite{p260803335}. Each head retains its own attention output while compatible heads execute together.

\paragraph{Scheduling}
Cost-based dispatch and persistent scheduling address uneven task sizes and idle workers during sparse evaluation. They organize the remaining work so that available workers can continue processing useful tasks. FG-Attn's persistent work scheduler estimates each query group's cost from its KV-list length and prioritizes longer tasks. Within its persistent kernel, each thread block takes the highest-cost remaining output tile from the GPU queue, processes it, and then requests another tile~\cite{p250916518}. This lets workers that finish short tasks continue processing the remaining workload.

\paragraph{Access}
Cooperative gather and coalesced access reduce the cost of retrieving the operands required by sparse computation. They coordinate address preparation and memory requests to limit access overhead. FG-Attn loads sparse indices cooperatively, distributes them across warps, and lets threads calculate addresses and gather different KV vectors into a contiguous shared-memory tile~\cite{p250916518}. PISA organizes pooled key/value summaries into consecutive groups and streams them through coalesced loads when computing omitted-block contributions~\cite{p260201077}. These access schemes coordinate irregular exact-operand retrieval and regular summary streaming, respectively.

Selective communication reduces data exchange by restricting transfers to operands required by sparse computation. DSA exploits spatially restricted support to exchange KV only with neighboring devices in a partial ring, while distributing attention sinks separately~\cite{pdsaiclr2026}. This limits the communication range to the devices supplying the required spatial neighborhood and preserves access to global sink tokens.

\paragraph{Pipelining}
Prefetching starts retrieval ahead of the computation that will consume the data, reducing stalls when future sparse operands are needed. ReWorld prefetches retrieved landmark KV from host memory on a separate stream while denoising is in progress, advancing data delivery before the landmarks are consumed by attention~\cite{p260823565}.

Asynchronous transfer decouples issuing a data movement operation from waiting for its completion, allowing independent work to proceed while data are in transit. Synchronization ensures the transferred operands are ready before consumption. Veda issues asynchronous TMA loads for selected, noncontiguous KV tiles into shared-memory buffers while attention matrix multiplication proceeds on tiles already available~\cite{p260530325}.

Double buffering separates storage for current and upcoming data, so loading new operands does not overwrite those still being consumed. The two buffers exchange roles as execution advances. PAROAttention loads the next configured sparse mask into one buffer while attention reads the current mask from the other, allowing mask loading and attention computation to proceed concurrently~\cite{p250616054}.

Warp specialization assigns data preparation and attention arithmetic to different warps within a thread block. Compute warps perform regular matrix operations while producer warps handle sparse indexing and data delivery. STA assigns mask calculation and KV transfer to producer warpgroups while consumer warpgroups evaluate attention on delivered tiles~\cite{p250204507}. PASA applies this division to correction: load warps prefetch grouped first-order statistics while compute warps execute attention~\cite{p260412219}.

\paragraph{Residency}
On-chip retention keeps reusable sparse metadata and intermediate state close to computation, reducing repeated preparation and temporary storage. FlashOmni decodes each block-skip byte once and retains the resulting decisions in registers for subsequent KV-block iterations~\cite{p250925401}. Reusing these decisions avoids repeating the same bitwise decoding at every block load.

Offloading moves infrequently used sparse state out of device memory, freeing capacity for active computation and allowing longer generation within a fixed GPU memory budget. ReWorld keeps its landmark bank in pinned host memory and stages only the retrieved working set on the GPU~\cite{p260823565}. Device memory therefore follows the fixed active-context budget rather than the full stored history, enabling long rollouts to retain access to older landmarks. Selective retrieval limits the amount transferred back, and prefetching overlaps these transfers with denoising to reduce the additional waiting caused by host retrieval.

%% file: figures/sparse_attention_pattern_atlas.tex

\definecolor{wine}{HTML}{5C142F}
\definecolor{navy}{HTML}{1E386B}
\definecolor{orange}{HTML}{A6300D}
\definecolor{ink}{HTML}{342B2F}
\definecolor{supportprimary}{HTML}{345D8C}
\definecolor{supportstructural}{HTML}{2A2F36}
\definecolor{supportauxiliary}{HTML}{B9703C}
\newcommand{\patternpanel}[2]{%
  \begin{minipage}[t]{2.82cm}
    \centering
    \parbox[c][4.8mm][c]{2.80cm}{%
      \centering\fontsize{7.0}{7.6}\selectfont\bfseries\color{ink}#1}%
    \\[-0.35mm]
    \includegraphics[width=2.66cm]{figures/support_pattern_panels/#2}%
  \end{minipage}%
}

\newcommand{\legenditem}[2]{%
  \textcolor{#1}{\rule{2.8mm}{1.7mm}}%
  \hspace{0.6mm}{\fontsize{7.2}{8}\selectfont #2}%
}

\begin{minipage}{17.55cm}
\centering
\setlength{\tabcolsep}{0.40mm}
\renewcommand{\arraystretch}{1}
\begin{tabular}{@{}cccccc@{}}
\patternpanel{Sliding window}{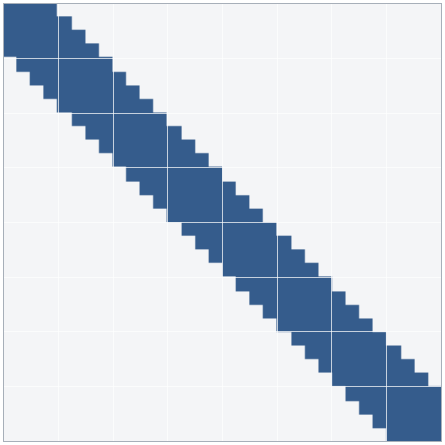}
& \patternpanel{Causal local window}{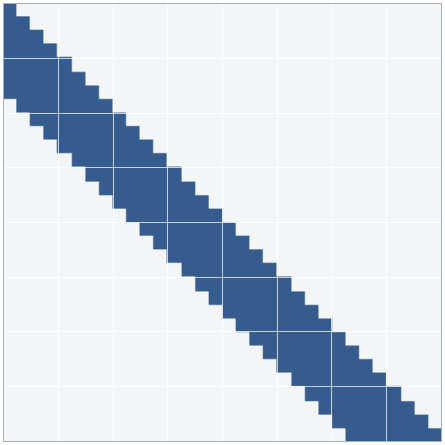}
& \patternpanel{Block diagonal}{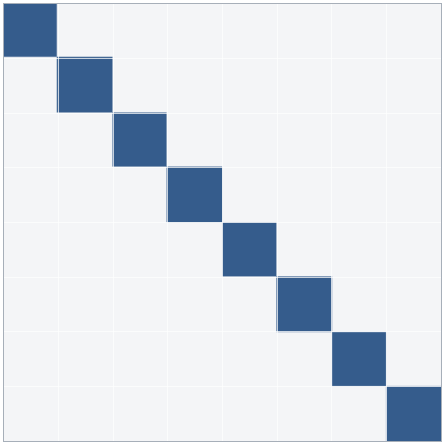}
& \patternpanel{Dilated / strided}{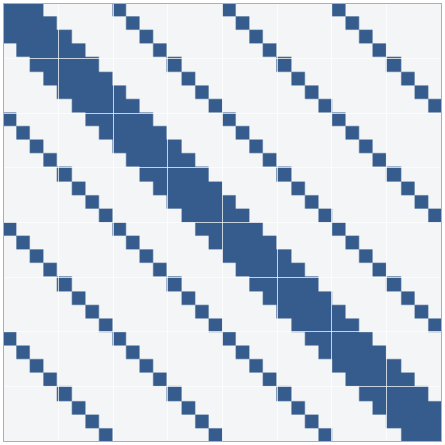}
& \patternpanel{Global + local}{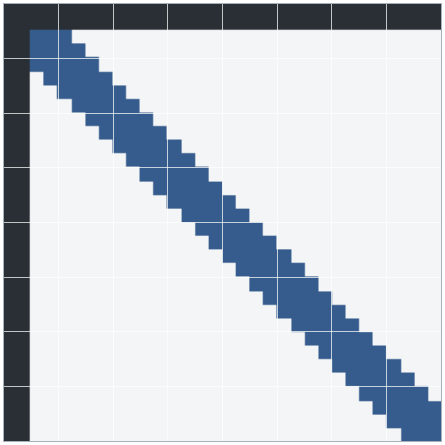}
& \patternpanel{Local + random + global}{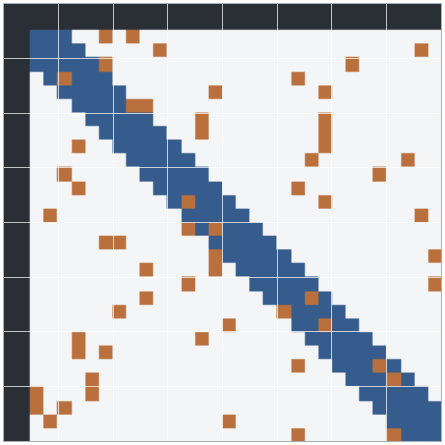}
\\[0.8mm]
\patternpanel{Axial / factorized}{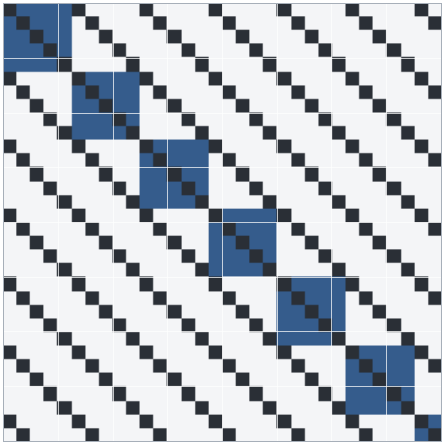}
& \patternpanel{Radial / multiscale}{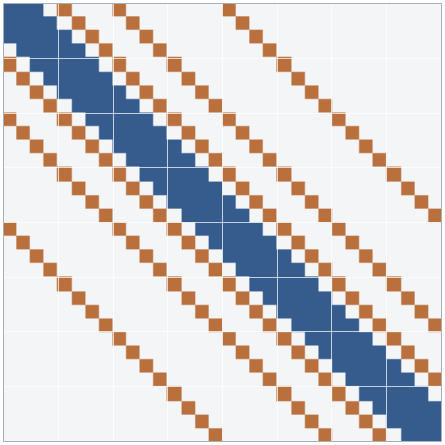}
& \patternpanel{Reordered diagonal}{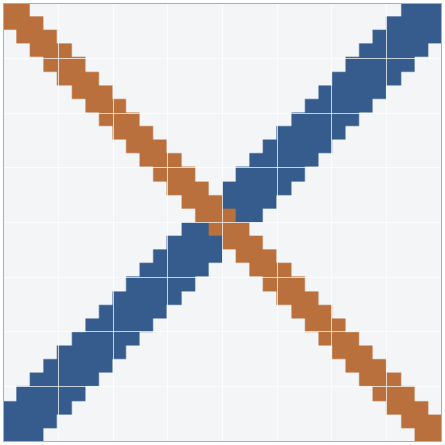}
& \patternpanel{Hilbert-reordered local}{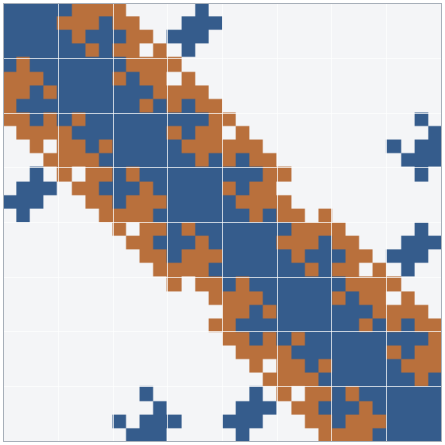}
& \patternpanel{Monarch / butterfly}{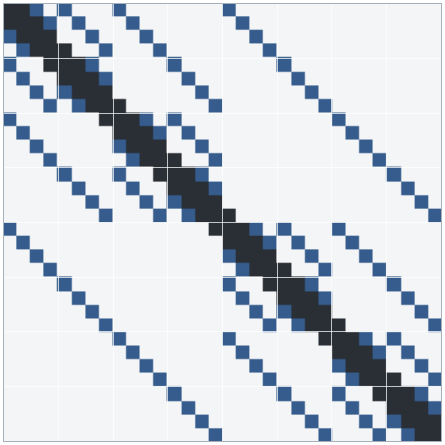}
& \patternpanel{Periodic N:M}{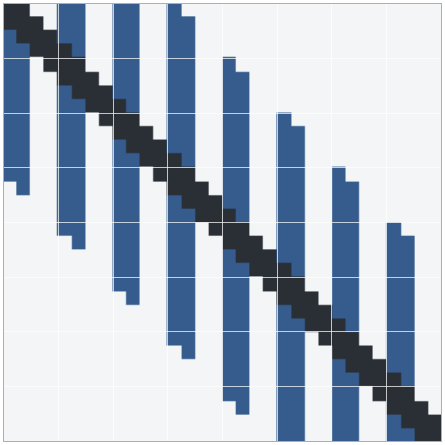}
\\[0.8mm]
\patternpanel{Column + slash}{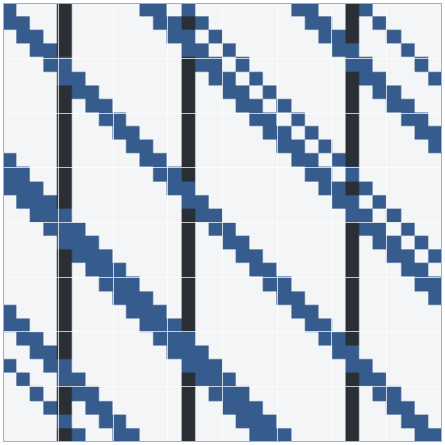}
& \patternpanel{Query-wise token Top-k}{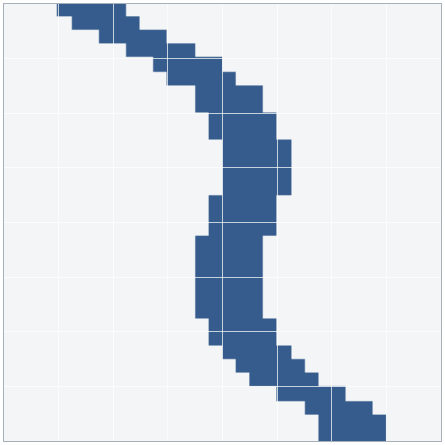}
& \patternpanel{Block Top-k}{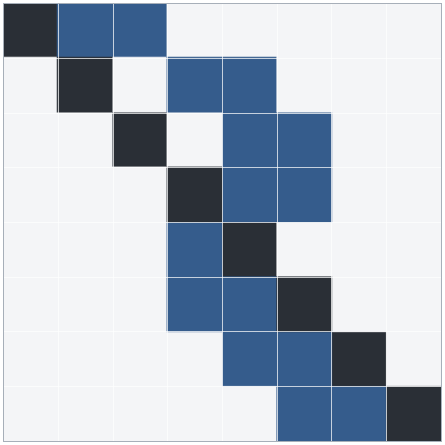}
& \patternpanel{Top-p / energy threshold}{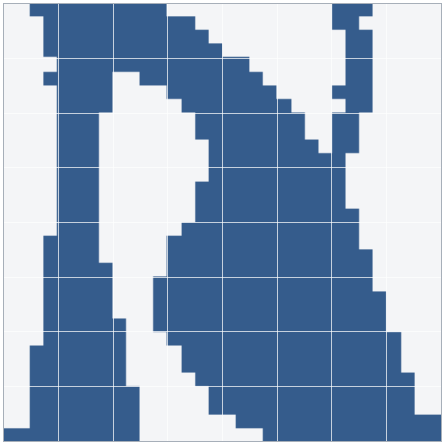}
& \patternpanel{Cluster / routed blocks}{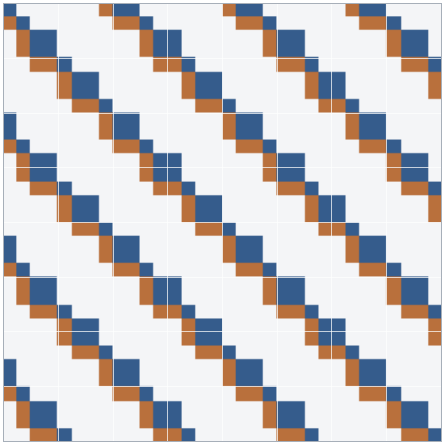}
& \patternpanel{Hash / bucket}{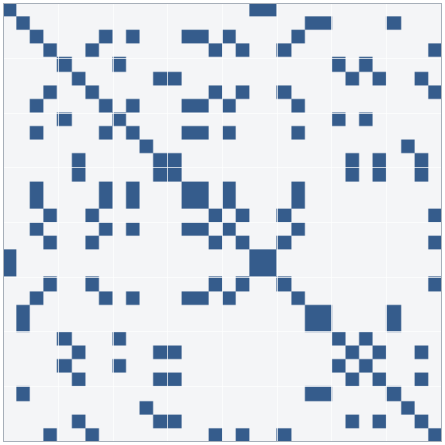}
\end{tabular}

\vspace{0.8mm}
\legenditem{supportprimary}{Primary selected support}
\hspace{5mm}
\legenditem{supportstructural}{Mandatory or structural support}
\hspace{5mm}
\legenditem{supportauxiliary}{Auxiliary or long-range support}
\end{minipage}

%% file: sections/05_qualitative_image.tex
\begin{figure*}[!t]
\centering
\footnotesize
\setlength{\tabcolsep}{2pt}
\resizebox{\textwidth}{!}{%
\begin{tabular}{@{}ccccccc@{}}
\textbf{FLUX.1-dev Dense}~\cite{flux2024} & CLEAR~\cite{p241216112} & SpargeAttention~\cite{p250218137} & PLADIS~\cite{p250307677} & Chipmunk~\cite{p250603275} & PISA~\cite{p260201077} & AccelAes~\cite{p260312575} \\\noalign{\vskip 2pt}
\raisebox{-.5\height}{\includegraphics[width=.132\textwidth,keepaspectratio]{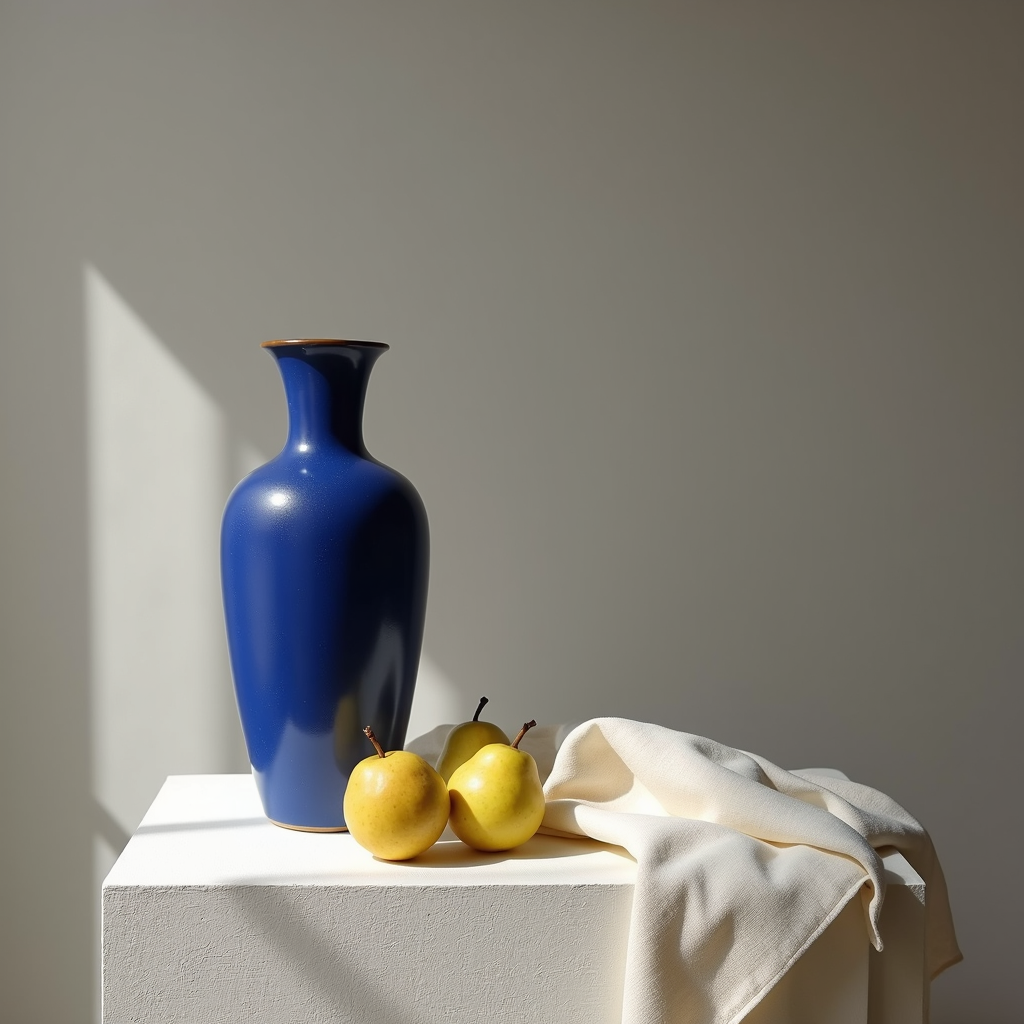}} & \raisebox{-.5\height}{\includegraphics[width=.132\textwidth,keepaspectratio]{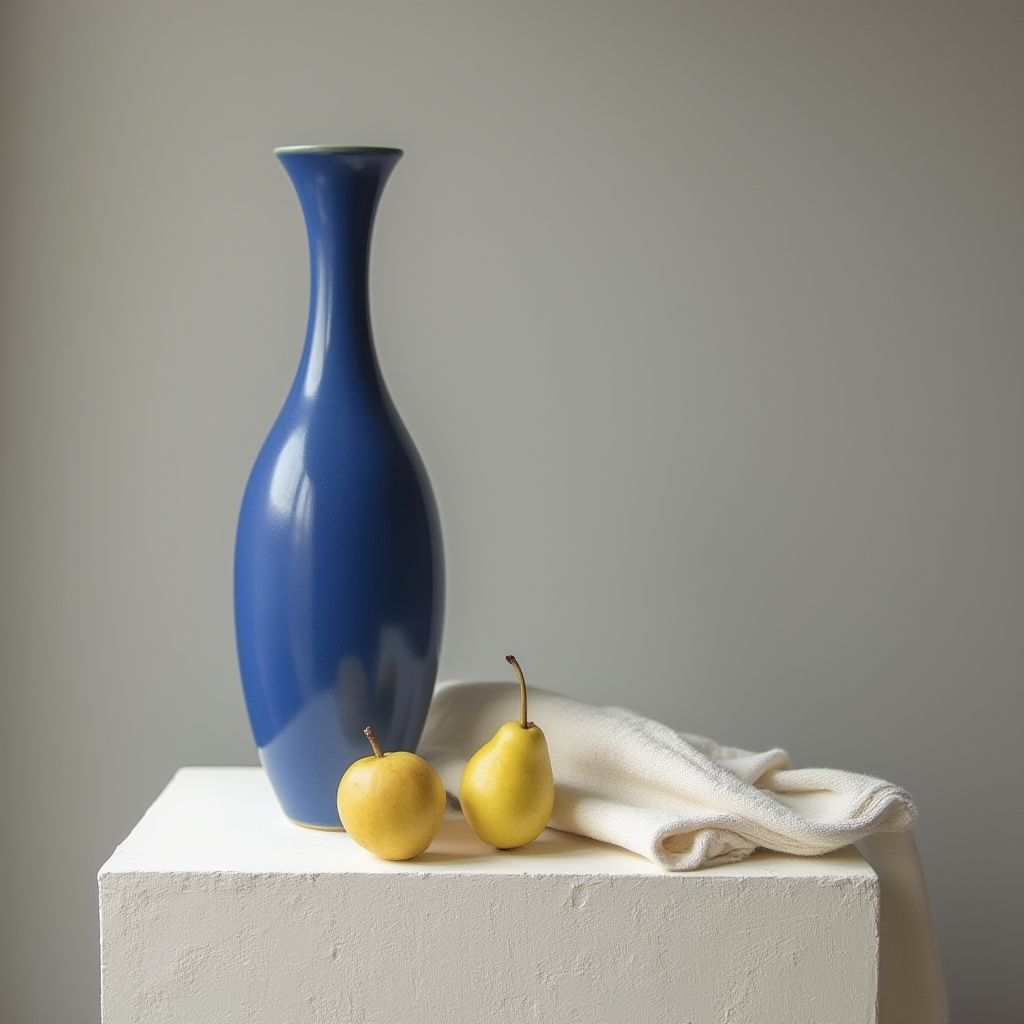}} & \raisebox{-.5\height}{\includegraphics[width=.132\textwidth,keepaspectratio]{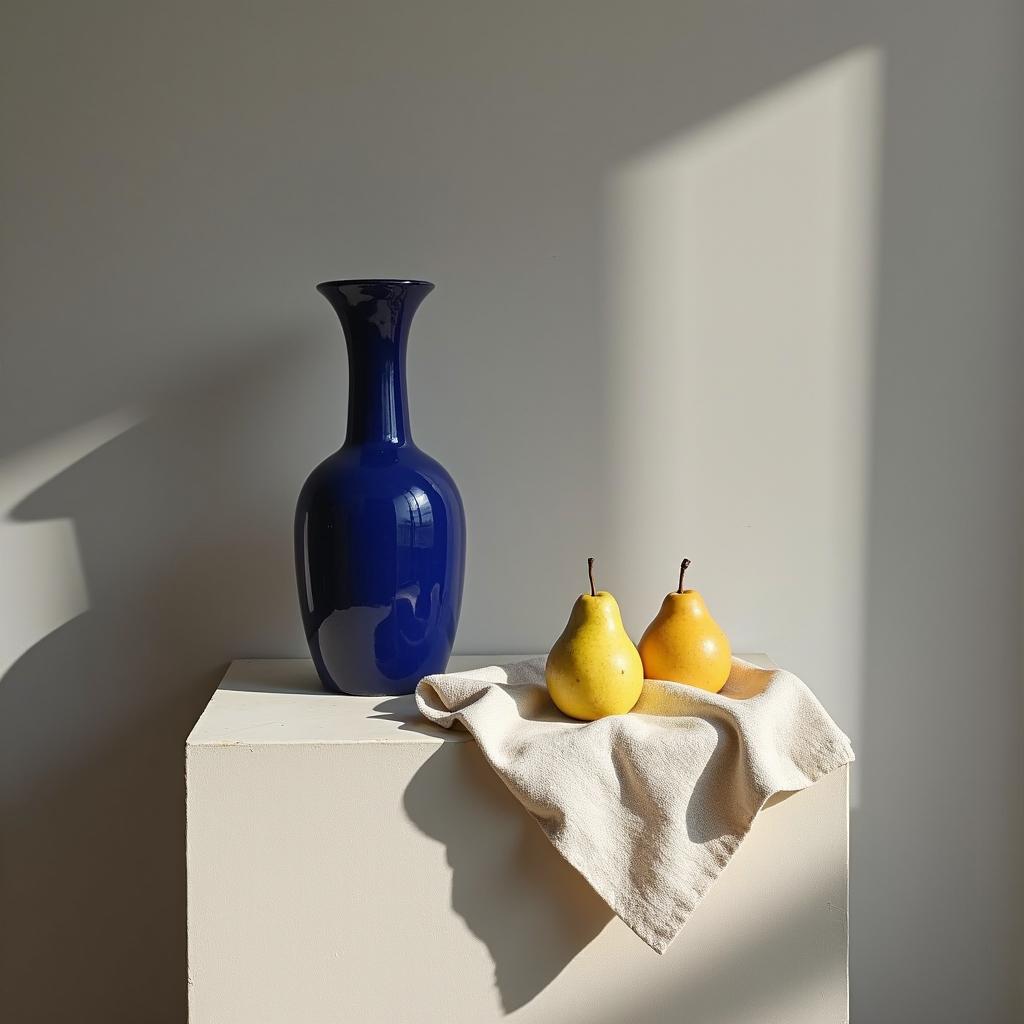}} & \raisebox{-.5\height}{\includegraphics[width=.132\textwidth,keepaspectratio]{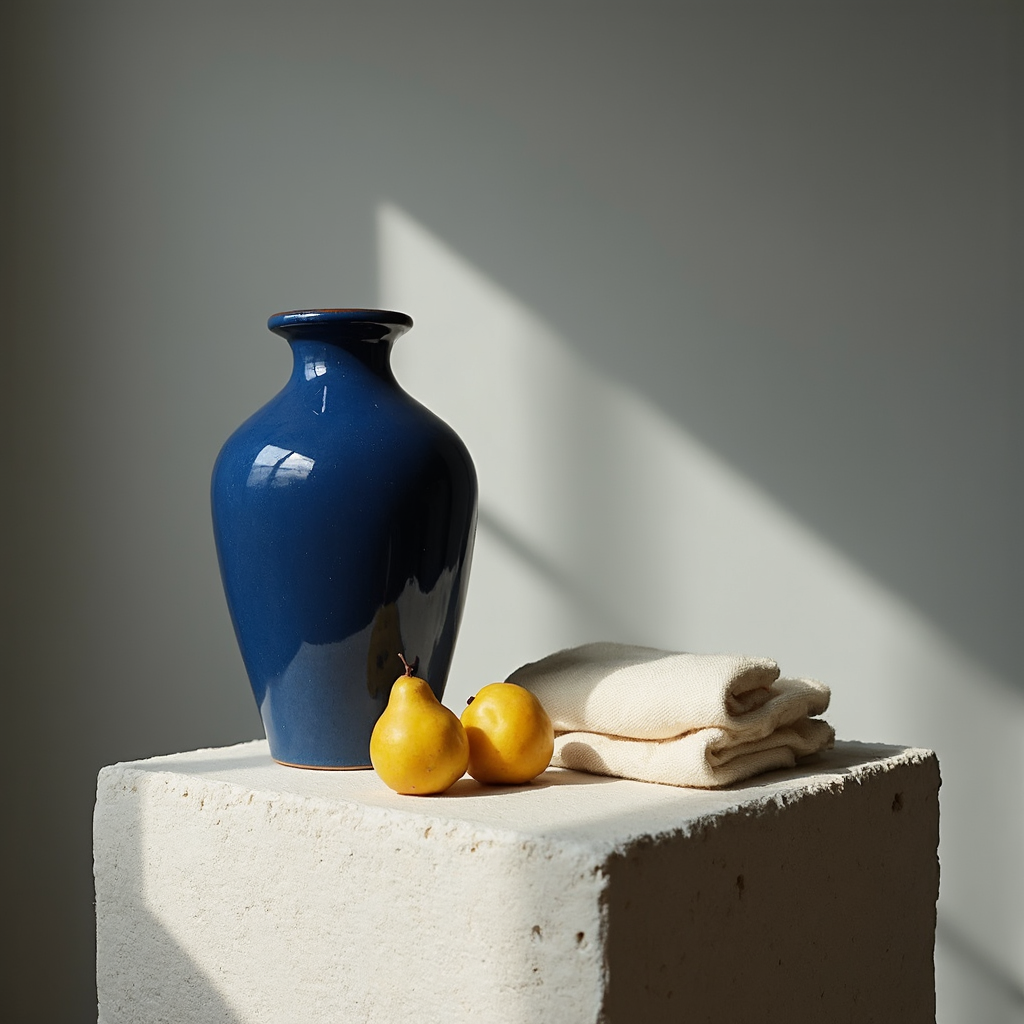}} & \raisebox{-.5\height}{\includegraphics[width=.132\textwidth,keepaspectratio]{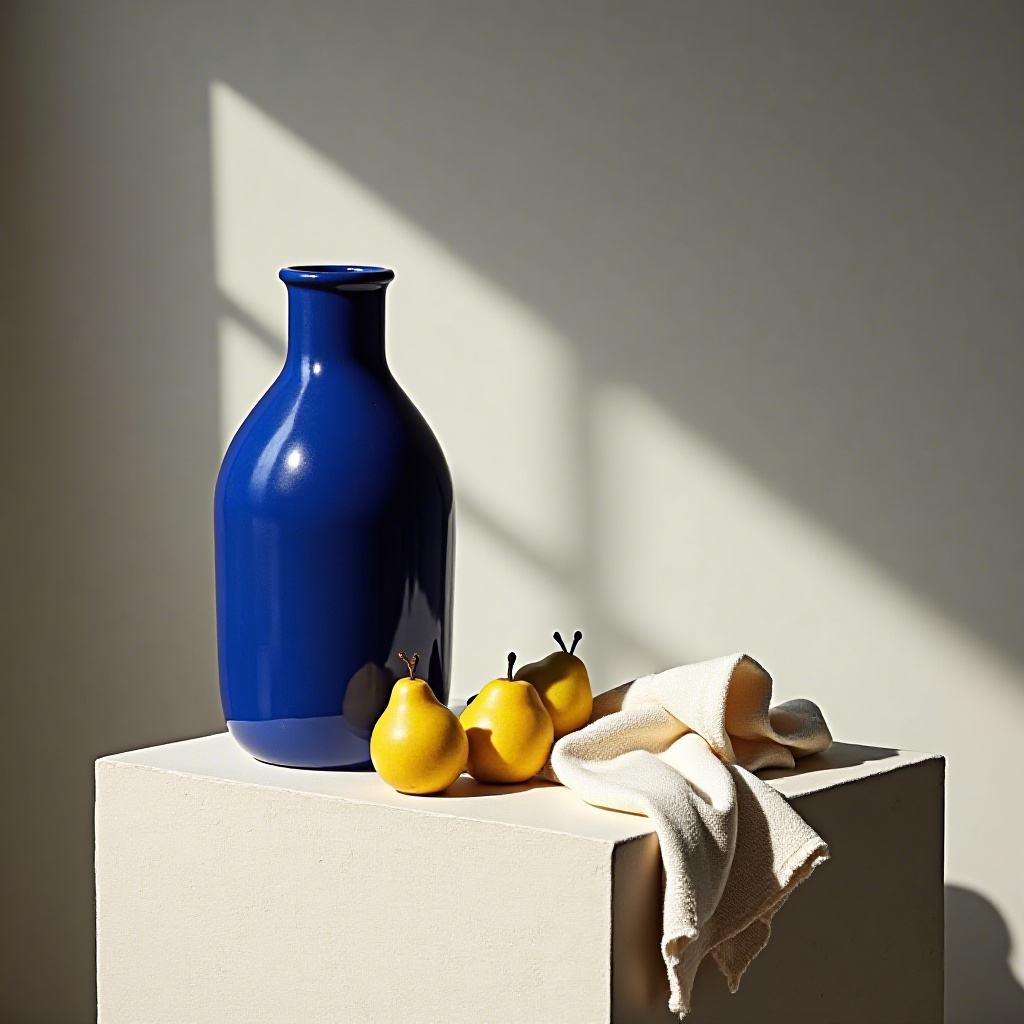}} & \raisebox{-.5\height}{\includegraphics[width=.132\textwidth,keepaspectratio]{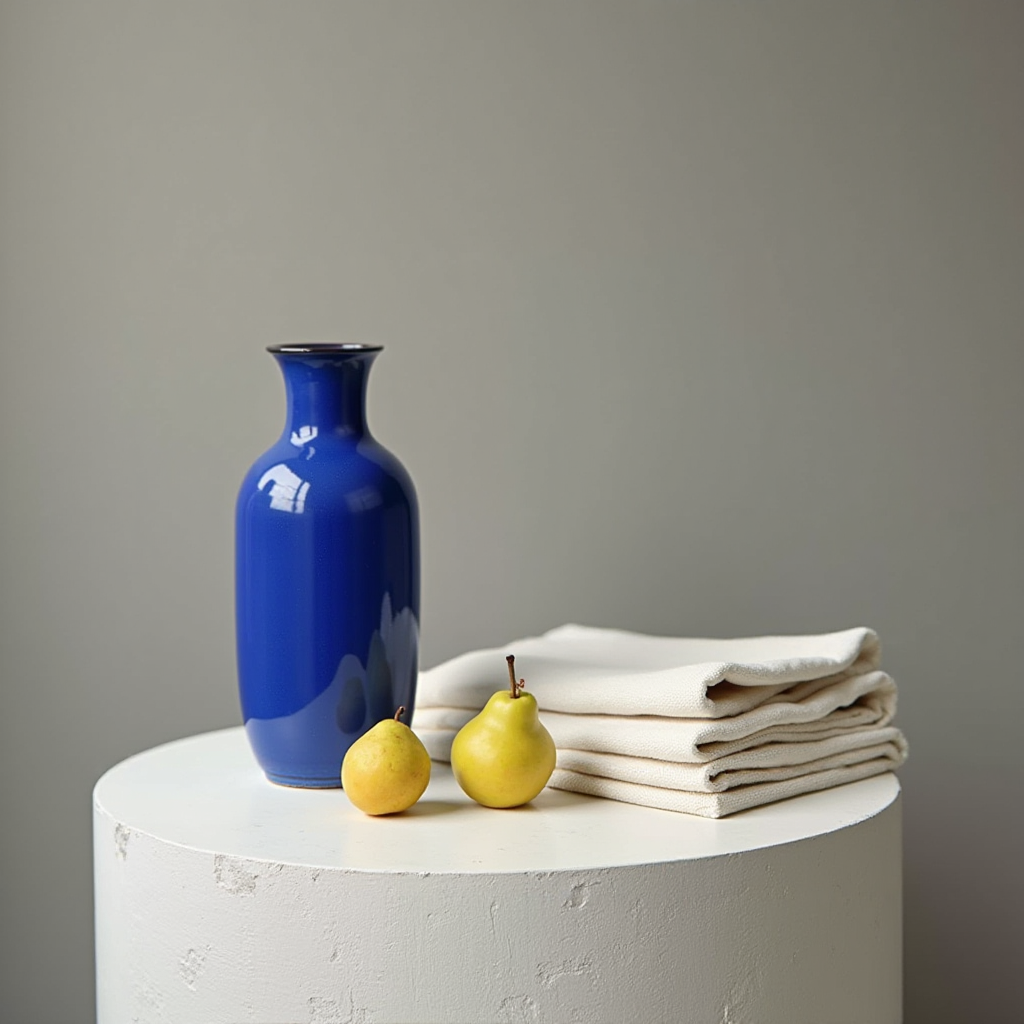}} & \raisebox{-.5\height}{\includegraphics[width=.132\textwidth,keepaspectratio]{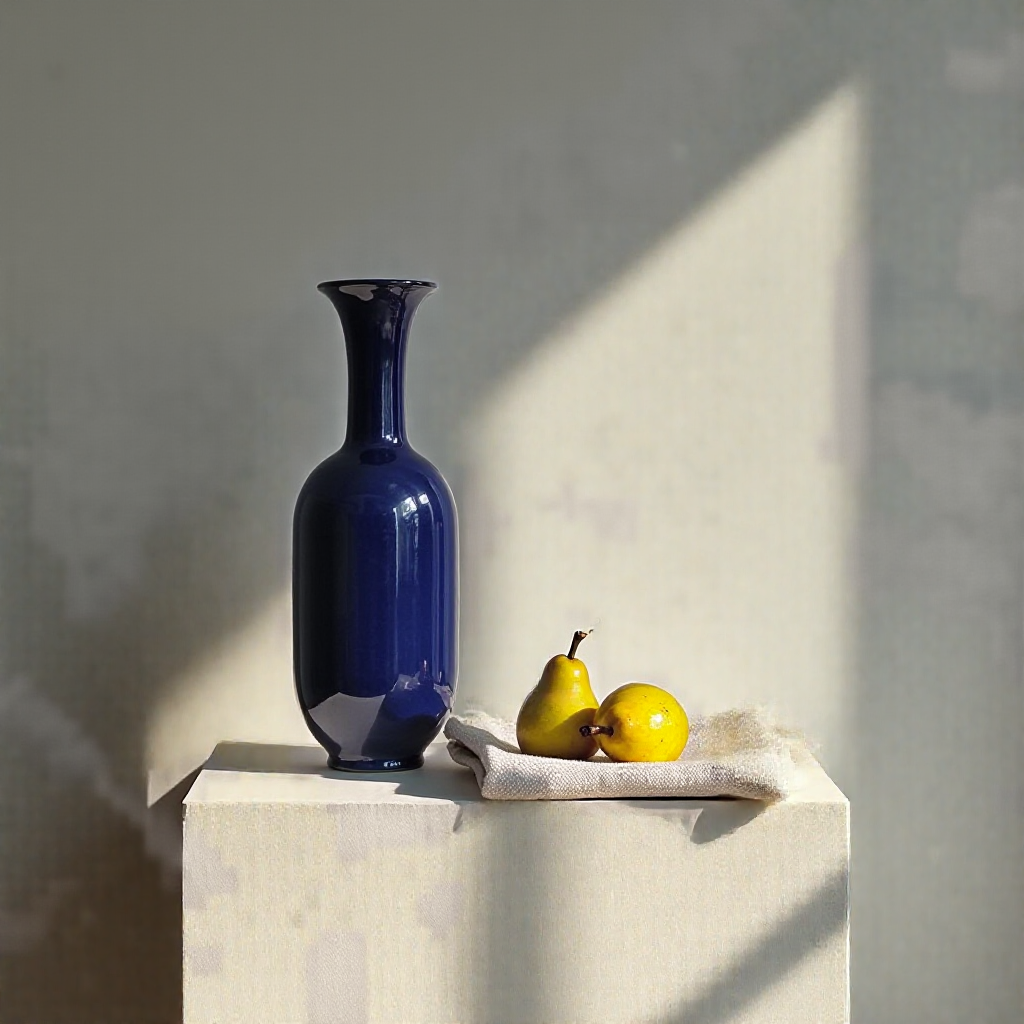}} \\\noalign{\vskip 2pt}
\raisebox{-.5\height}{\includegraphics[width=.132\textwidth,keepaspectratio]{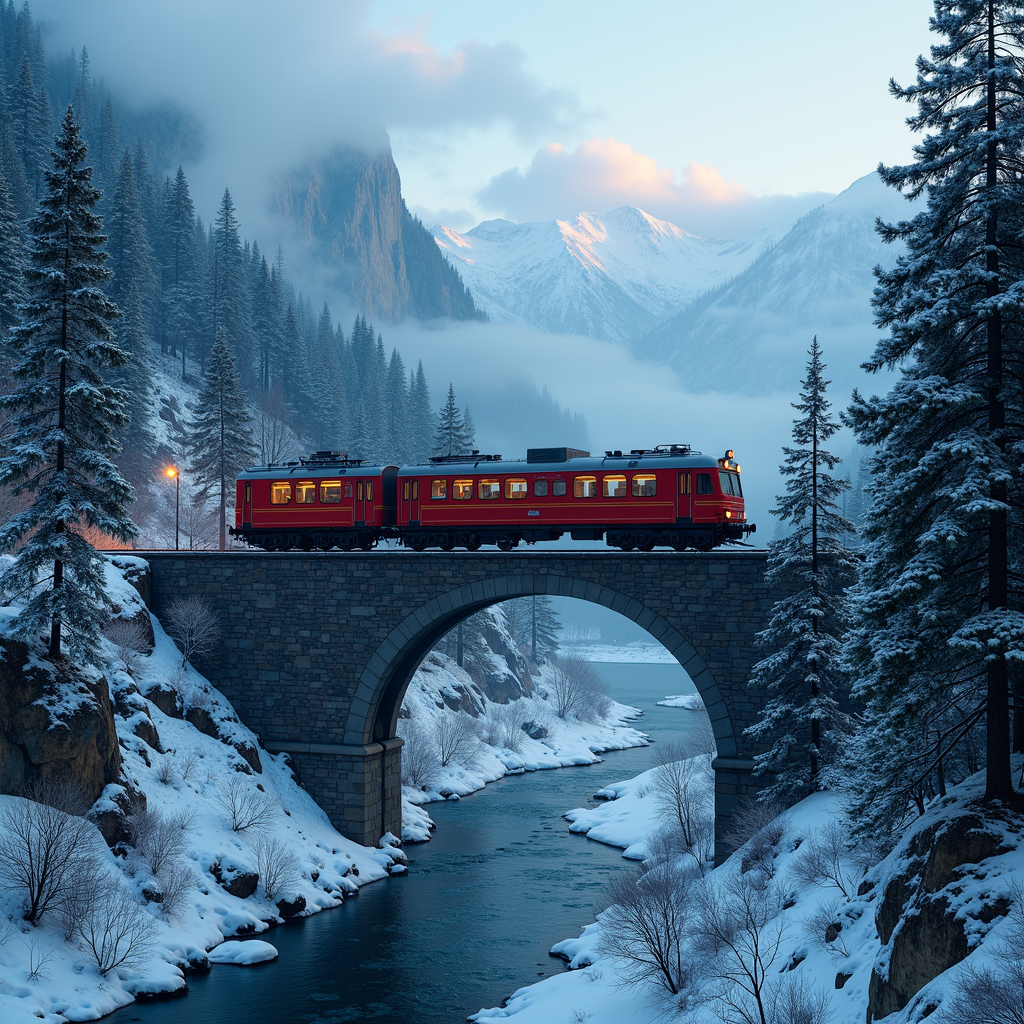}} & \raisebox{-.5\height}{\includegraphics[width=.132\textwidth,keepaspectratio]{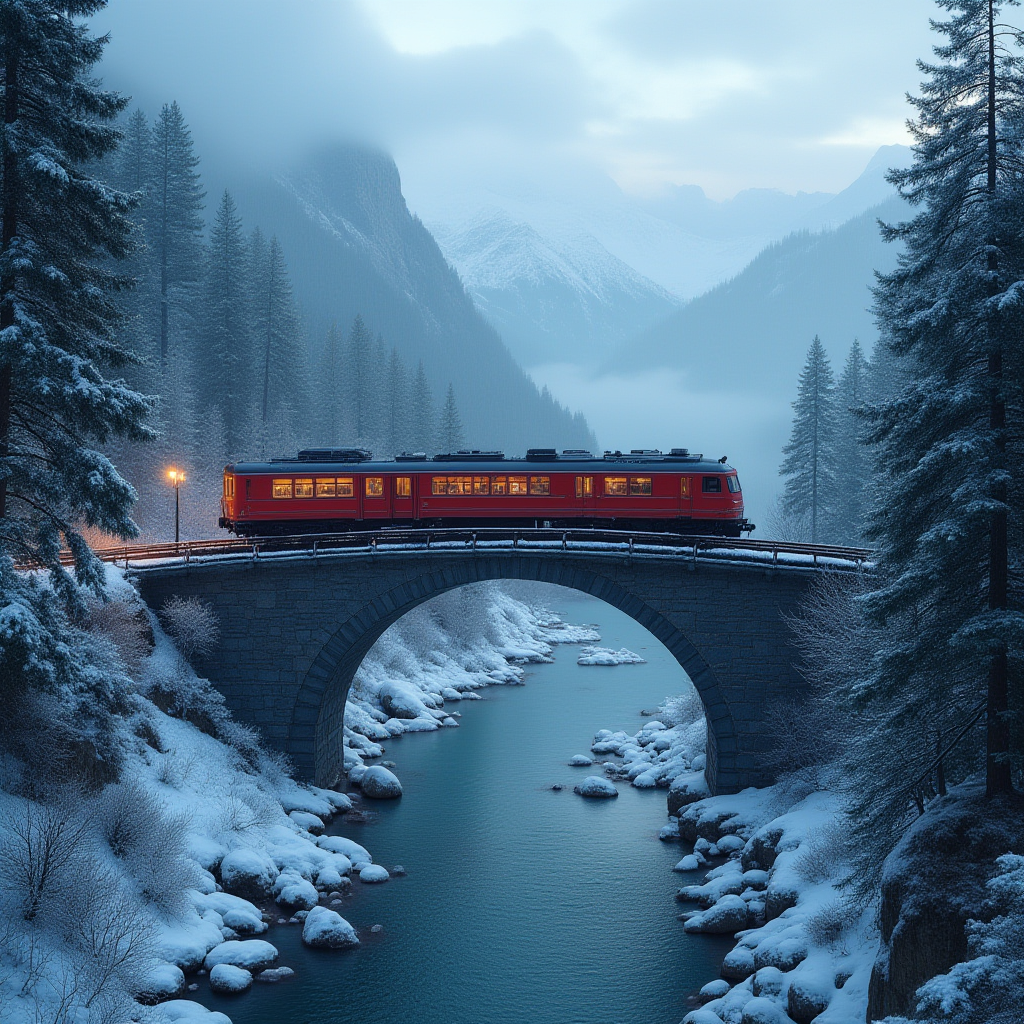}} & \raisebox{-.5\height}{\includegraphics[width=.132\textwidth,keepaspectratio]{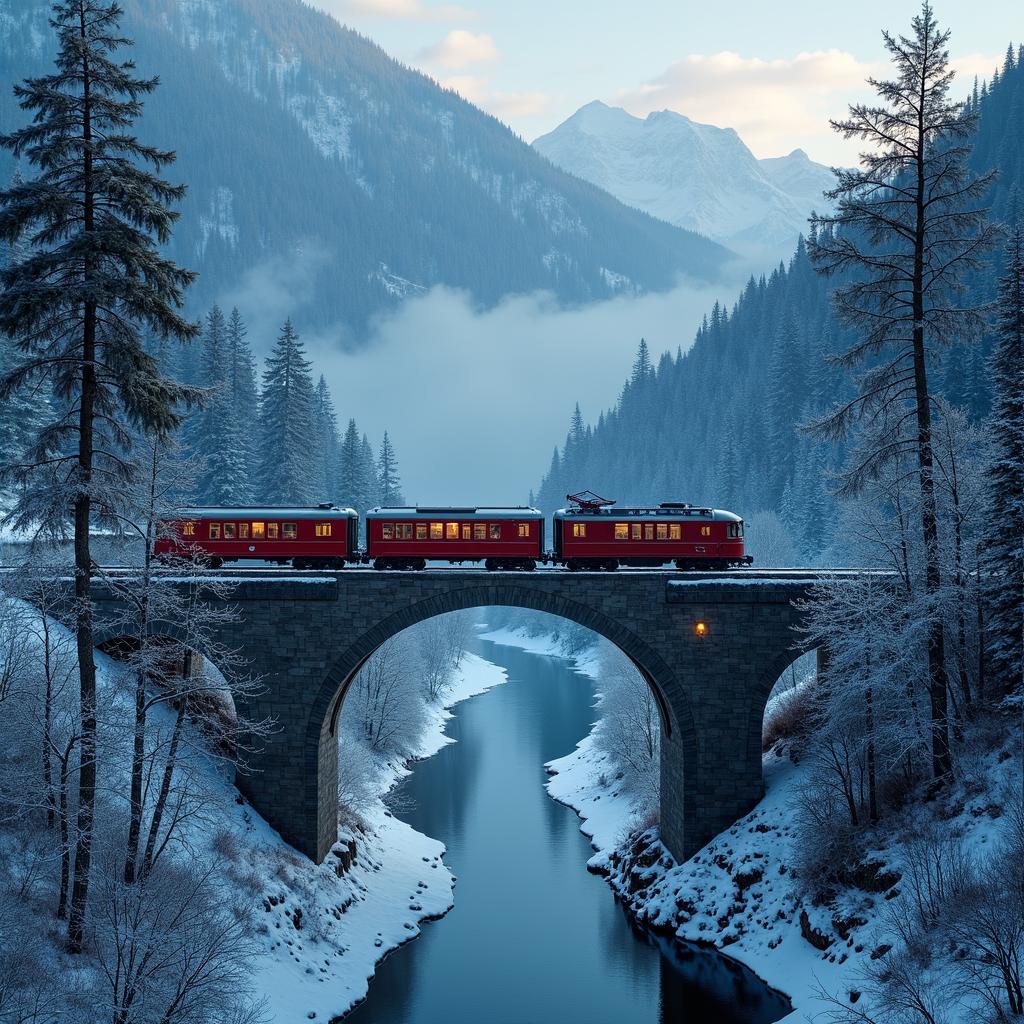}} & \raisebox{-.5\height}{\includegraphics[width=.132\textwidth,keepaspectratio]{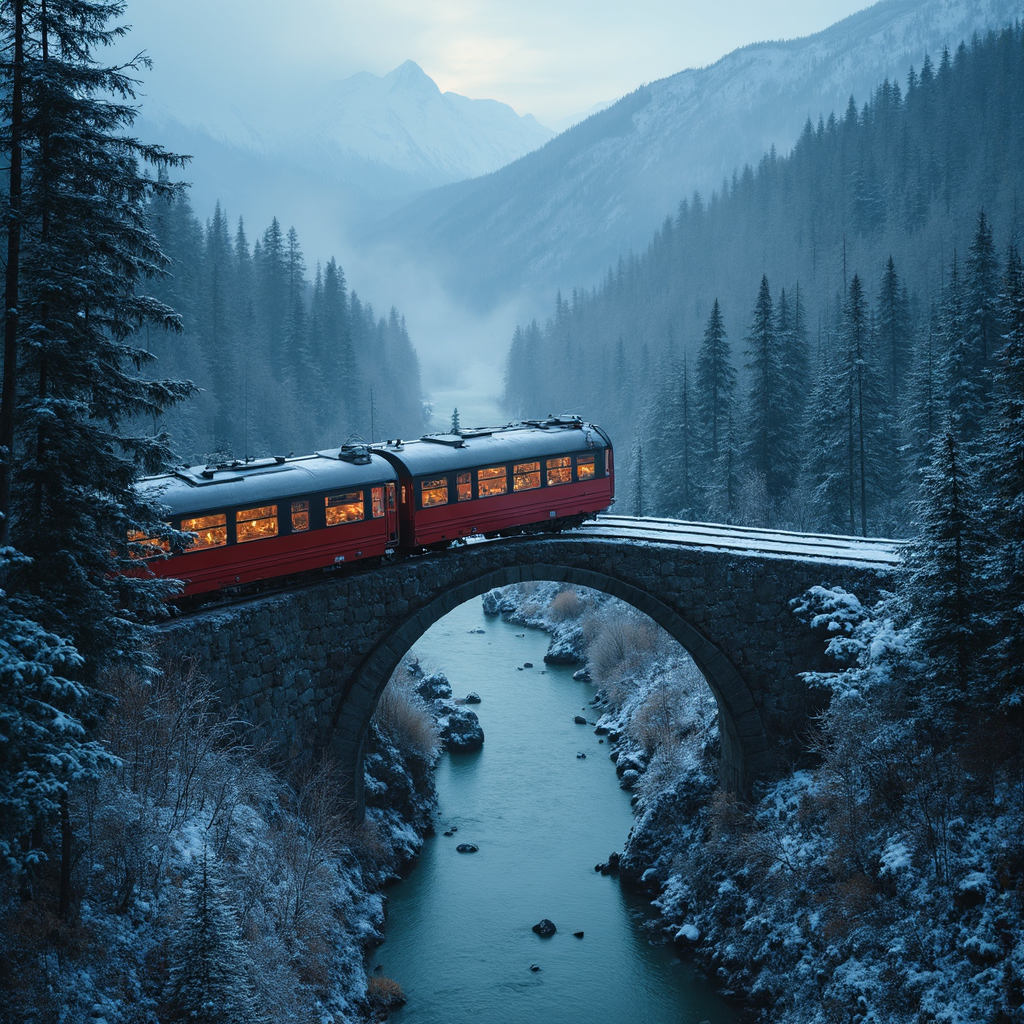}} & \raisebox{-.5\height}{\includegraphics[width=.132\textwidth,keepaspectratio]{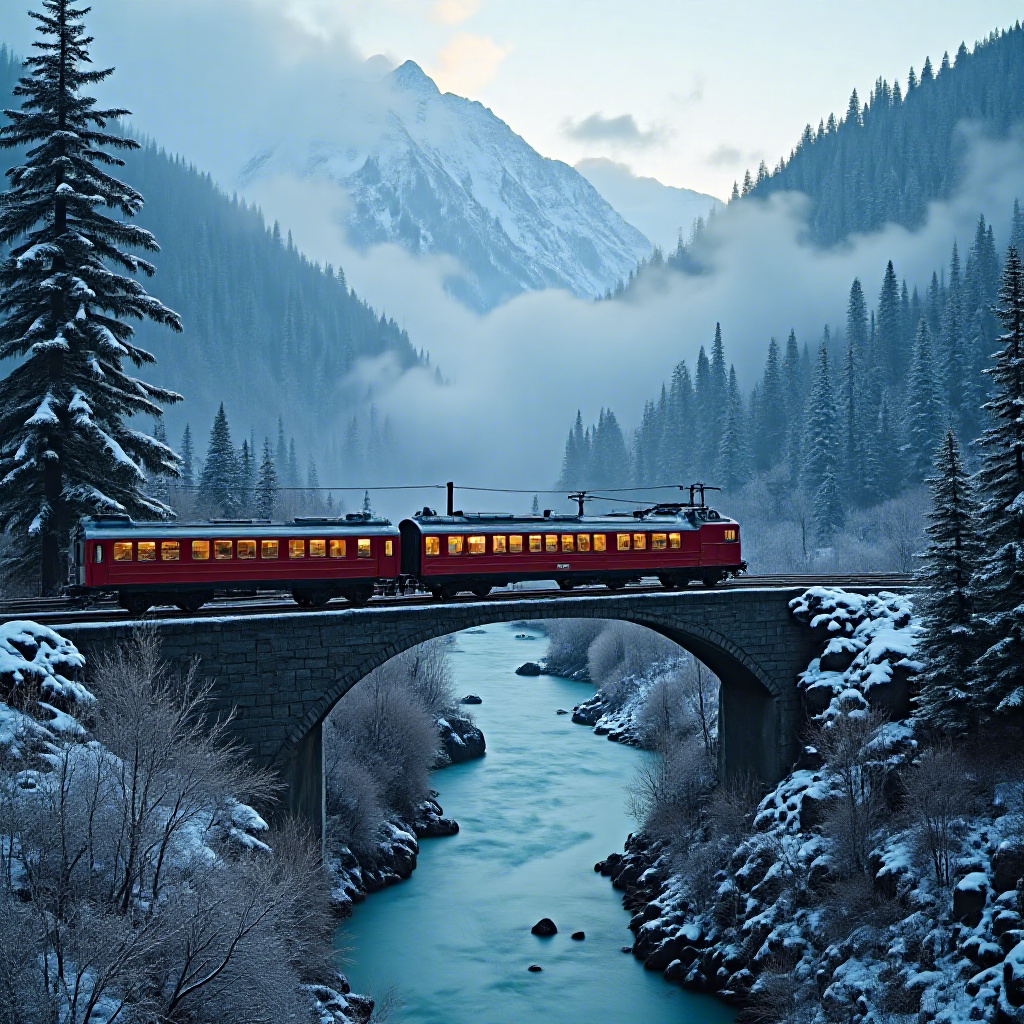}} & \raisebox{-.5\height}{\includegraphics[width=.132\textwidth,keepaspectratio]{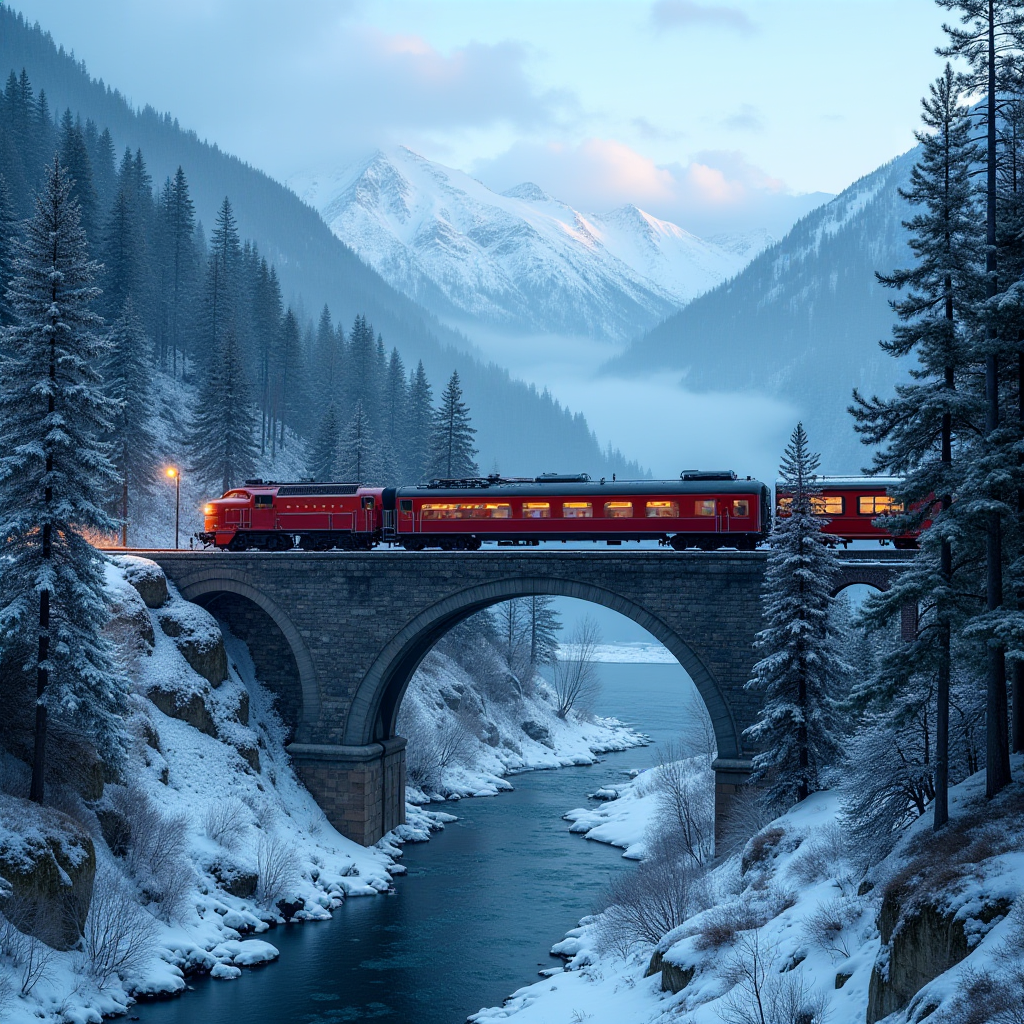}} & \raisebox{-.5\height}{\includegraphics[width=.132\textwidth,keepaspectratio]{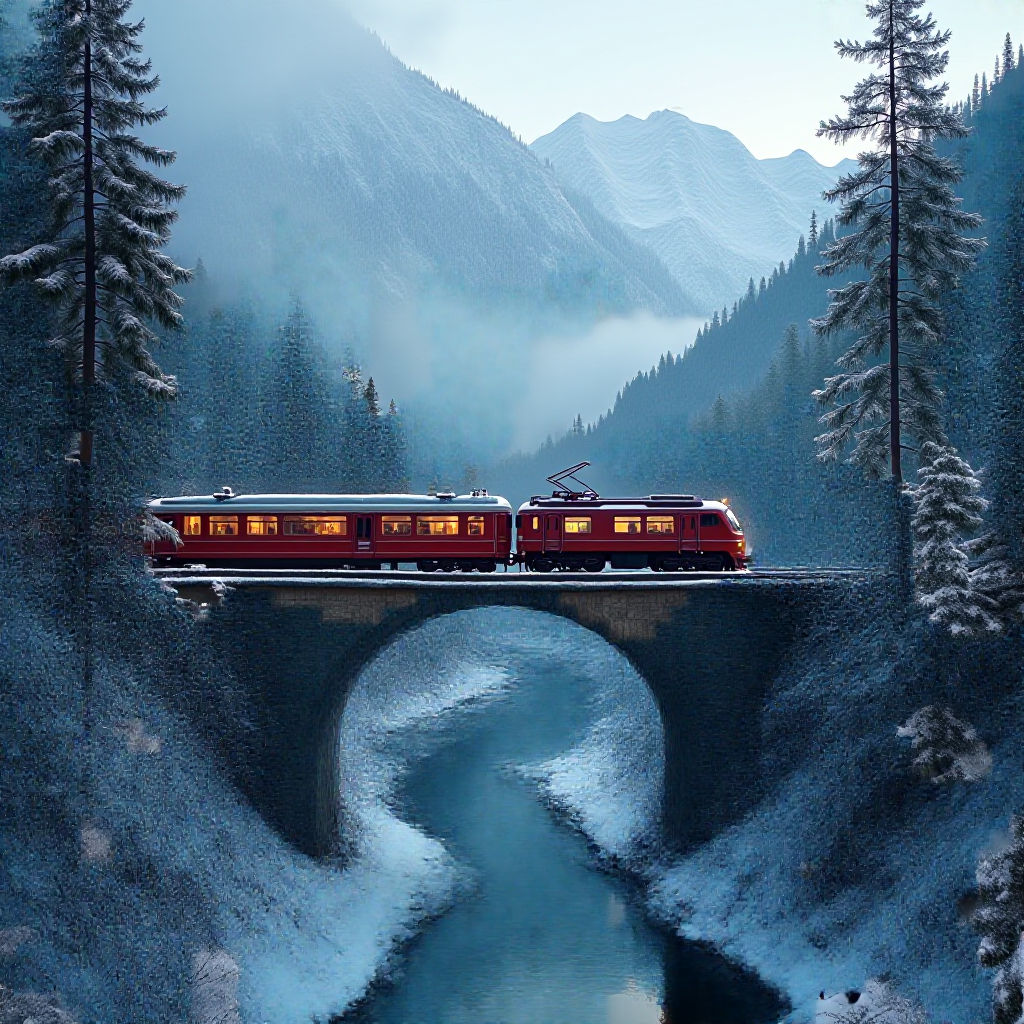}} \\\noalign{\vskip 2pt}
\raisebox{-.5\height}{\includegraphics[width=.132\textwidth,keepaspectratio]{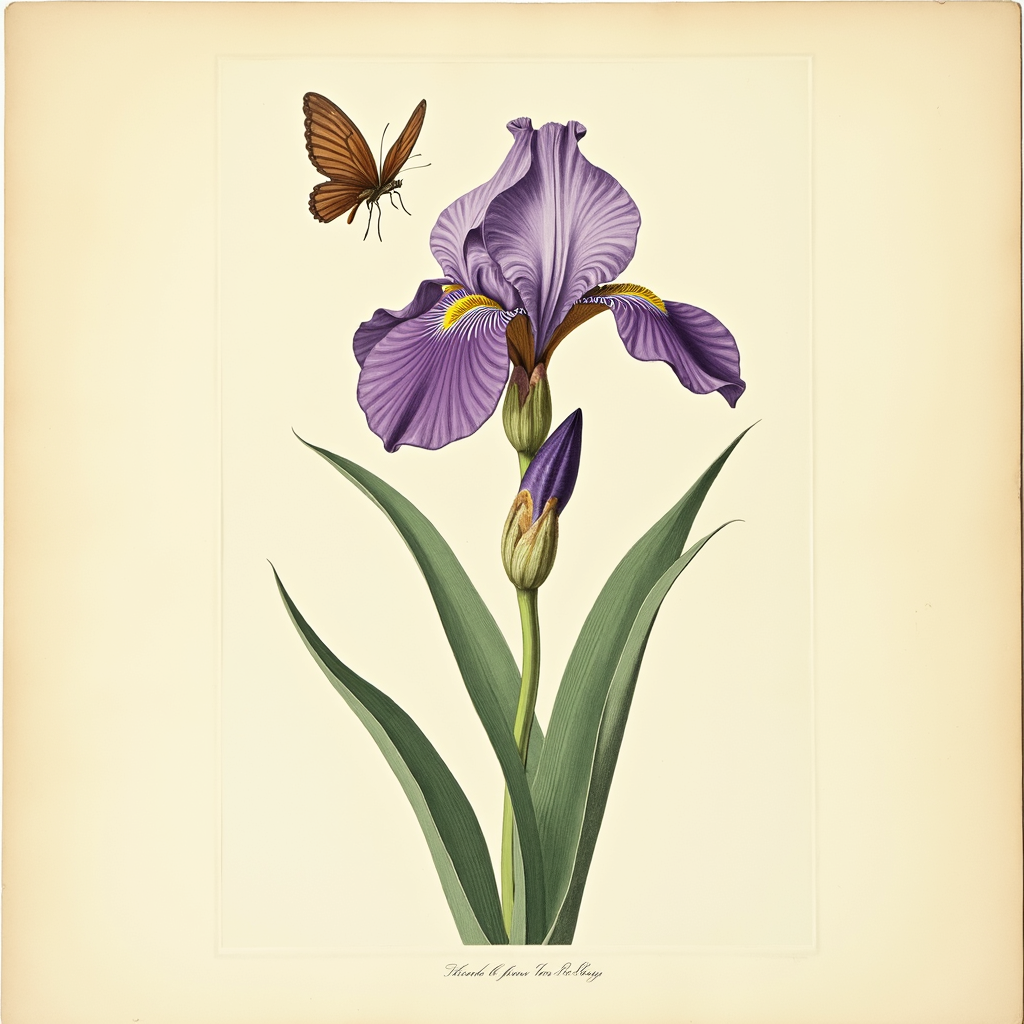}} & \raisebox{-.5\height}{\includegraphics[width=.132\textwidth,keepaspectratio]{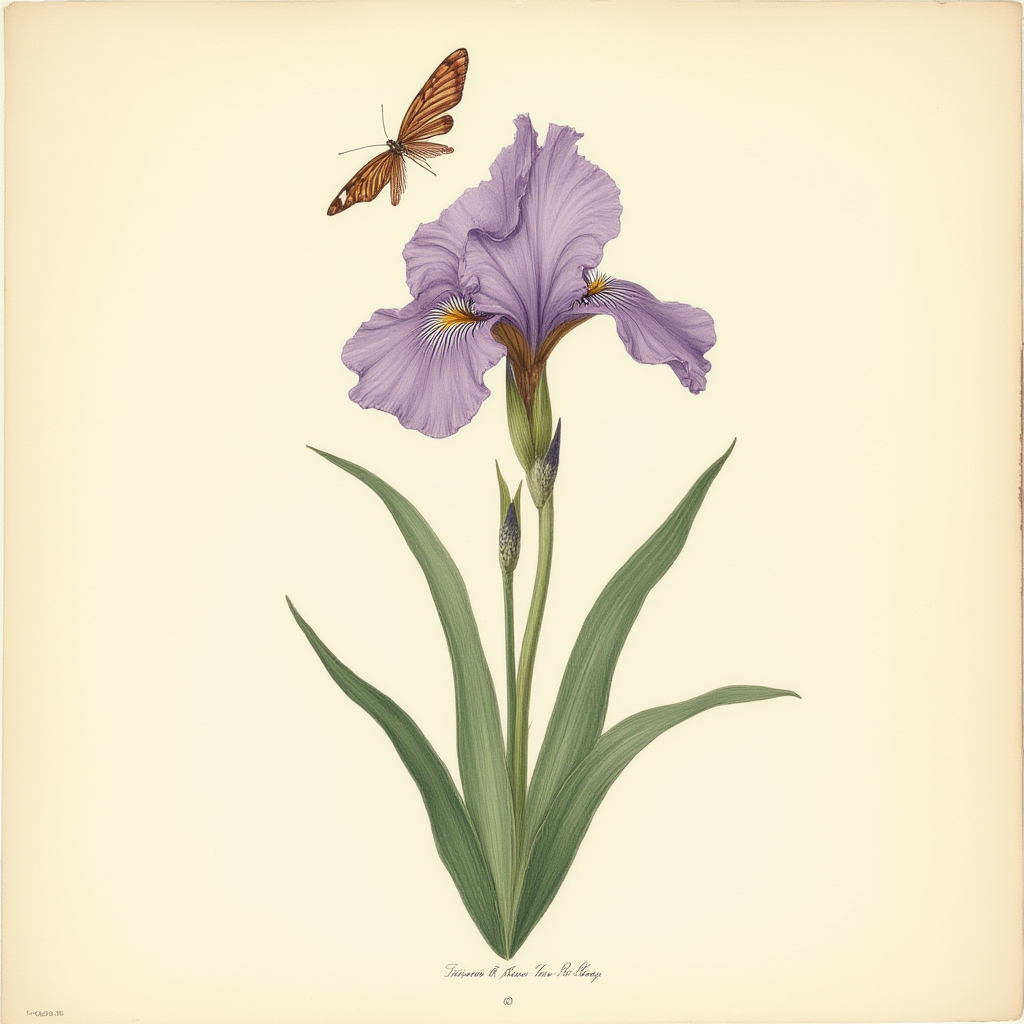}} & \raisebox{-.5\height}{\includegraphics[width=.132\textwidth,keepaspectratio]{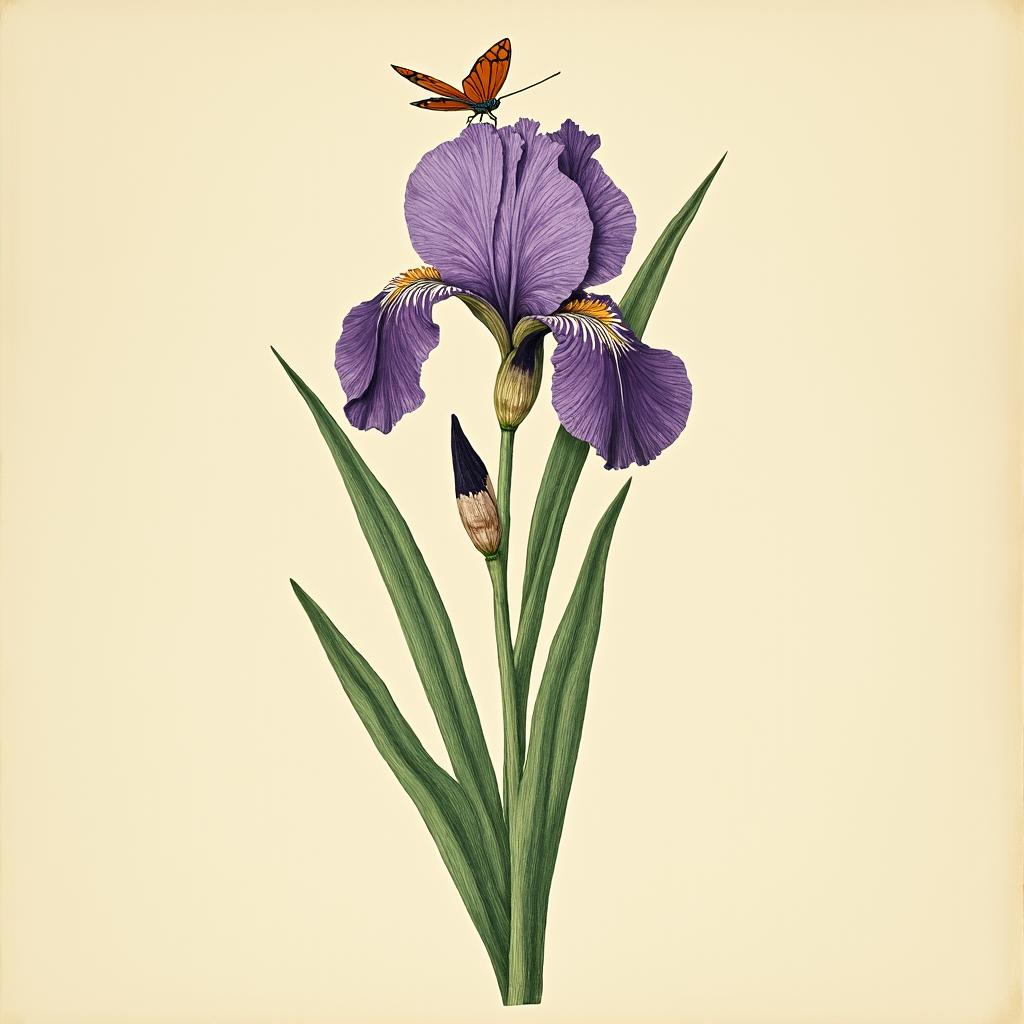}} & \raisebox{-.5\height}{\includegraphics[width=.132\textwidth,keepaspectratio]{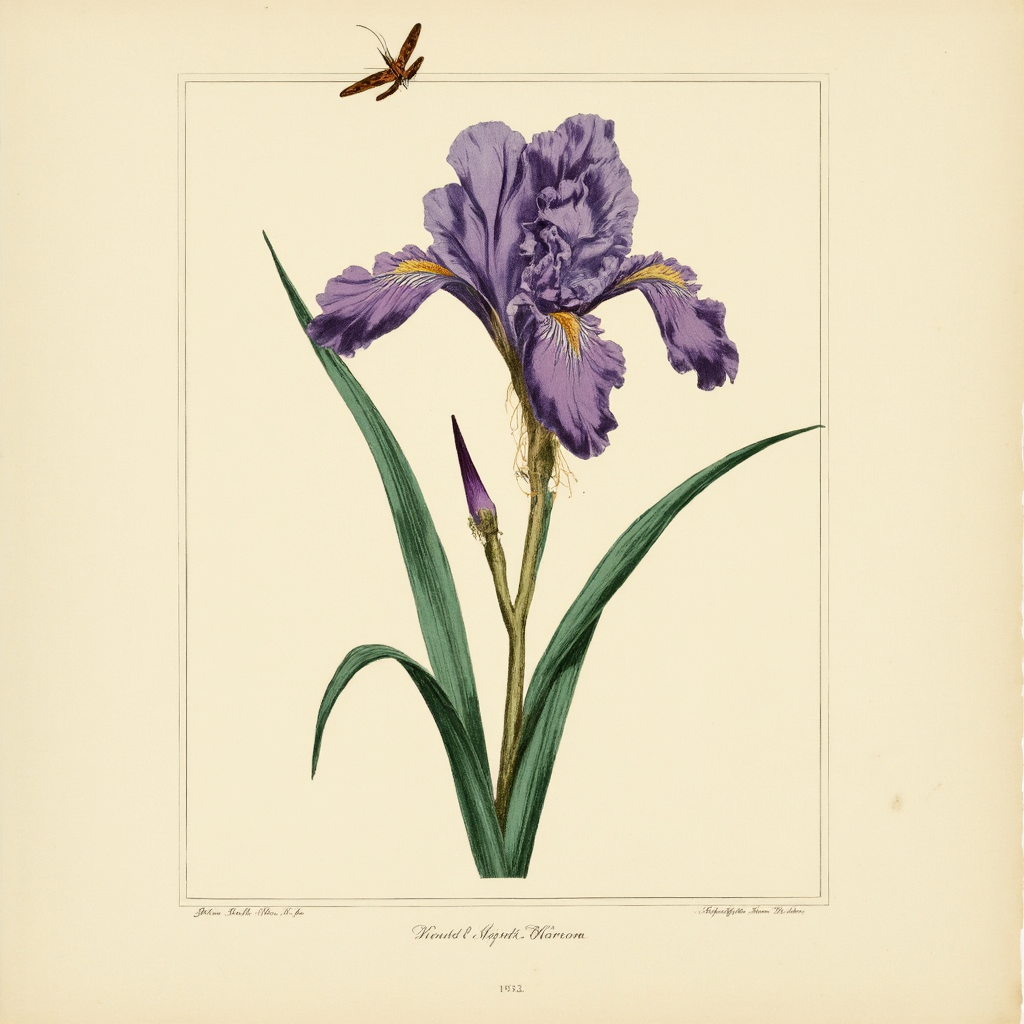}} & \raisebox{-.5\height}{\includegraphics[width=.132\textwidth,keepaspectratio]{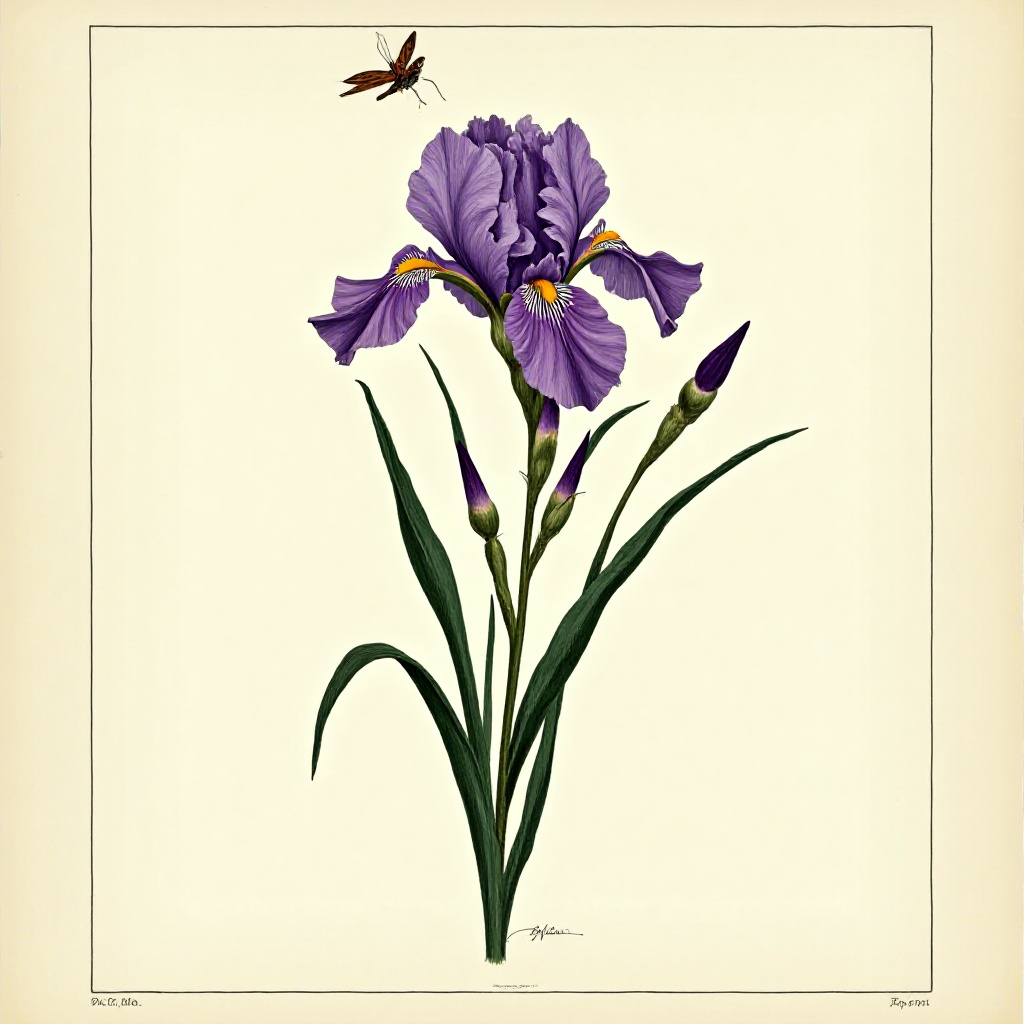}} & \raisebox{-.5\height}{\includegraphics[width=.132\textwidth,keepaspectratio]{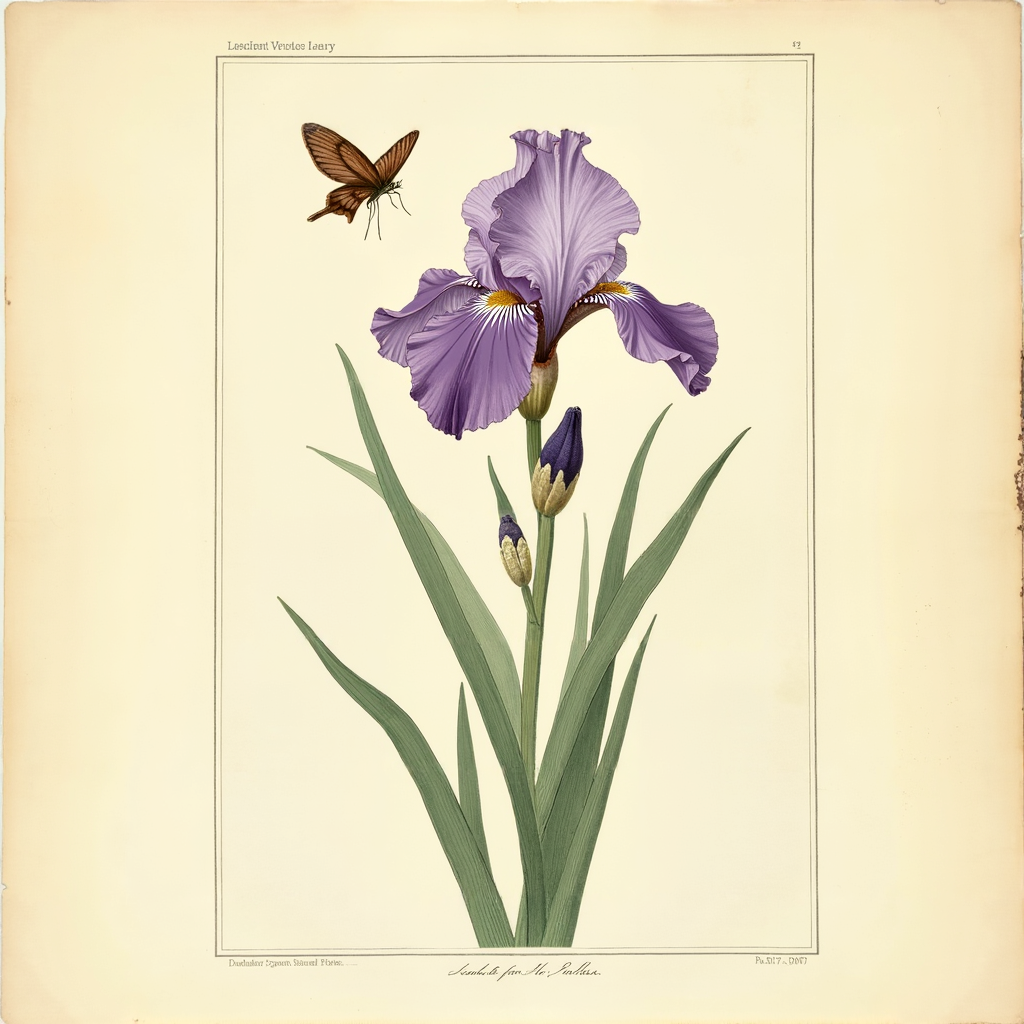}} & \raisebox{-.5\height}{\includegraphics[width=.132\textwidth,keepaspectratio]{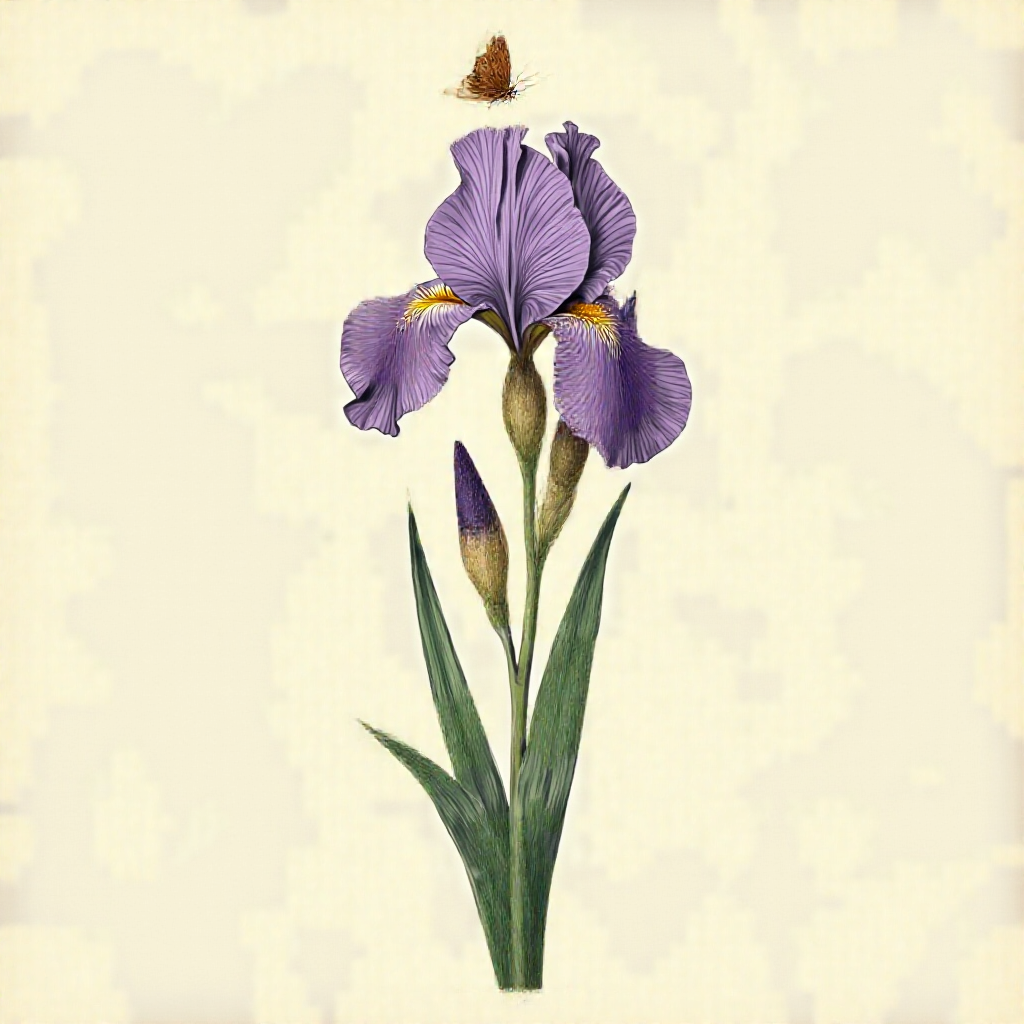}} \\\noalign{\vskip 2pt}
\raisebox{-.5\height}{\includegraphics[width=.132\textwidth,keepaspectratio]{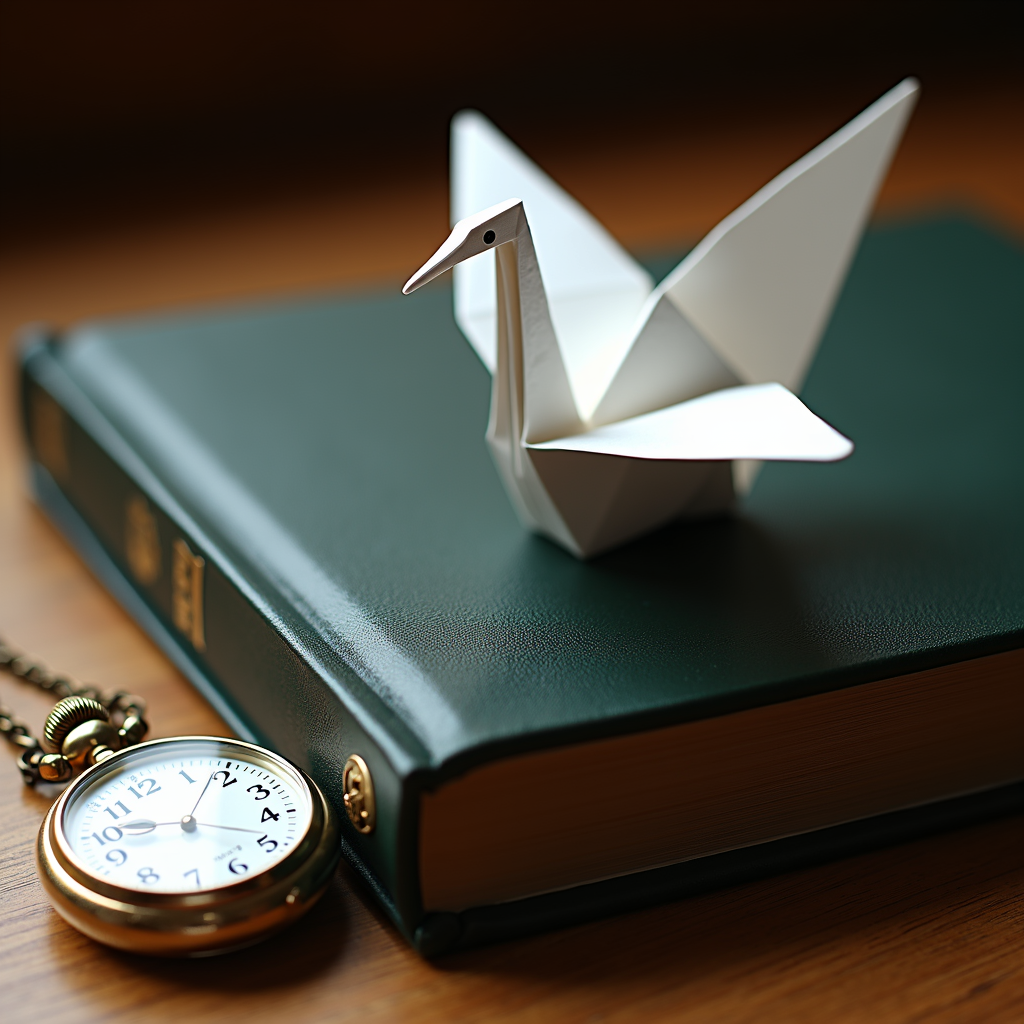}} & \raisebox{-.5\height}{\includegraphics[width=.132\textwidth,keepaspectratio]{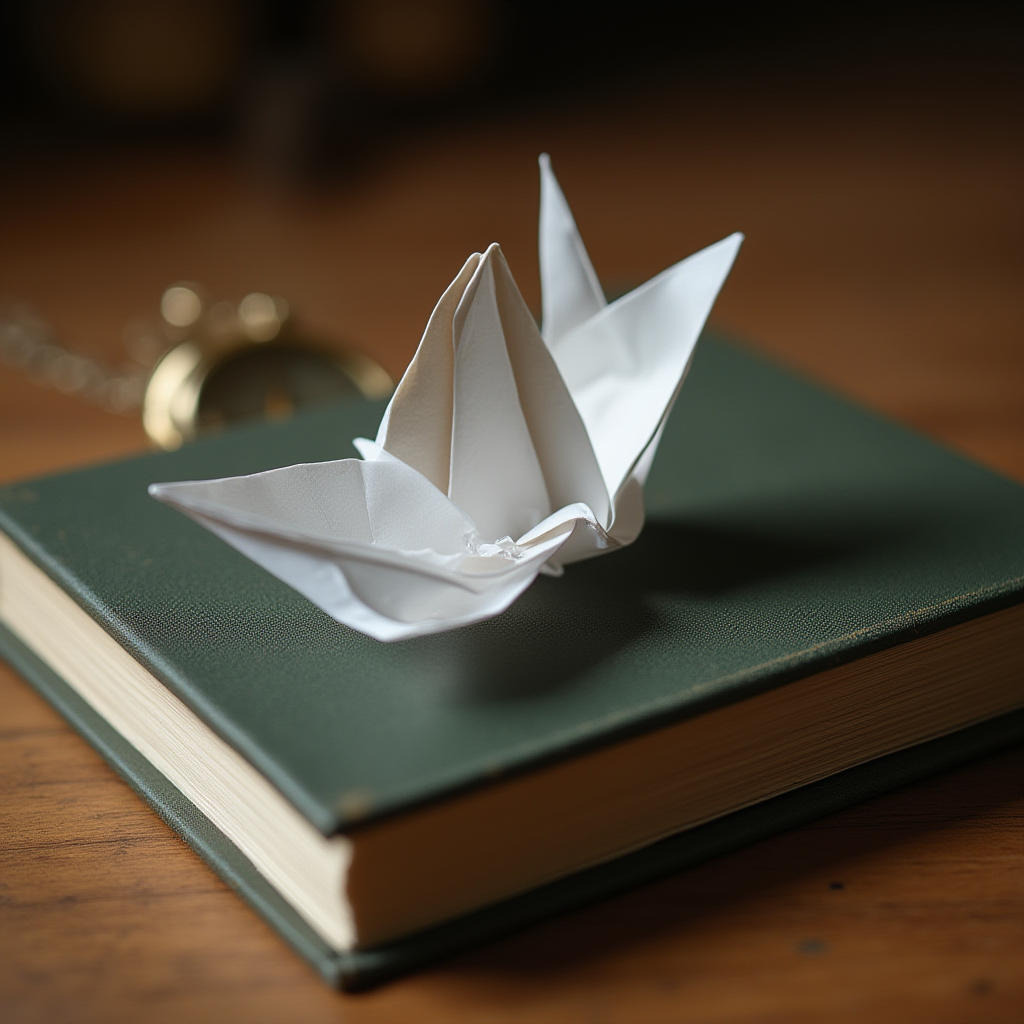}} & \raisebox{-.5\height}{\includegraphics[width=.132\textwidth,keepaspectratio]{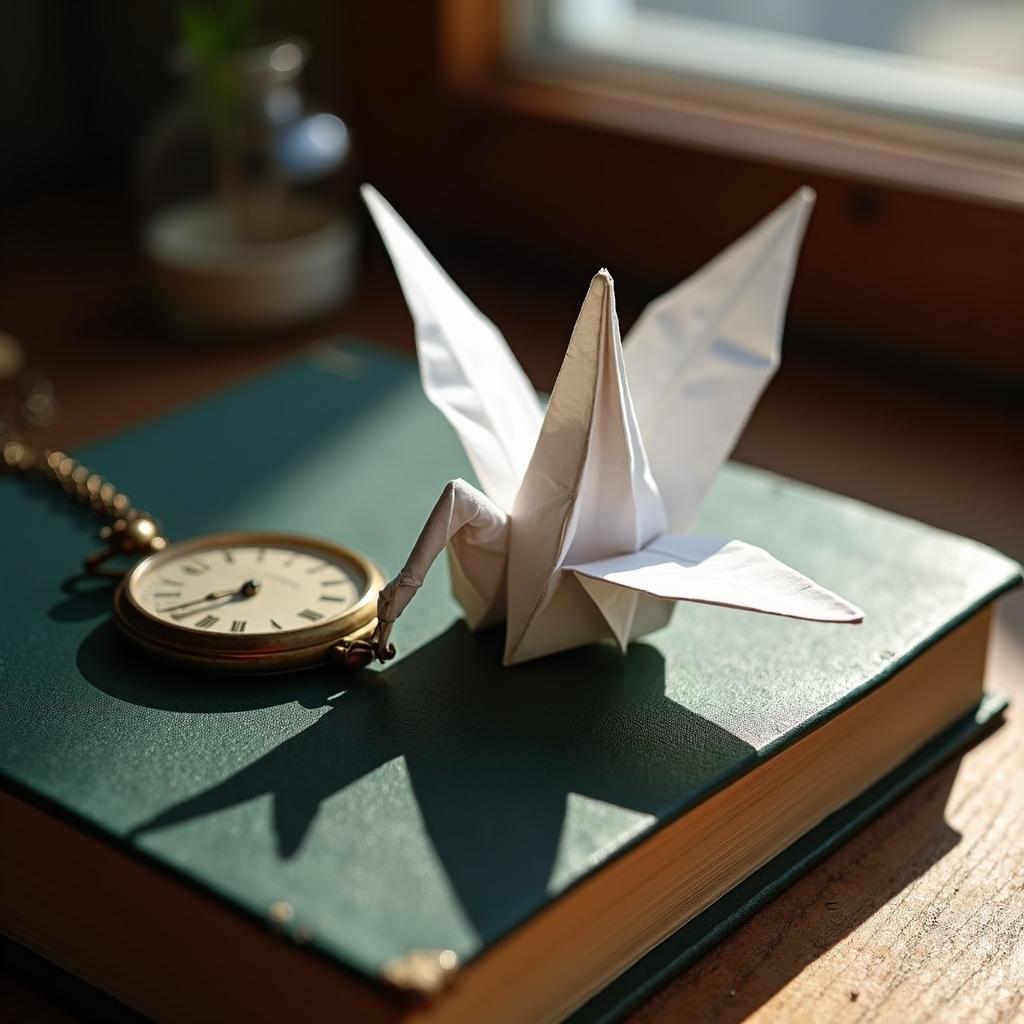}} & \raisebox{-.5\height}{\includegraphics[width=.132\textwidth,keepaspectratio]{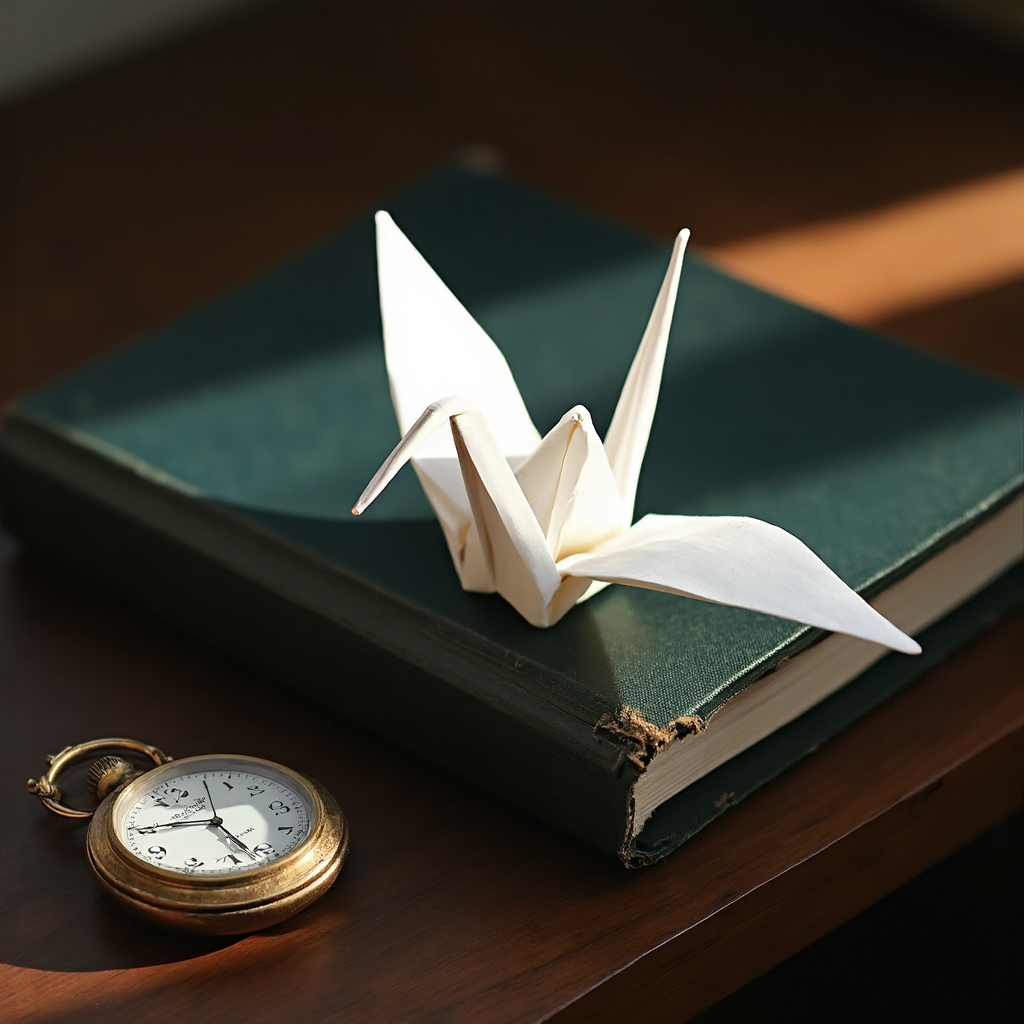}} & \raisebox{-.5\height}{\includegraphics[width=.132\textwidth,keepaspectratio]{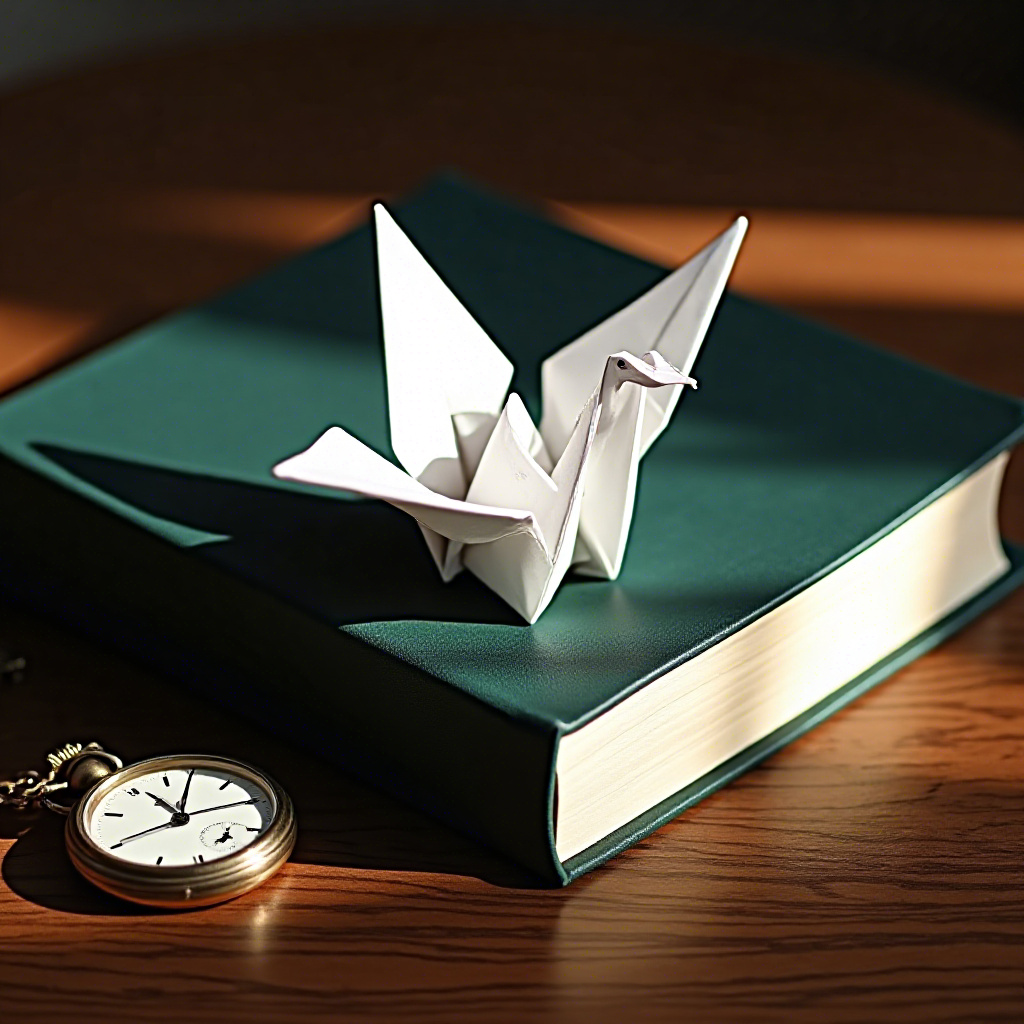}} & \raisebox{-.5\height}{\includegraphics[width=.132\textwidth,keepaspectratio]{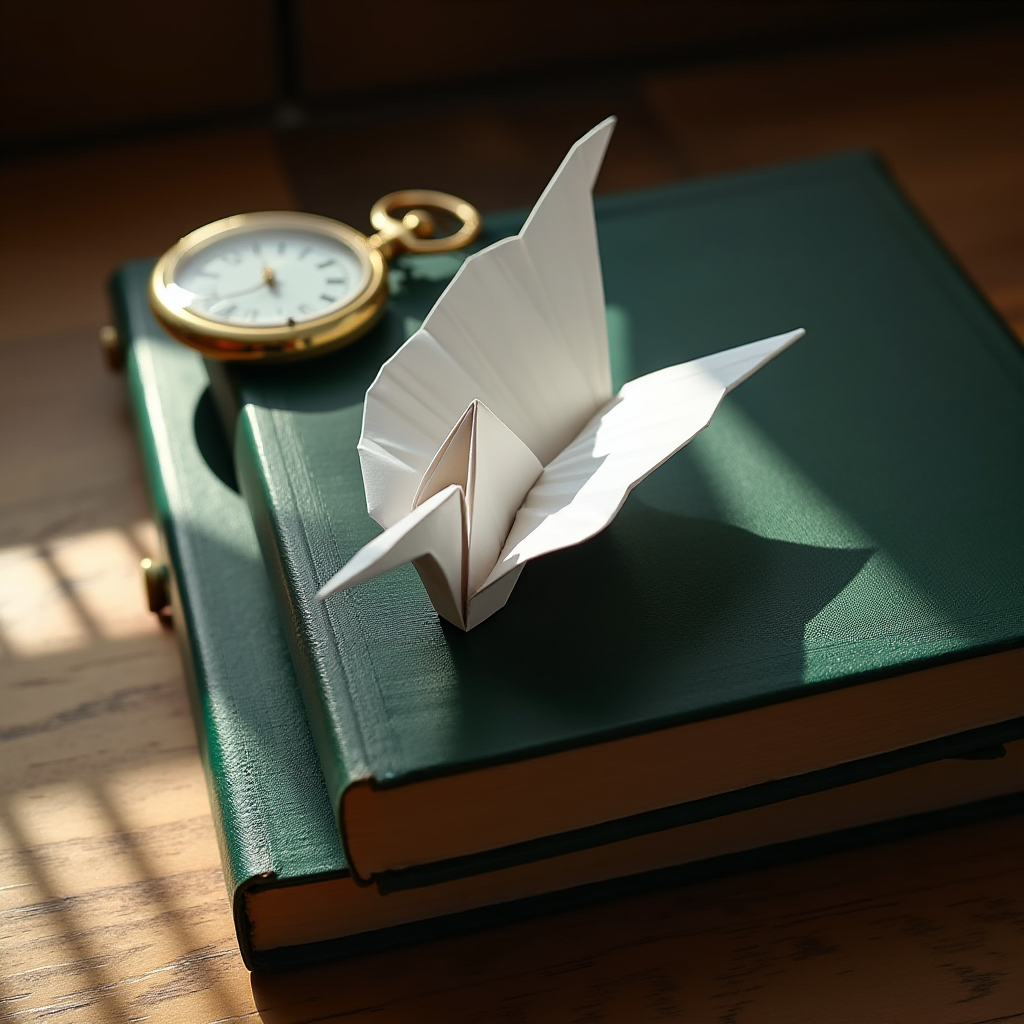}} & \raisebox{-.5\height}{\includegraphics[width=.132\textwidth,keepaspectratio]{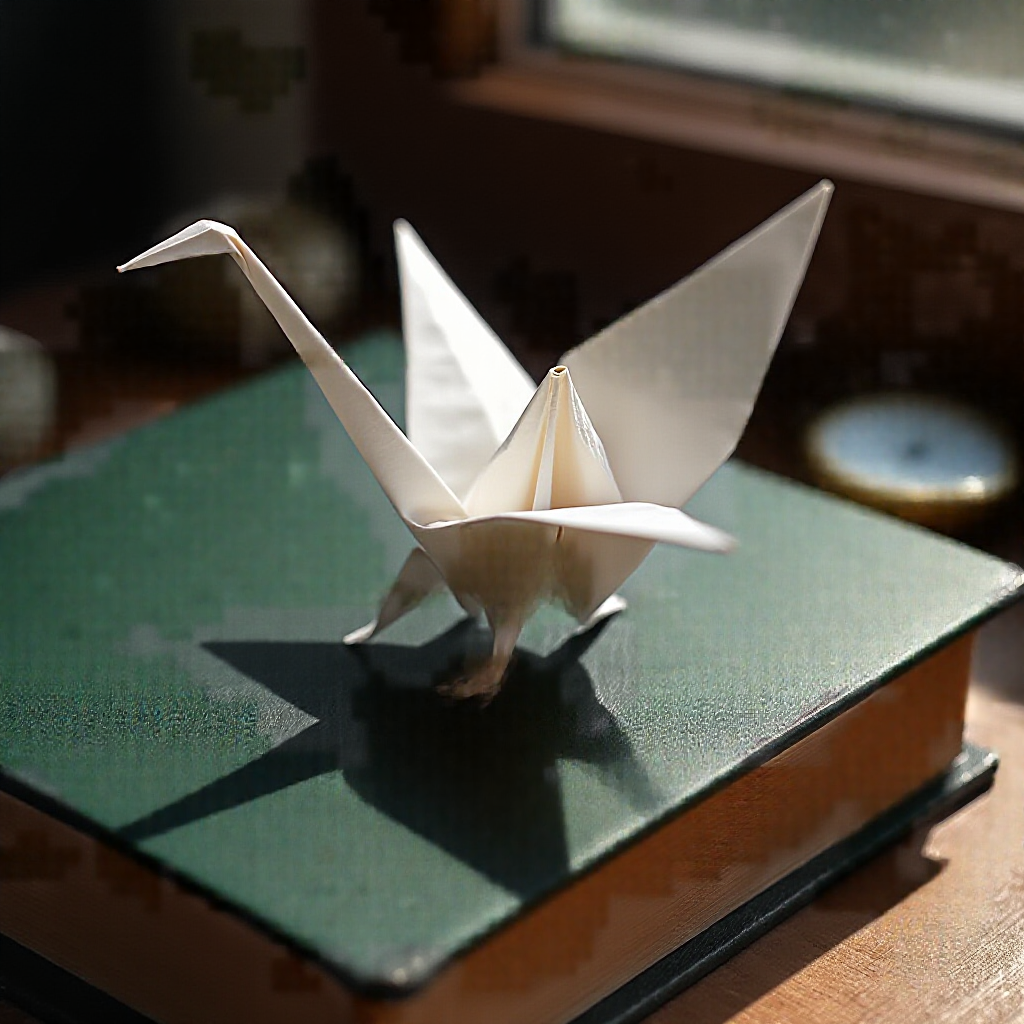}} \\
\end{tabular}%
}
\caption{Representative methods for text-to-image generation using their official configurations, with FLUX.1-dev as the shared backbone.}
\label{fig:sparse-t2i-qualitative}
\end{figure*}

%% file: sections/05_qualitative_video.tex
\begin{figure*}[!t]
\centering
\footnotesize
\setlength{\tabcolsep}{2pt}
\resizebox{\textwidth}{!}{%
\begin{tabular}{@{}lccc@{\hspace{8pt}}ccc@{}}
\raisebox{-.5\height}{\textbf{Wan2.1 14B Dense}~\cite{wanteam2025wan}} & \raisebox{-.5\height}{\includegraphics[width=.138\textwidth,keepaspectratio]{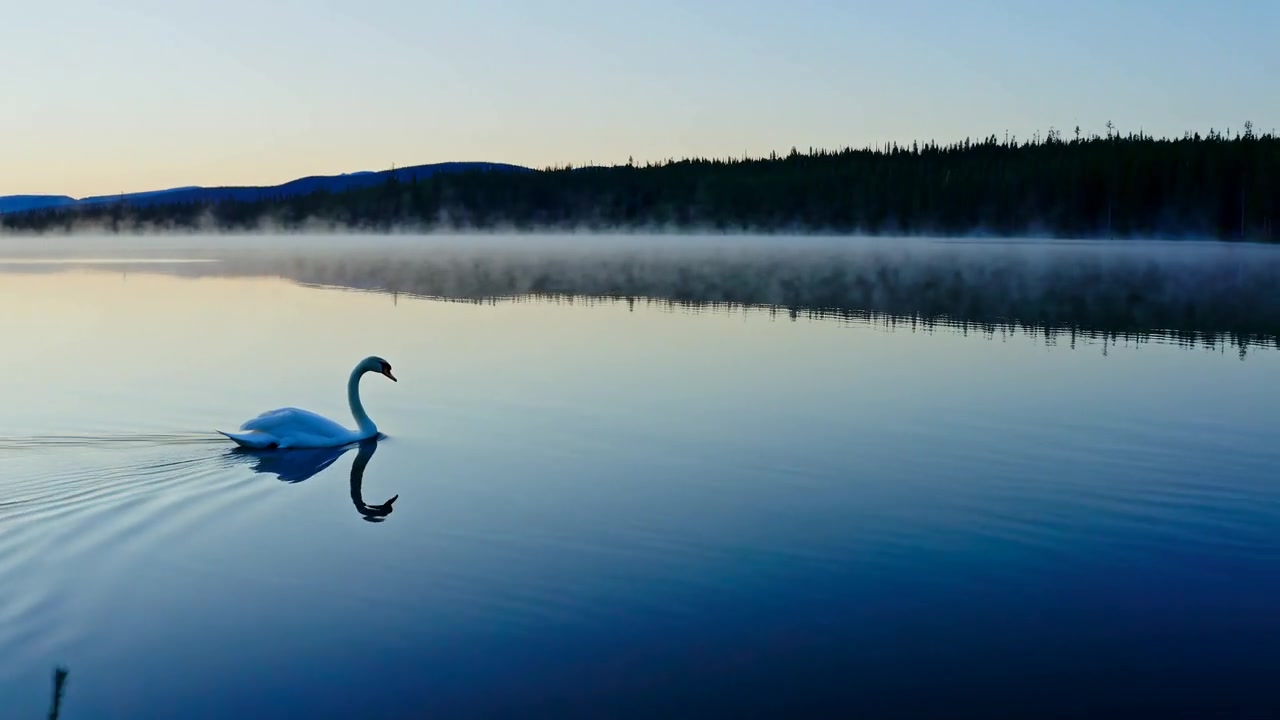}} & \raisebox{-.5\height}{\includegraphics[width=.138\textwidth,keepaspectratio]{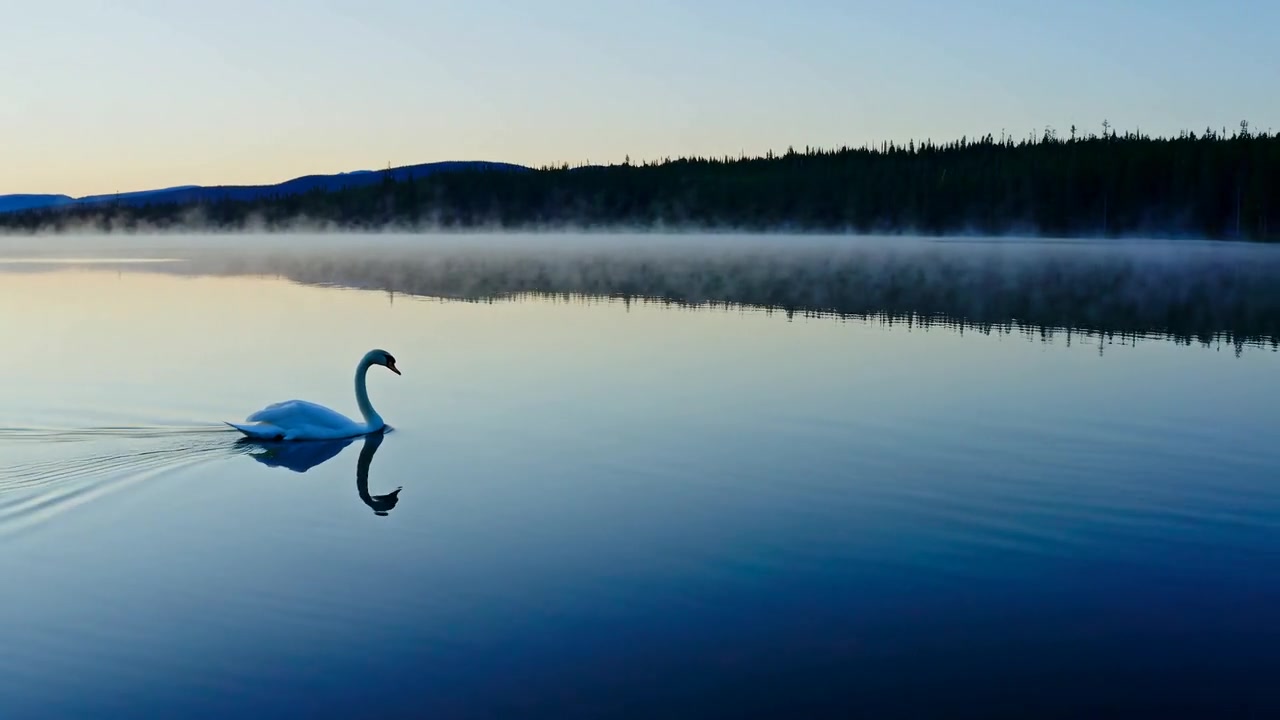}} & \raisebox{-.5\height}{\includegraphics[width=.138\textwidth,keepaspectratio]{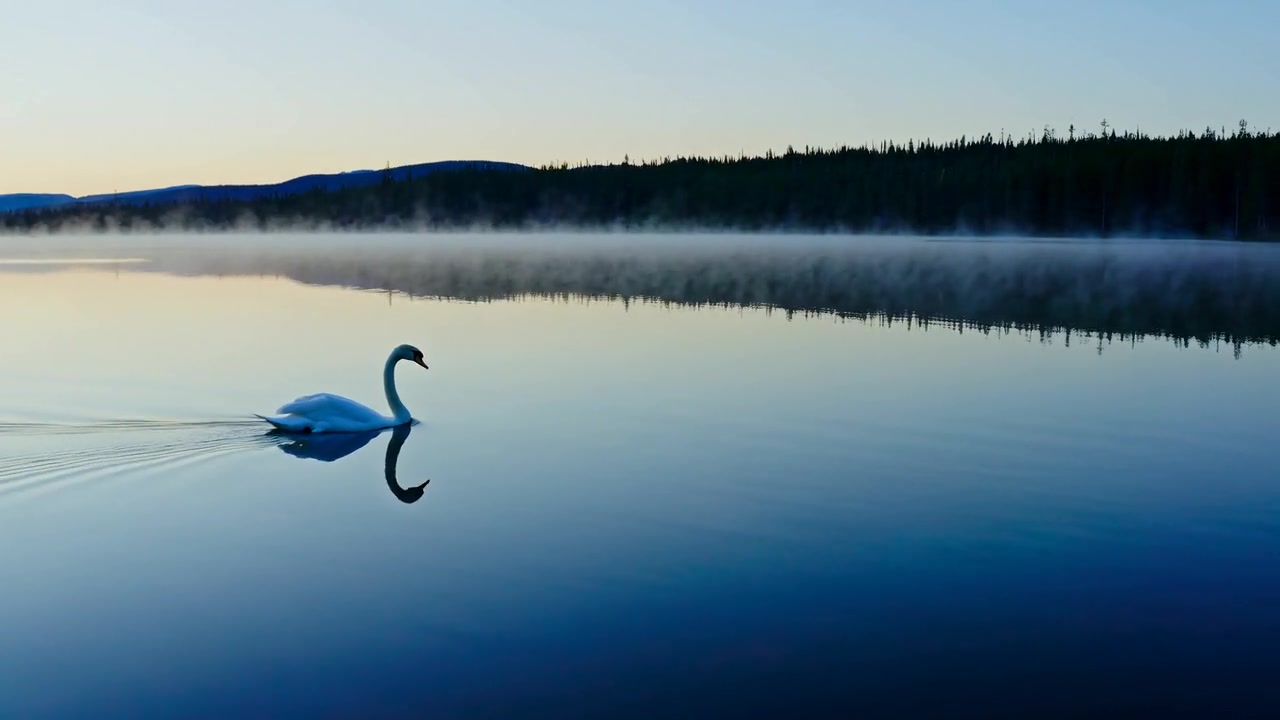}} & \raisebox{-.5\height}{\includegraphics[width=.138\textwidth,keepaspectratio]{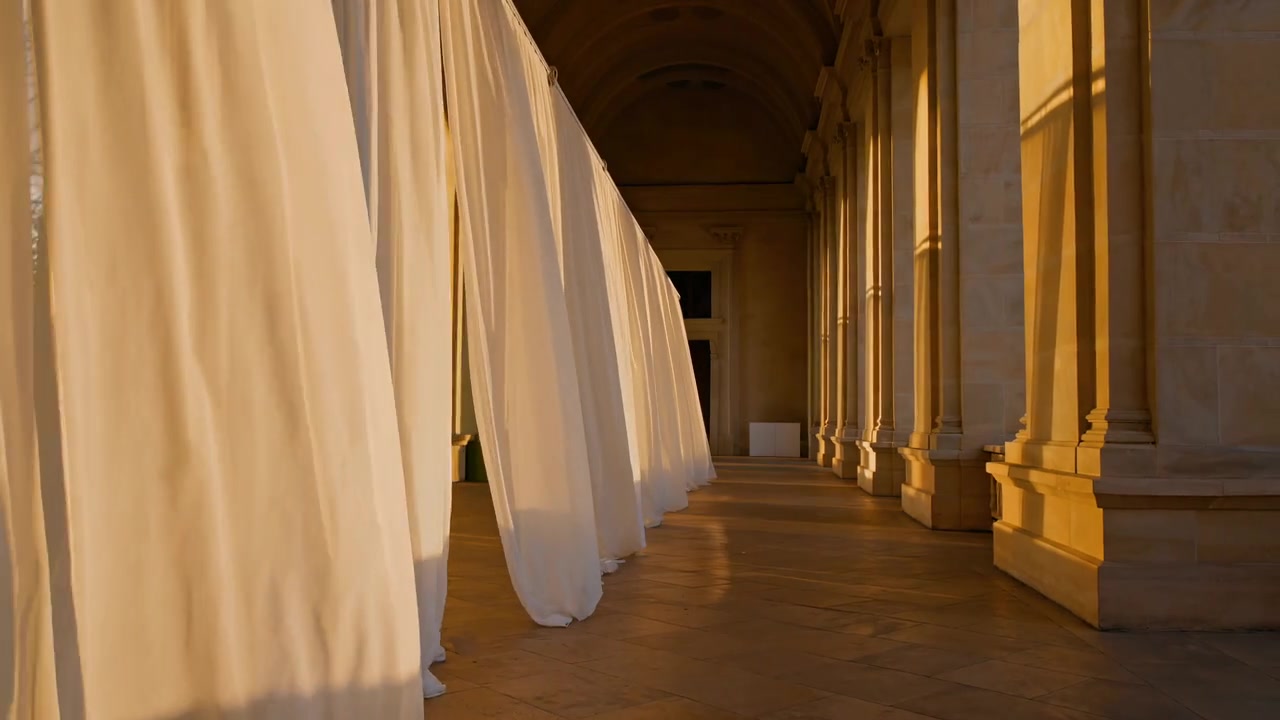}} & \raisebox{-.5\height}{\includegraphics[width=.138\textwidth,keepaspectratio]{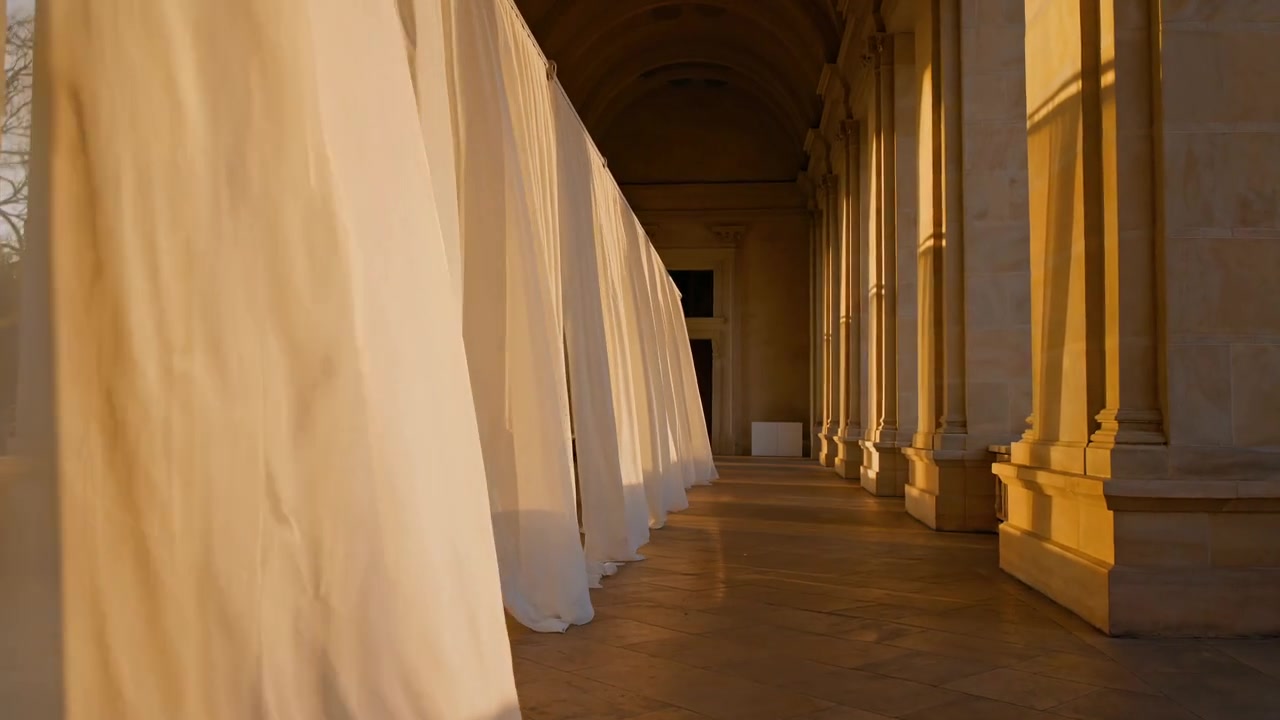}} & \raisebox{-.5\height}{\includegraphics[width=.138\textwidth,keepaspectratio]{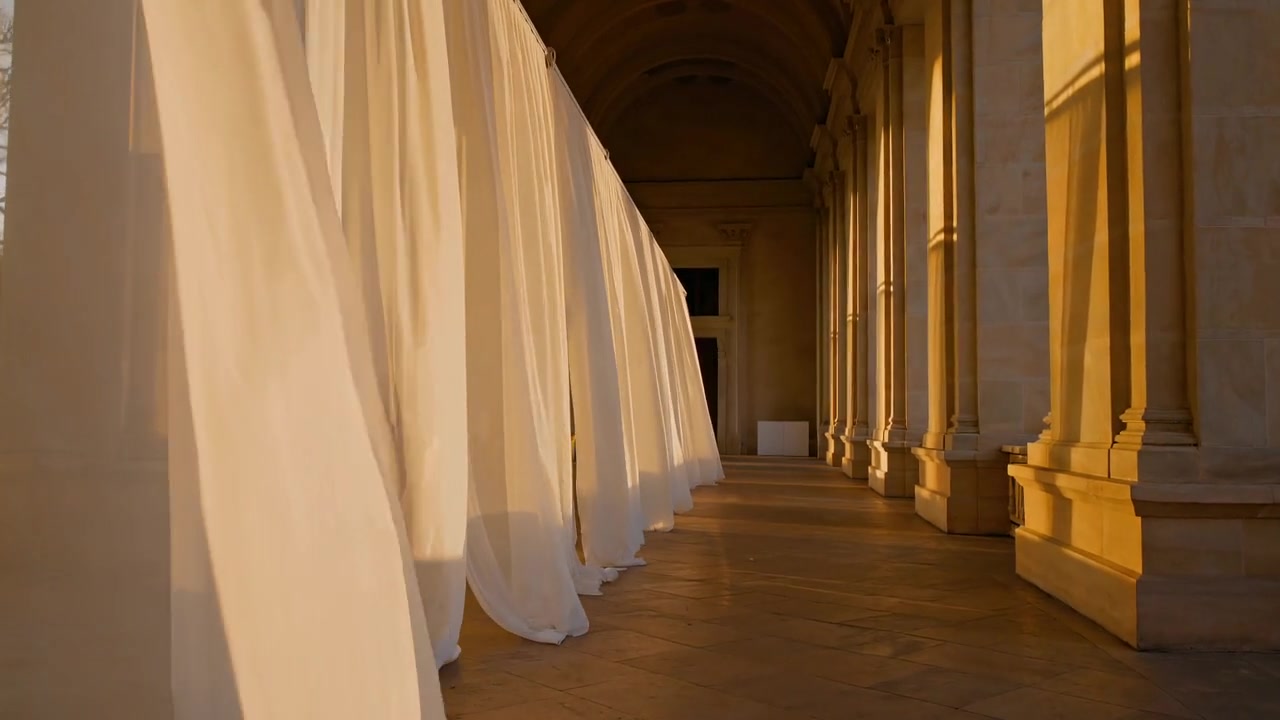}} \\\noalign{\vskip 2pt}
\raisebox{-.5\height}{Sparse VideoGen~\cite{p250201776}} & \raisebox{-.5\height}{\includegraphics[width=.138\textwidth,keepaspectratio]{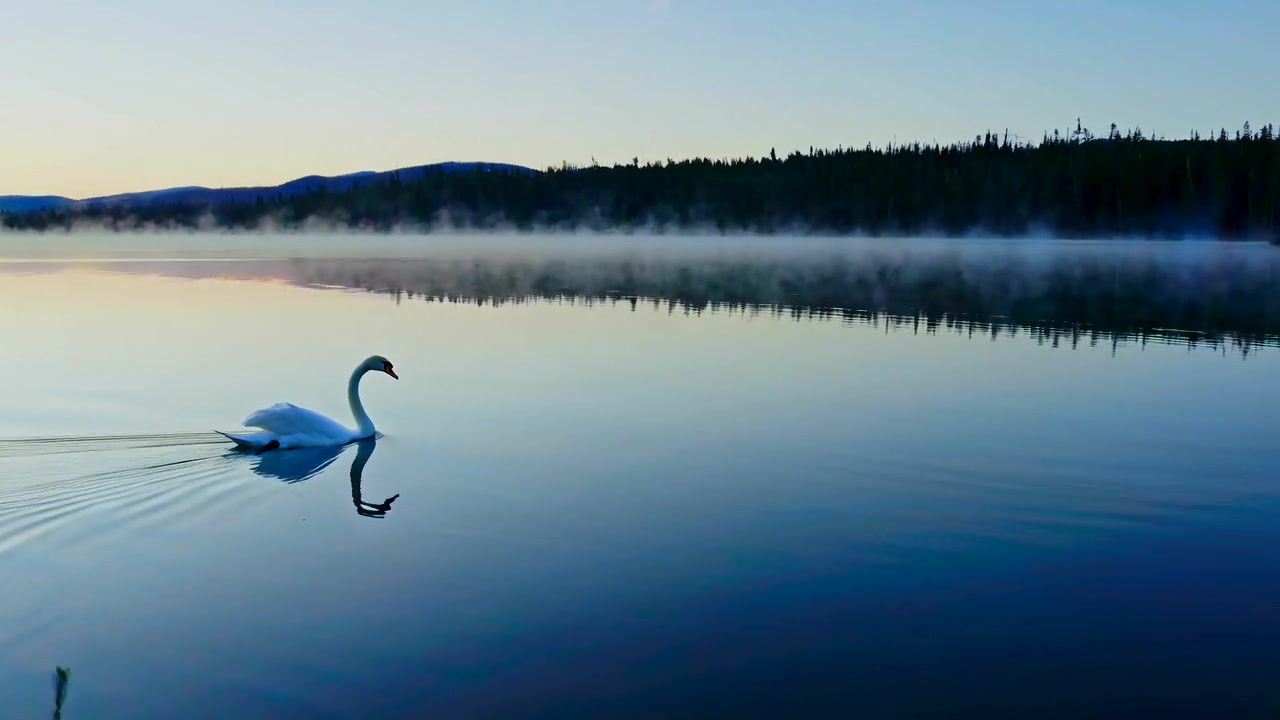}} & \raisebox{-.5\height}{\includegraphics[width=.138\textwidth,keepaspectratio]{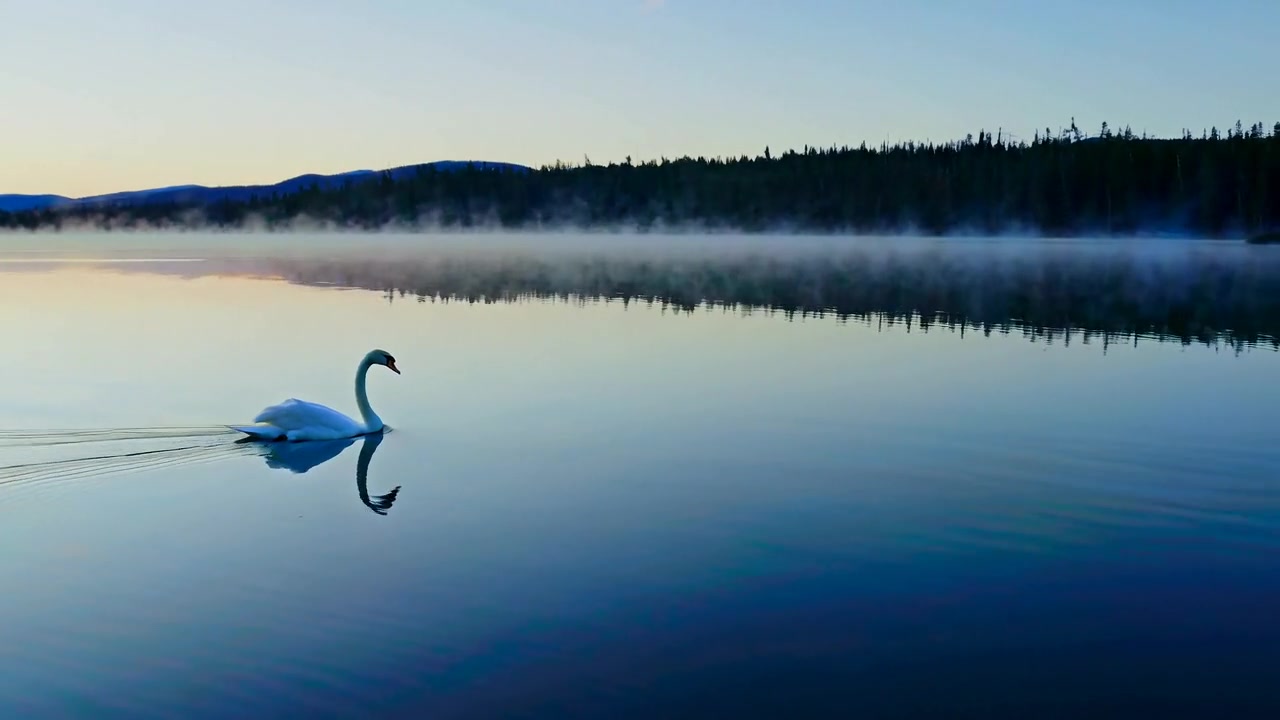}} & \raisebox{-.5\height}{\includegraphics[width=.138\textwidth,keepaspectratio]{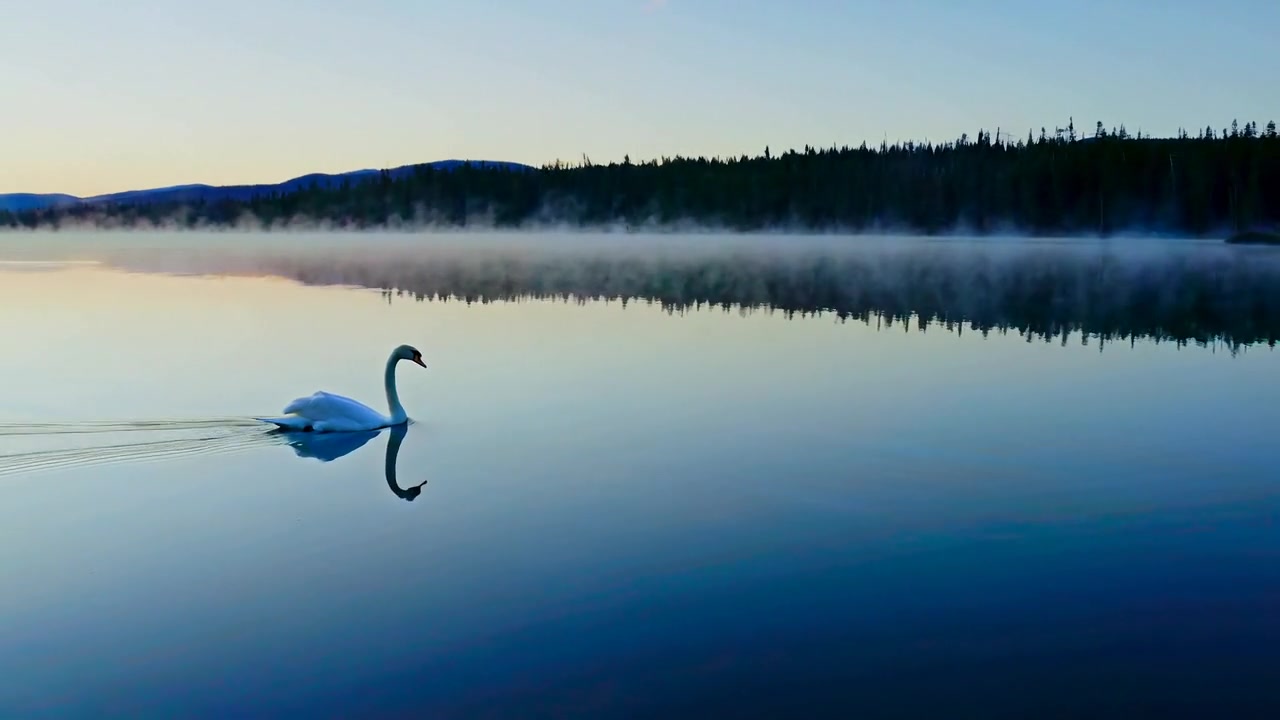}} & \raisebox{-.5\height}{\includegraphics[width=.138\textwidth,keepaspectratio]{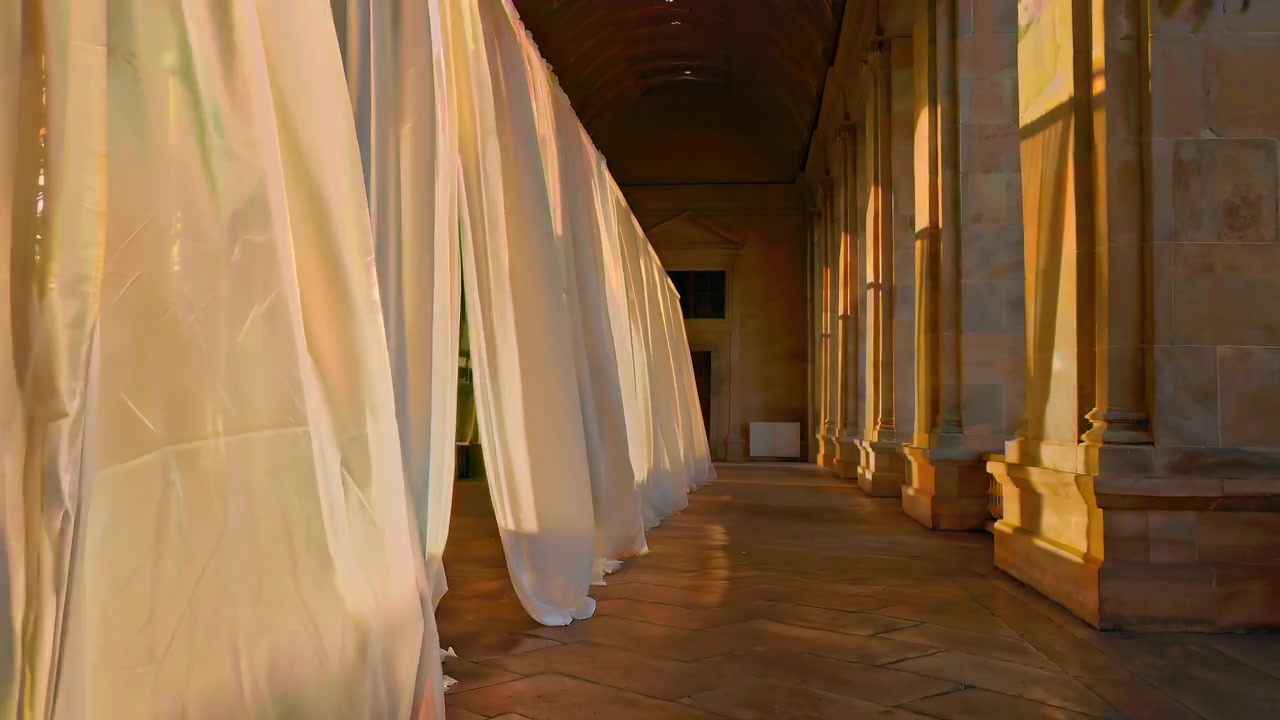}} & \raisebox{-.5\height}{\includegraphics[width=.138\textwidth,keepaspectratio]{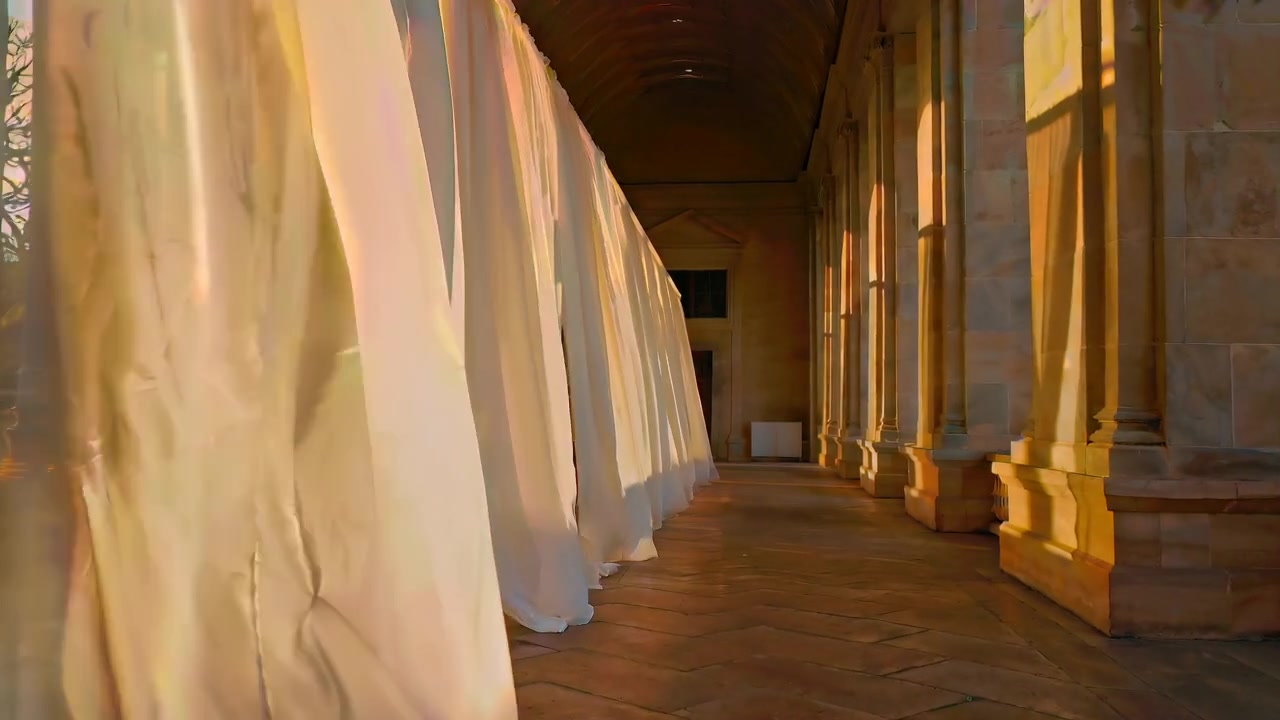}} & \raisebox{-.5\height}{\includegraphics[width=.138\textwidth,keepaspectratio]{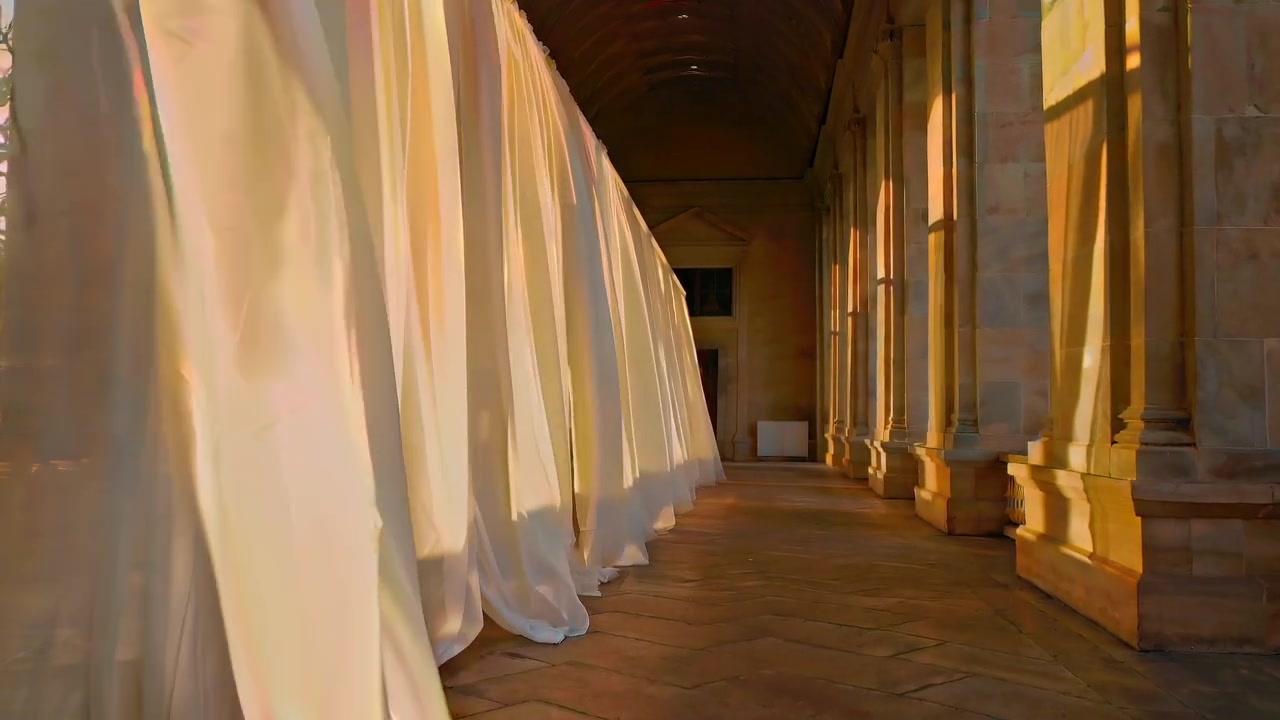}} \\\noalign{\vskip 2pt}
\raisebox{-.5\height}{SVG-EAR~\cite{p260308982}} & \raisebox{-.5\height}{\includegraphics[width=.138\textwidth,keepaspectratio]{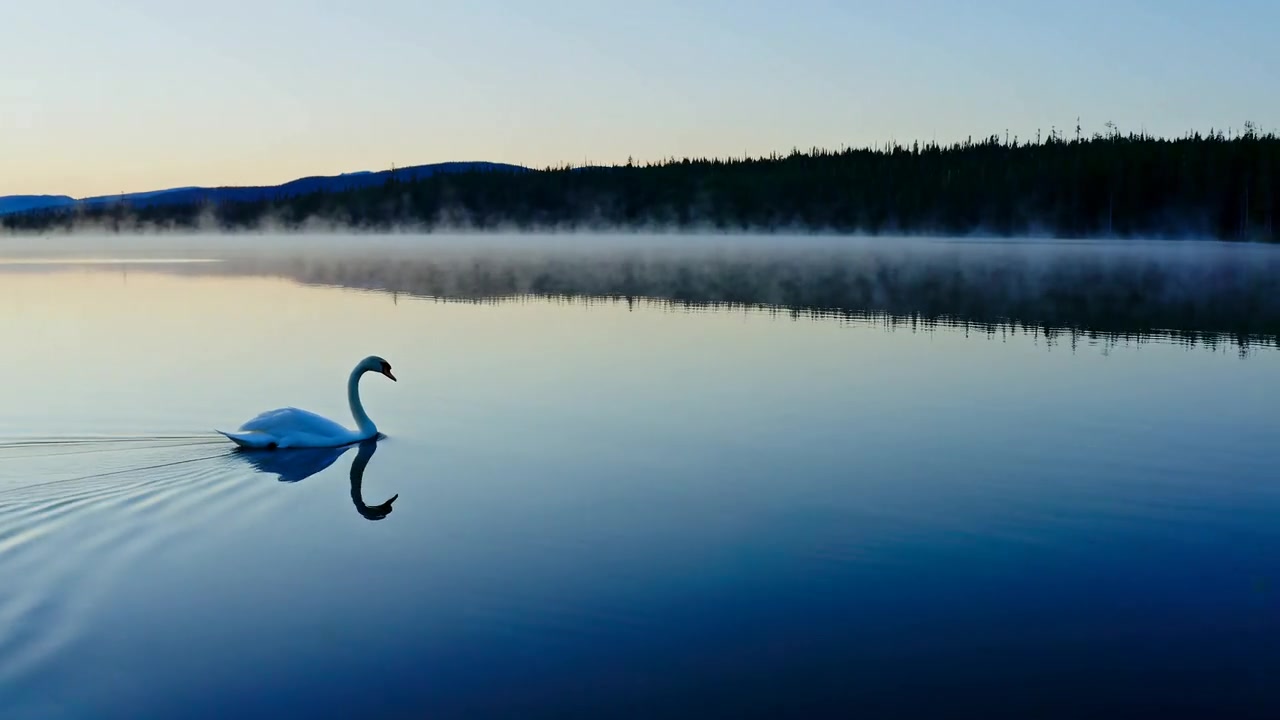}} & \raisebox{-.5\height}{\includegraphics[width=.138\textwidth,keepaspectratio]{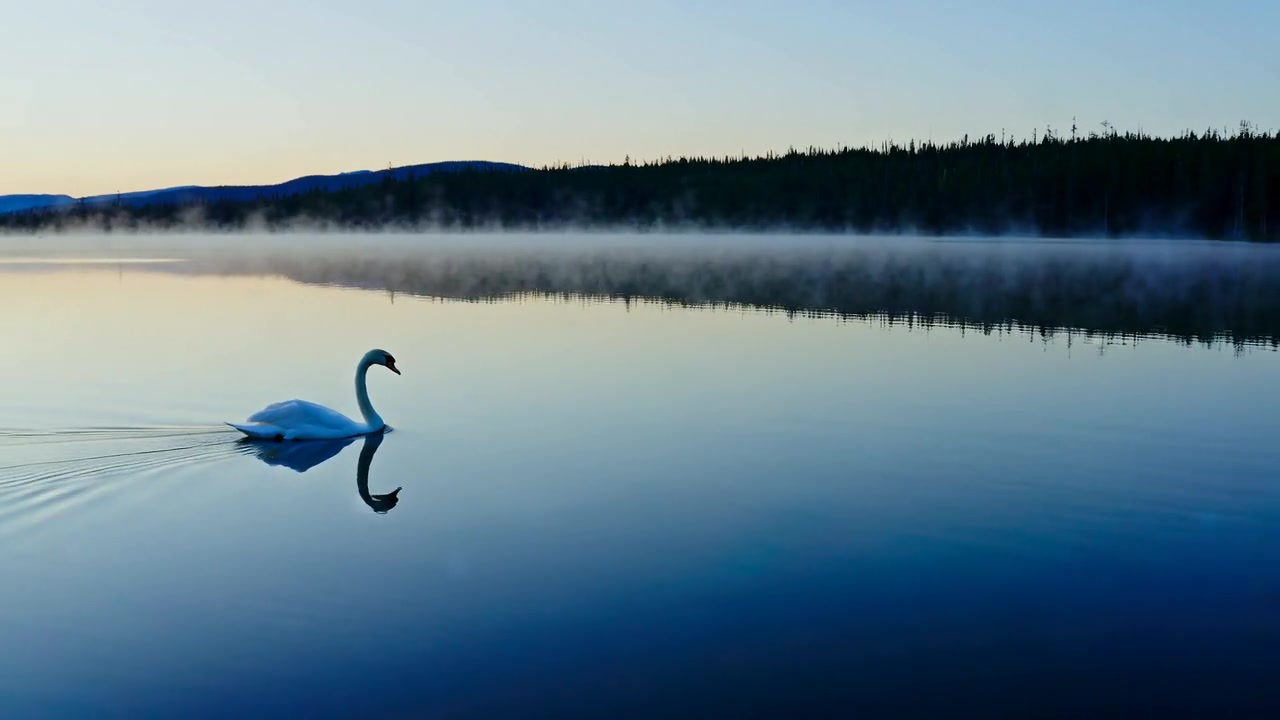}} & \raisebox{-.5\height}{\includegraphics[width=.138\textwidth,keepaspectratio]{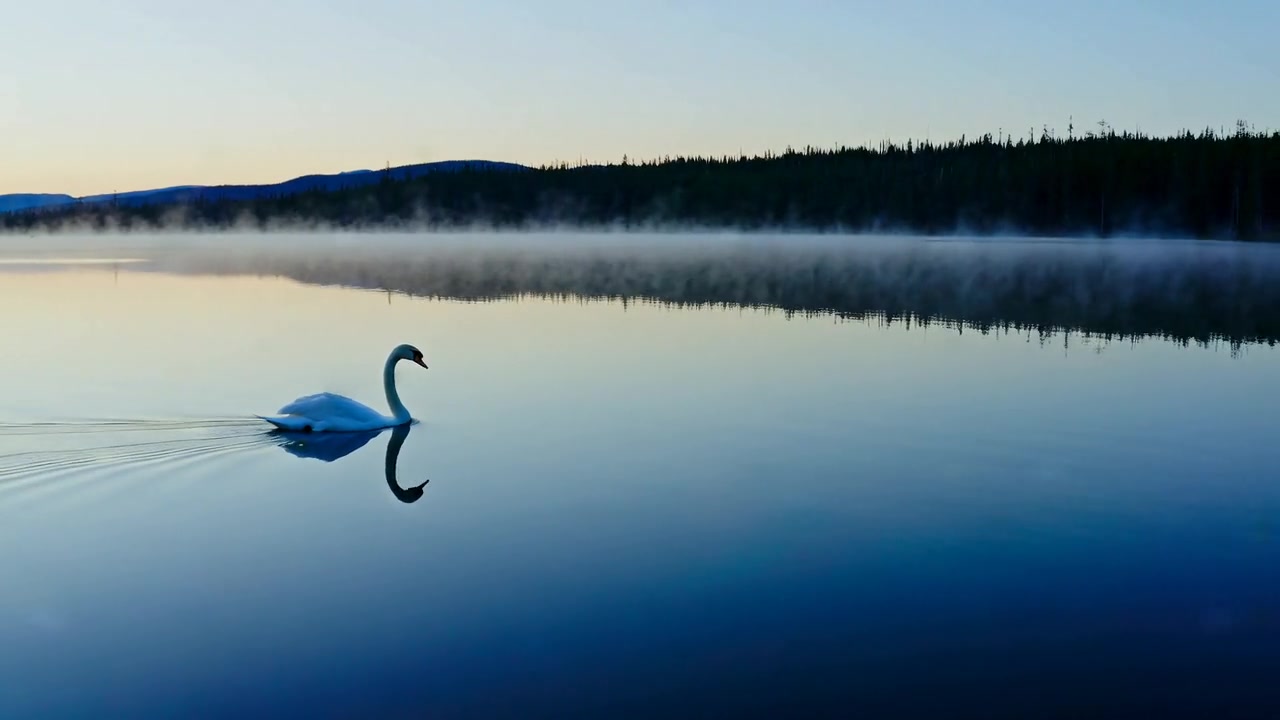}} & \raisebox{-.5\height}{\includegraphics[width=.138\textwidth,keepaspectratio]{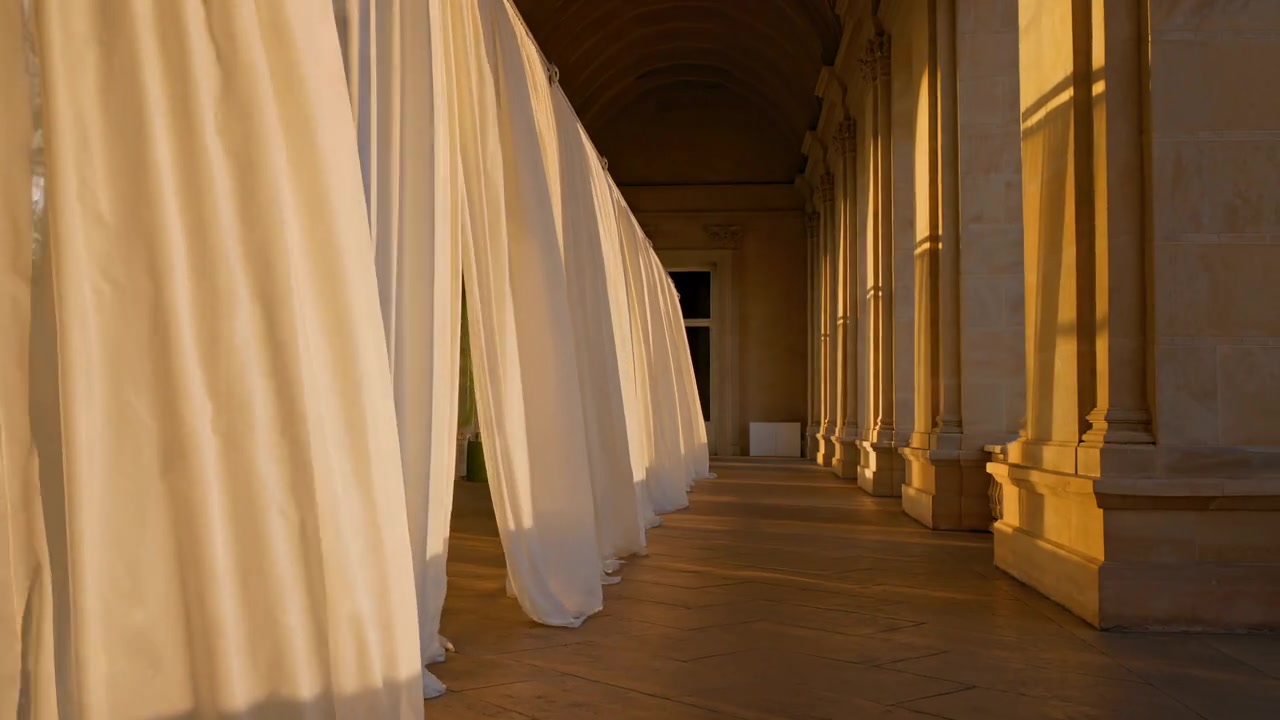}} & \raisebox{-.5\height}{\includegraphics[width=.138\textwidth,keepaspectratio]{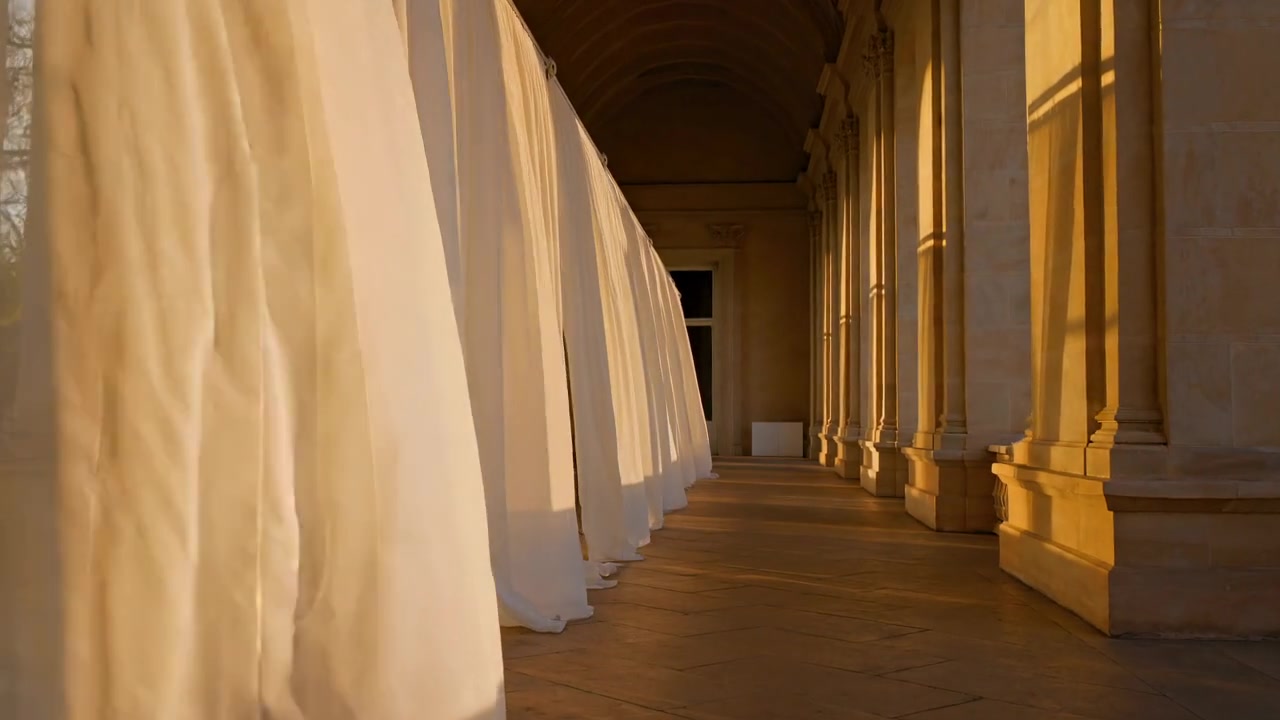}} & \raisebox{-.5\height}{\includegraphics[width=.138\textwidth,keepaspectratio]{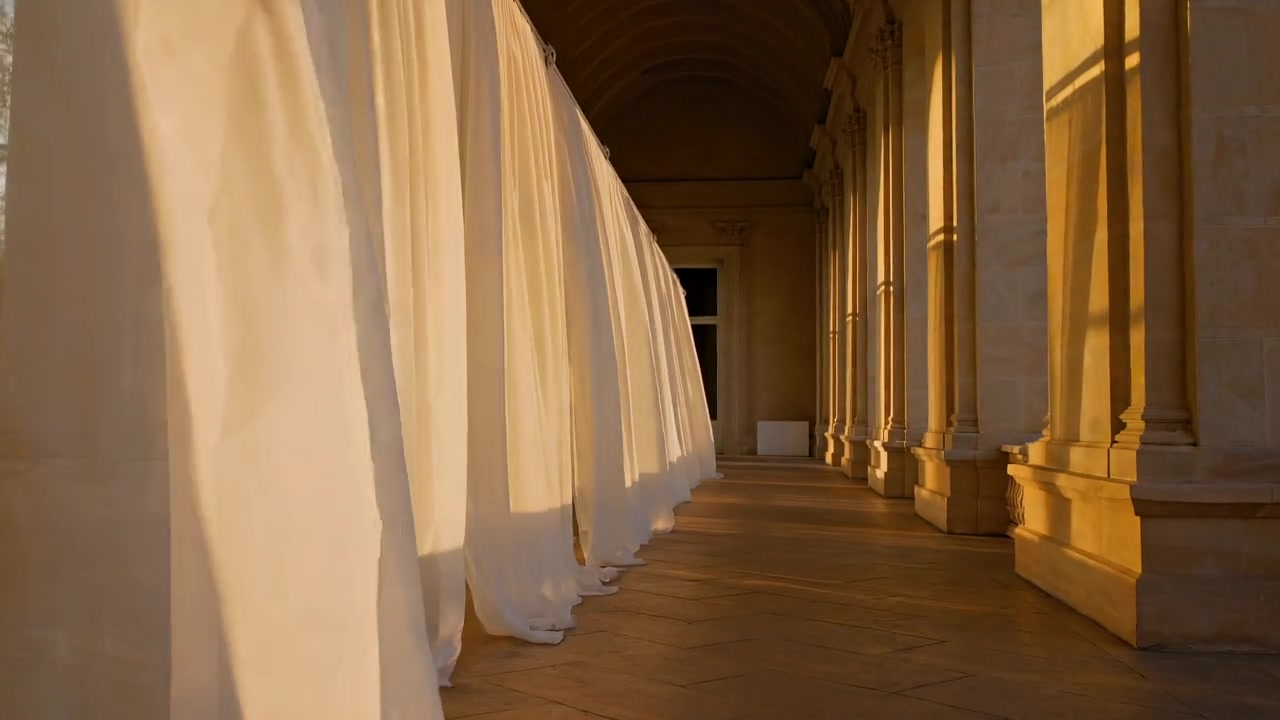}} \\\noalign{\vskip 2pt}
\raisebox{-.5\height}{SVOO~\cite{p260318636}} & \raisebox{-.5\height}{\includegraphics[width=.138\textwidth,keepaspectratio]{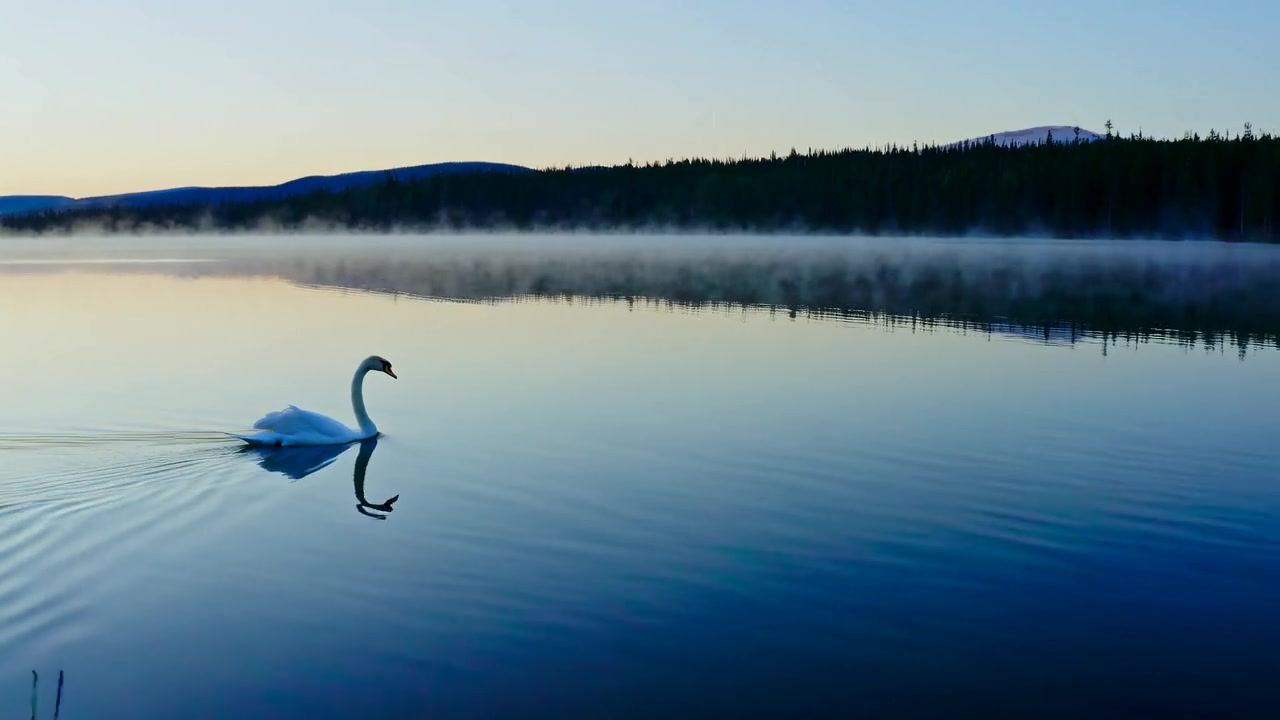}} & \raisebox{-.5\height}{\includegraphics[width=.138\textwidth,keepaspectratio]{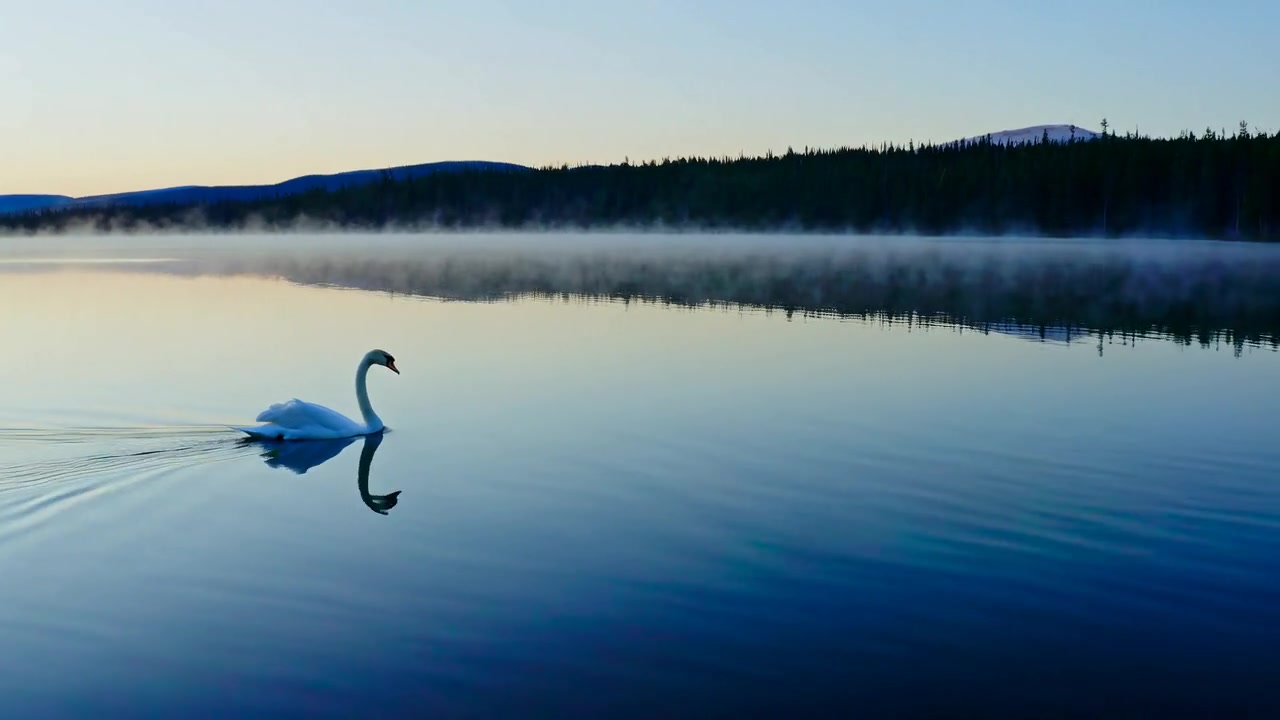}} & \raisebox{-.5\height}{\includegraphics[width=.138\textwidth,keepaspectratio]{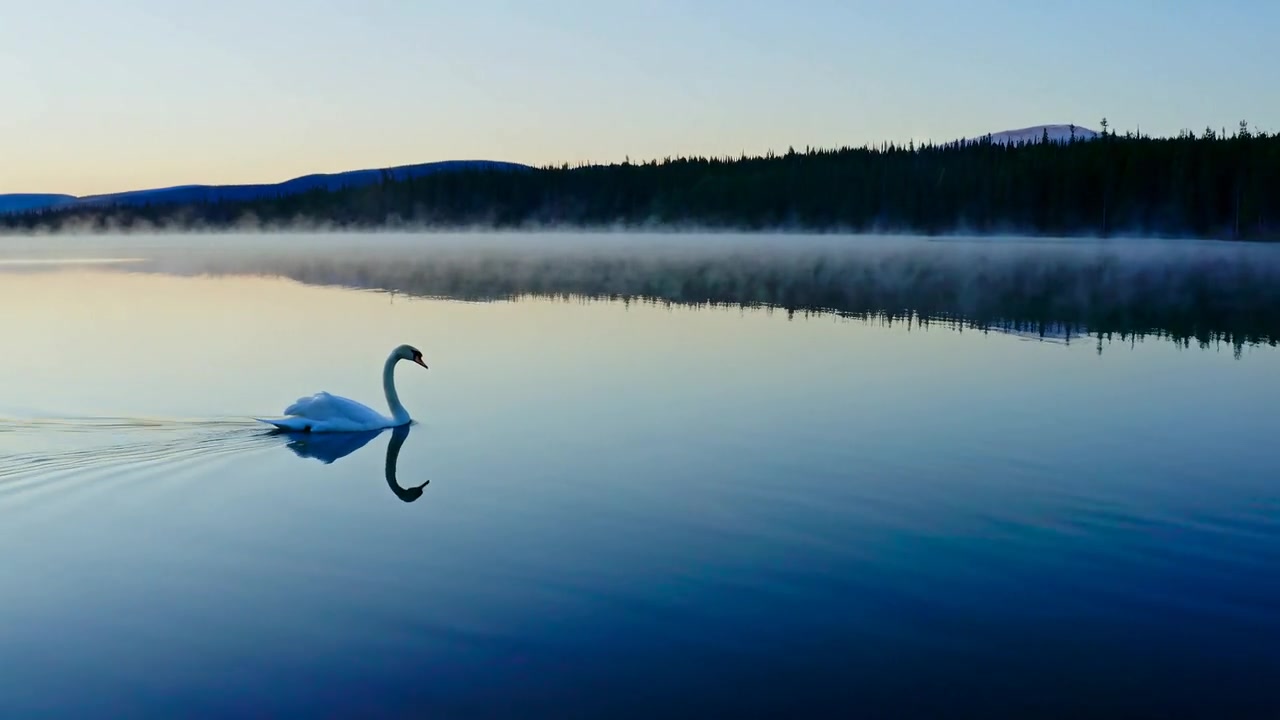}} & \raisebox{-.5\height}{\includegraphics[width=.138\textwidth,keepaspectratio]{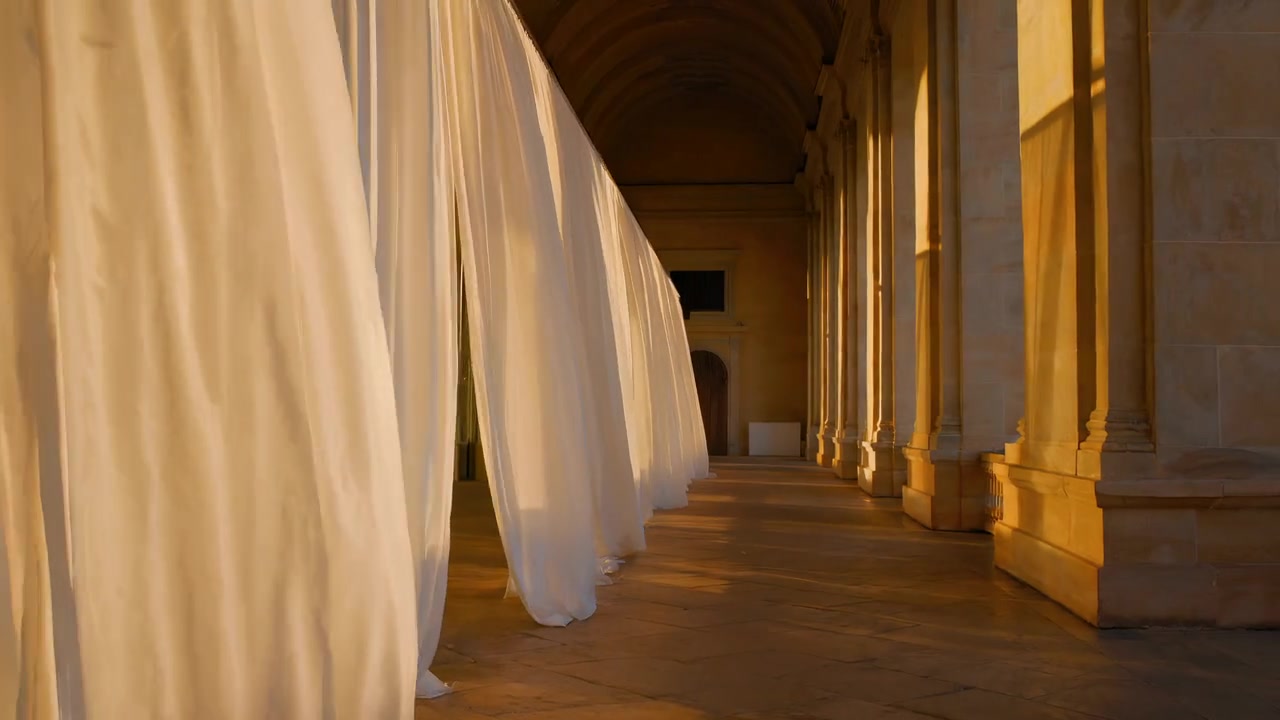}} & \raisebox{-.5\height}{\includegraphics[width=.138\textwidth,keepaspectratio]{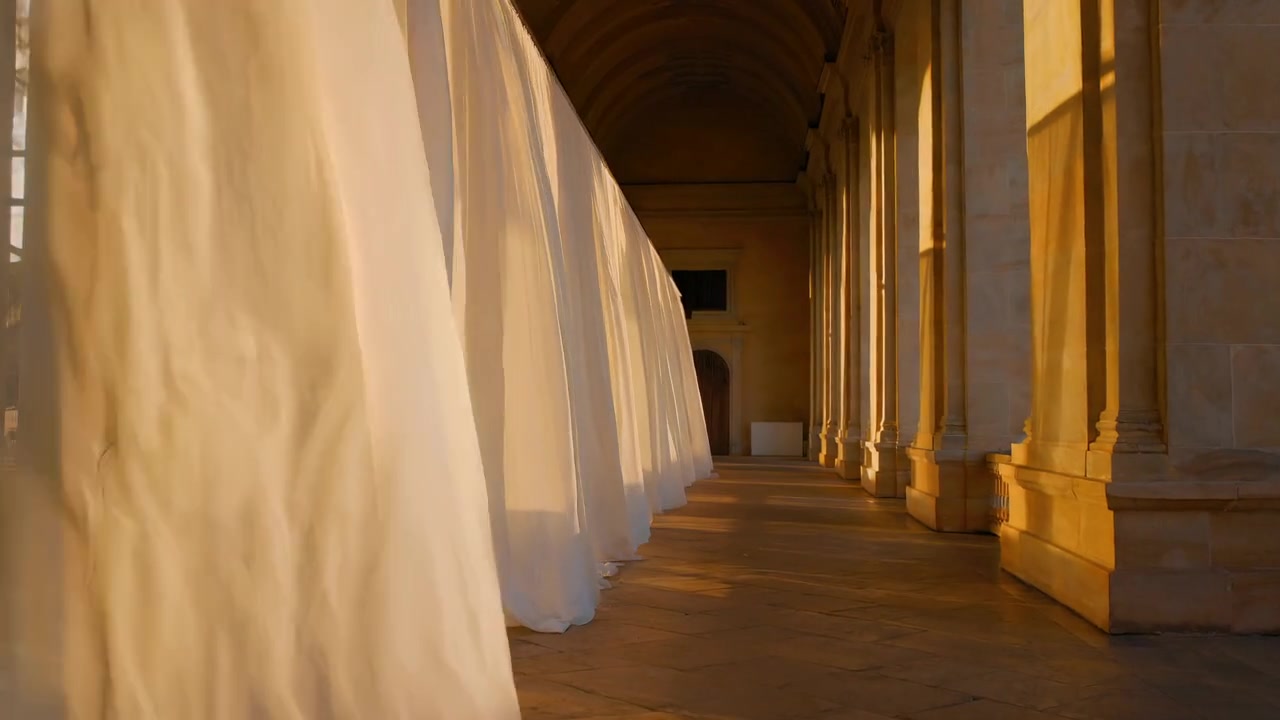}} & \raisebox{-.5\height}{\includegraphics[width=.138\textwidth,keepaspectratio]{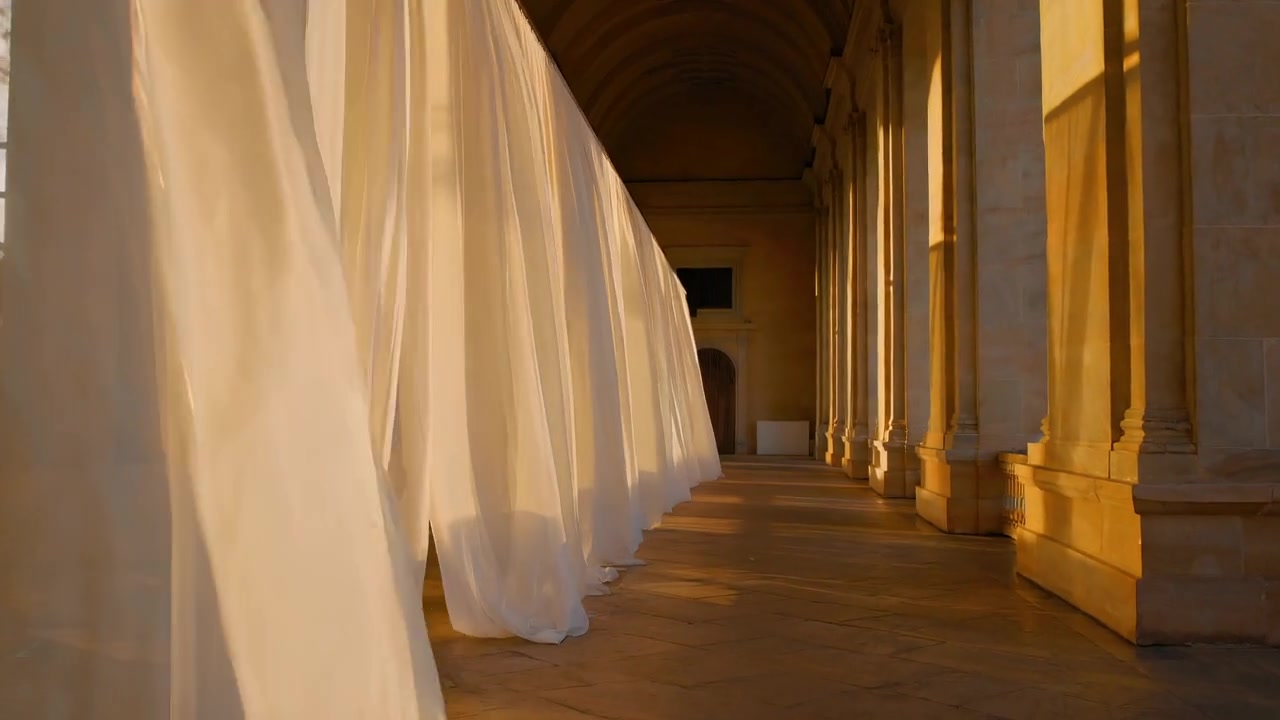}} \\\noalign{\vskip 2pt}
\raisebox{-.5\height}{LVSA~\cite{p260531057}} & \raisebox{-.5\height}{\includegraphics[width=.138\textwidth,keepaspectratio]{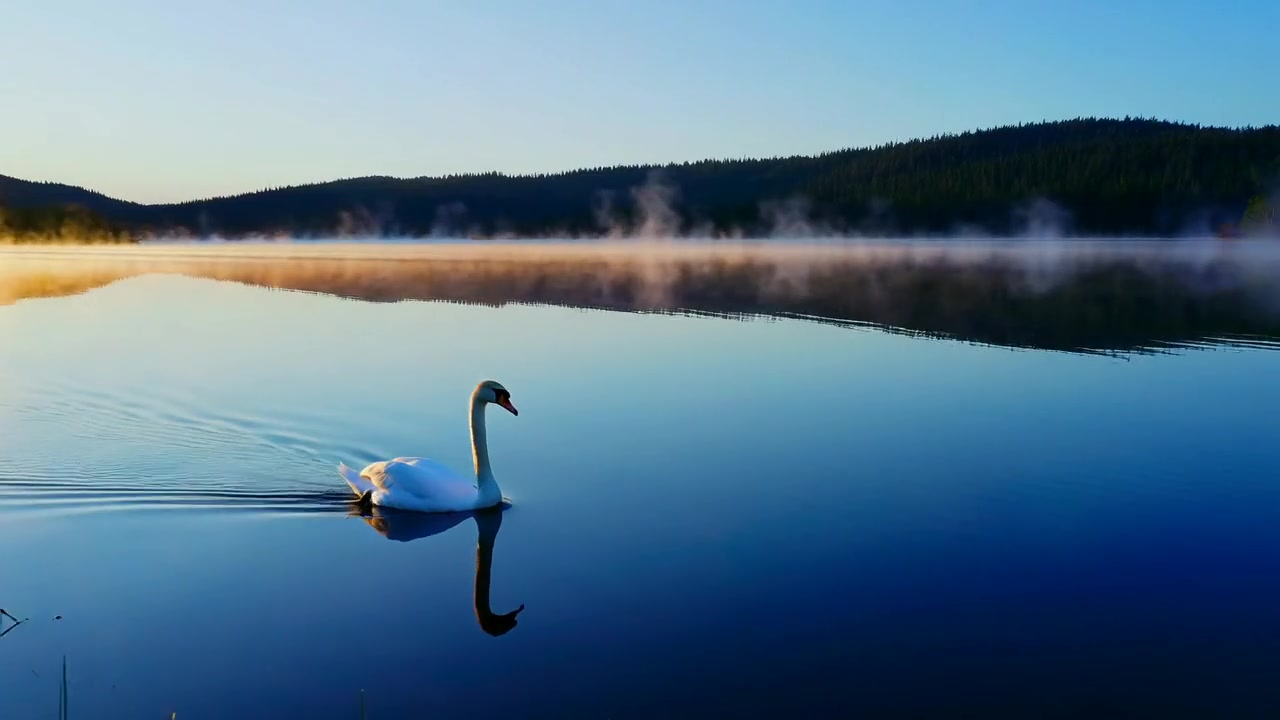}} & \raisebox{-.5\height}{\includegraphics[width=.138\textwidth,keepaspectratio]{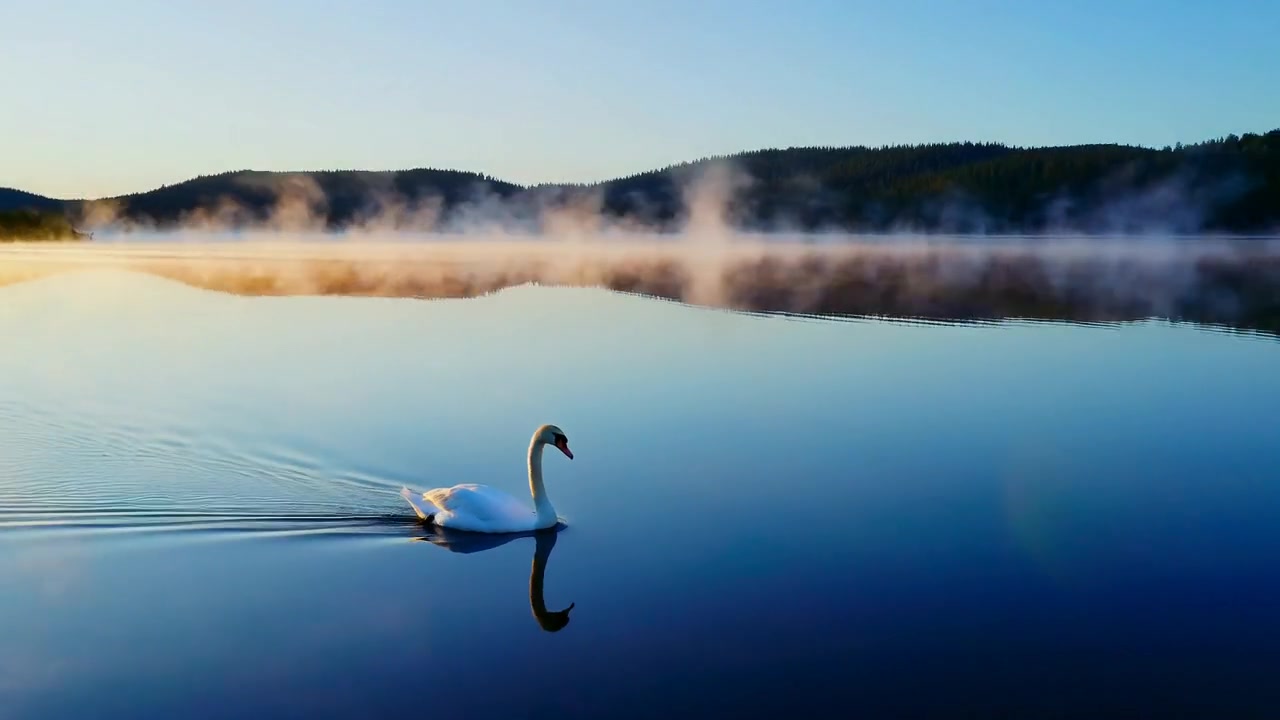}} & \raisebox{-.5\height}{\includegraphics[width=.138\textwidth,keepaspectratio]{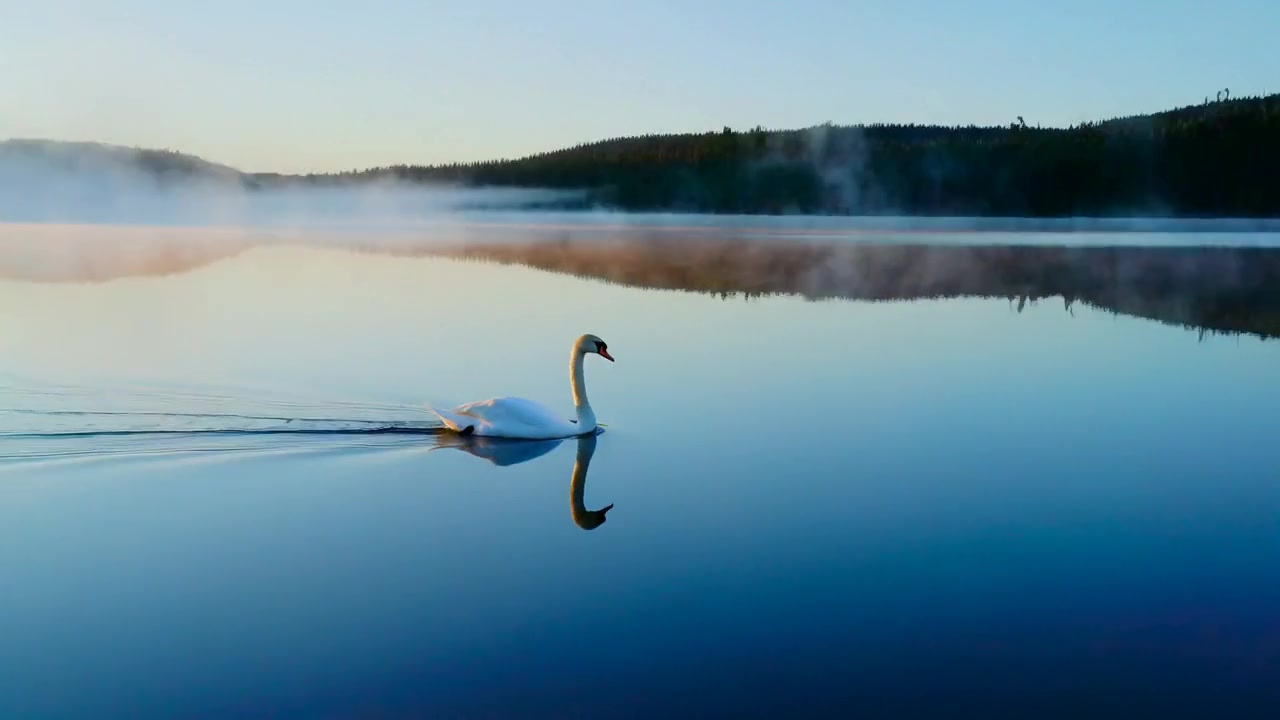}} & \raisebox{-.5\height}{\includegraphics[width=.138\textwidth,keepaspectratio]{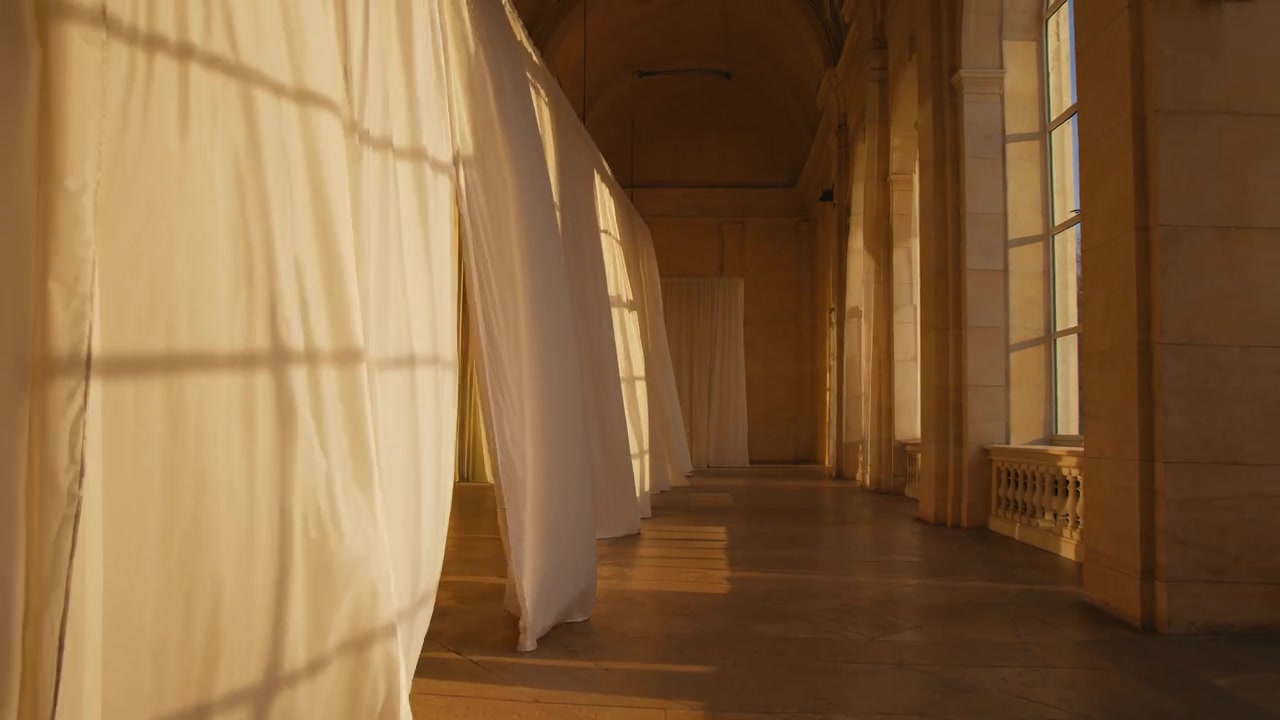}} & \raisebox{-.5\height}{\includegraphics[width=.138\textwidth,keepaspectratio]{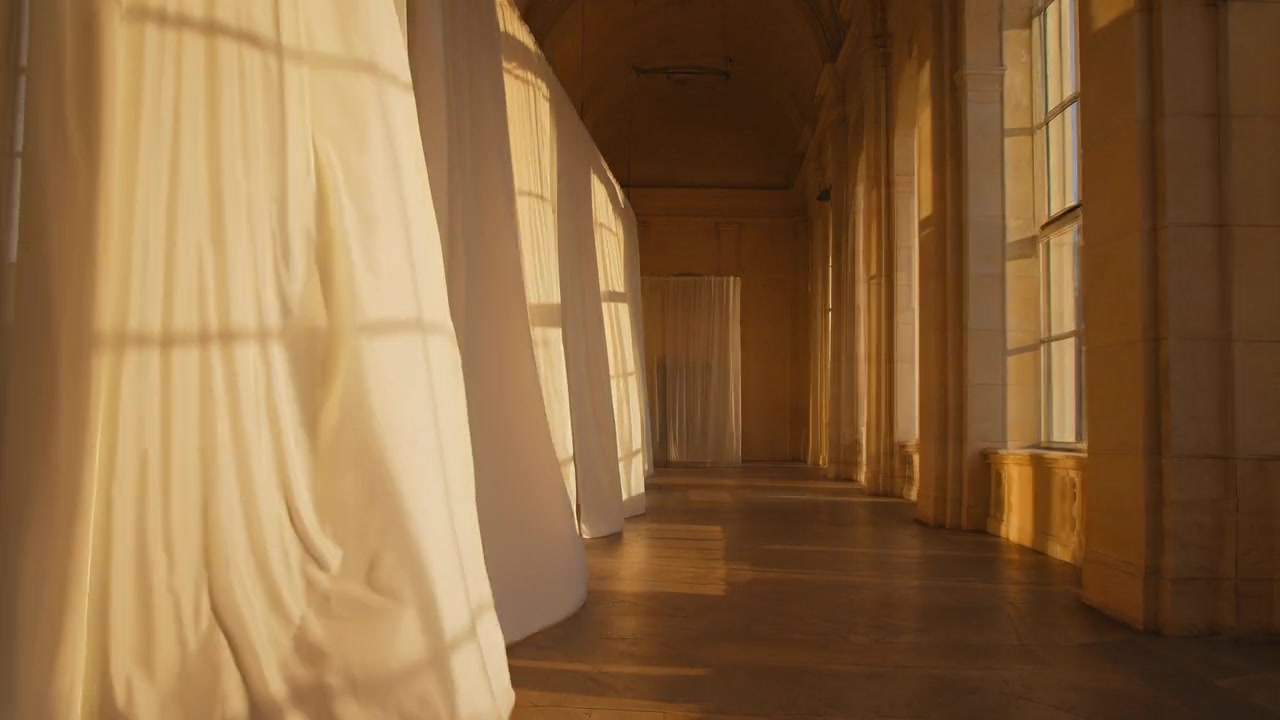}} & \raisebox{-.5\height}{\includegraphics[width=.138\textwidth,keepaspectratio]{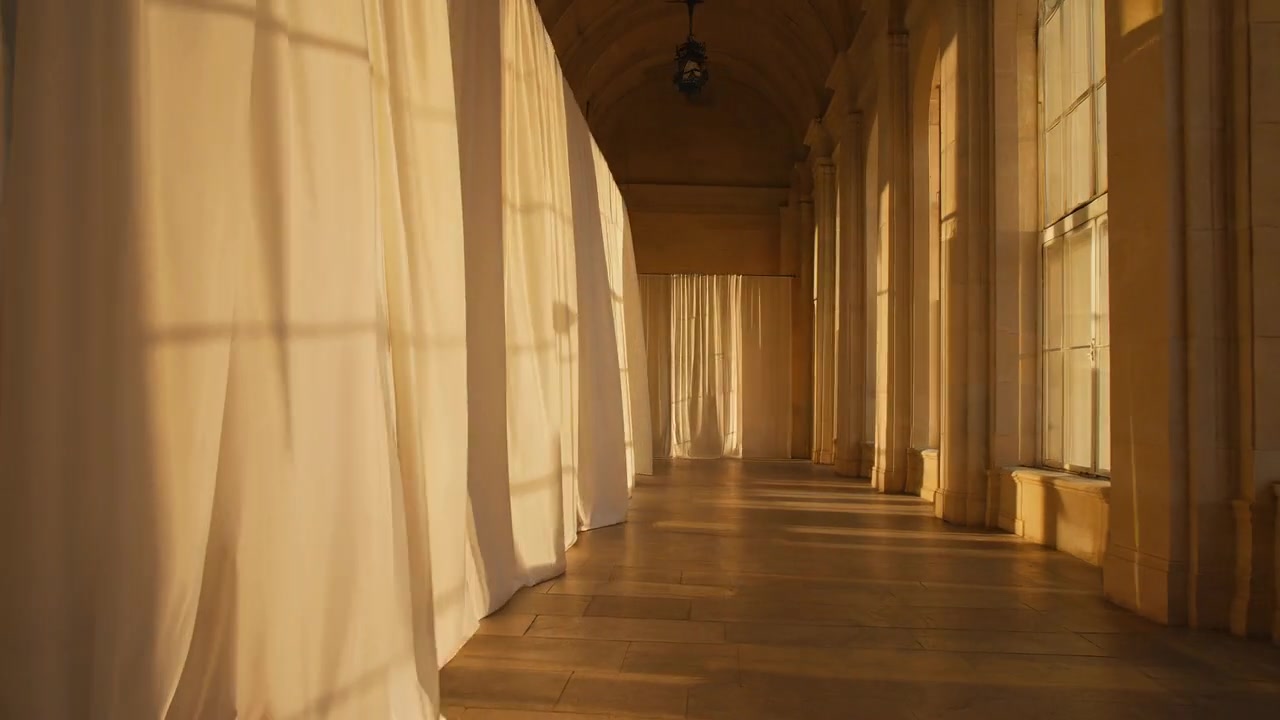}} \\\noalign{\vskip 2pt}
\raisebox{-.5\height}{Sol-Attn~\cite{p260724027}} & \raisebox{-.5\height}{\includegraphics[width=.138\textwidth,keepaspectratio]{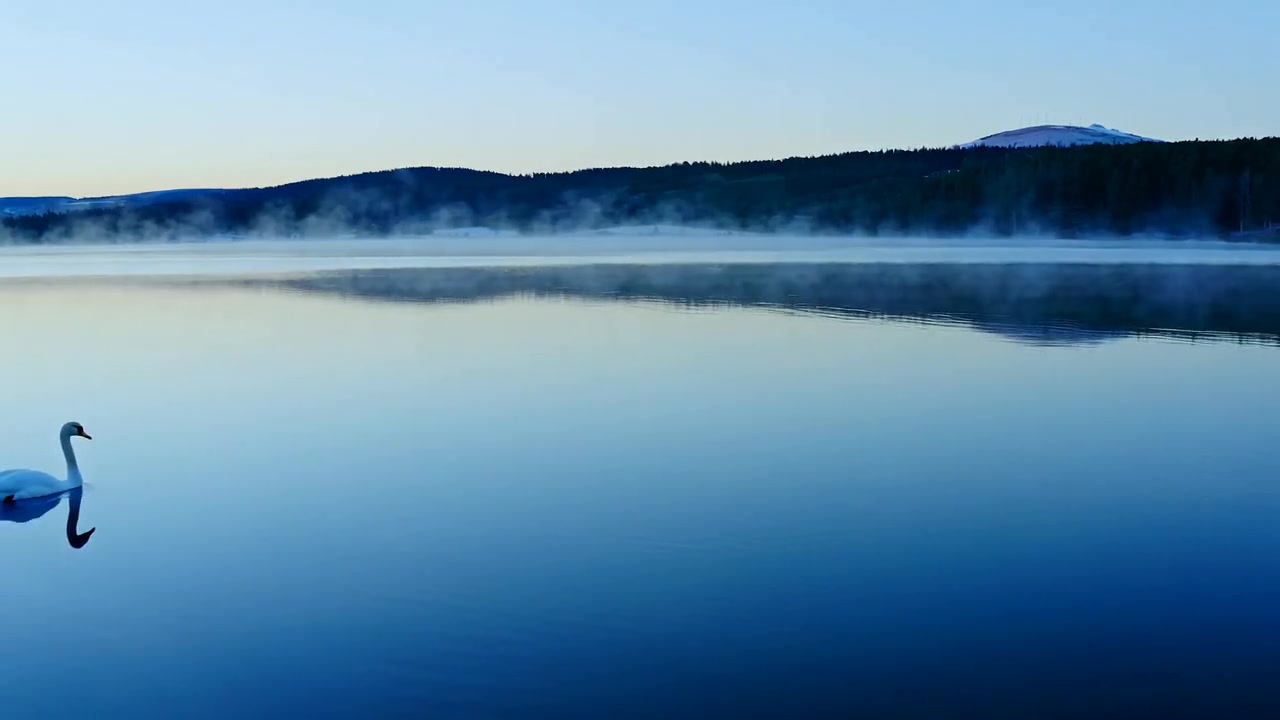}} & \raisebox{-.5\height}{\includegraphics[width=.138\textwidth,keepaspectratio]{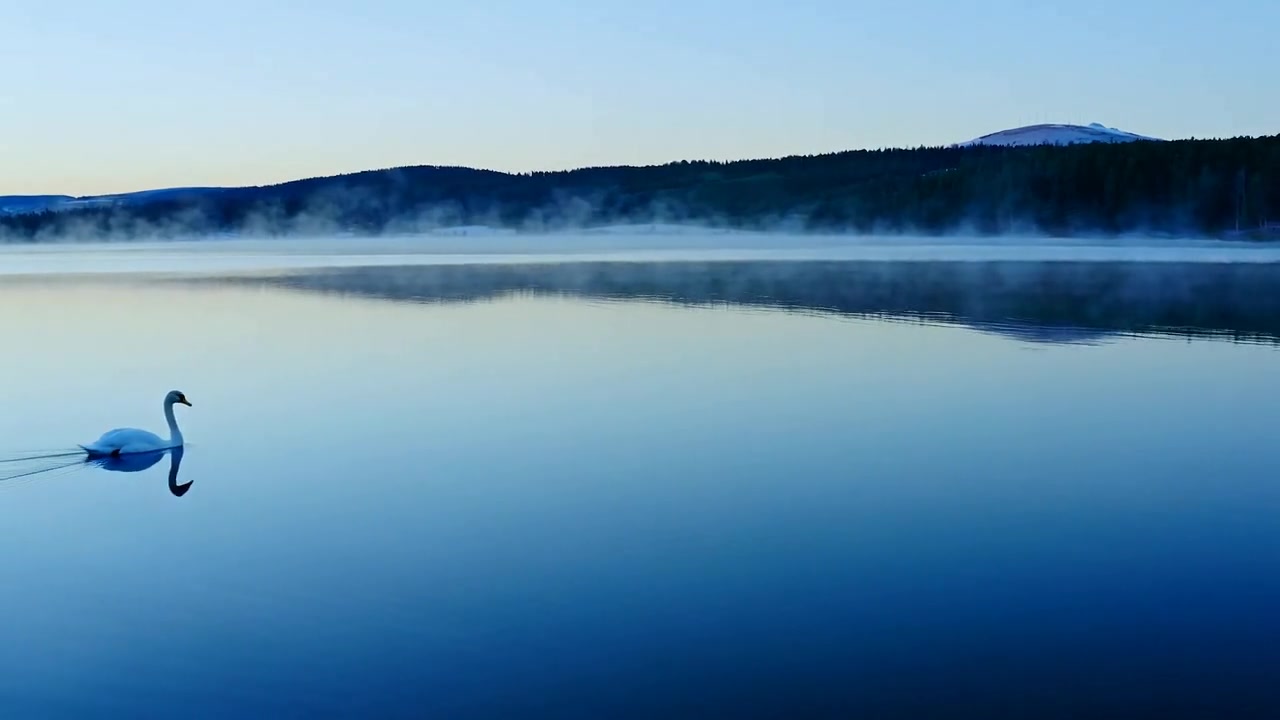}} & \raisebox{-.5\height}{\includegraphics[width=.138\textwidth,keepaspectratio]{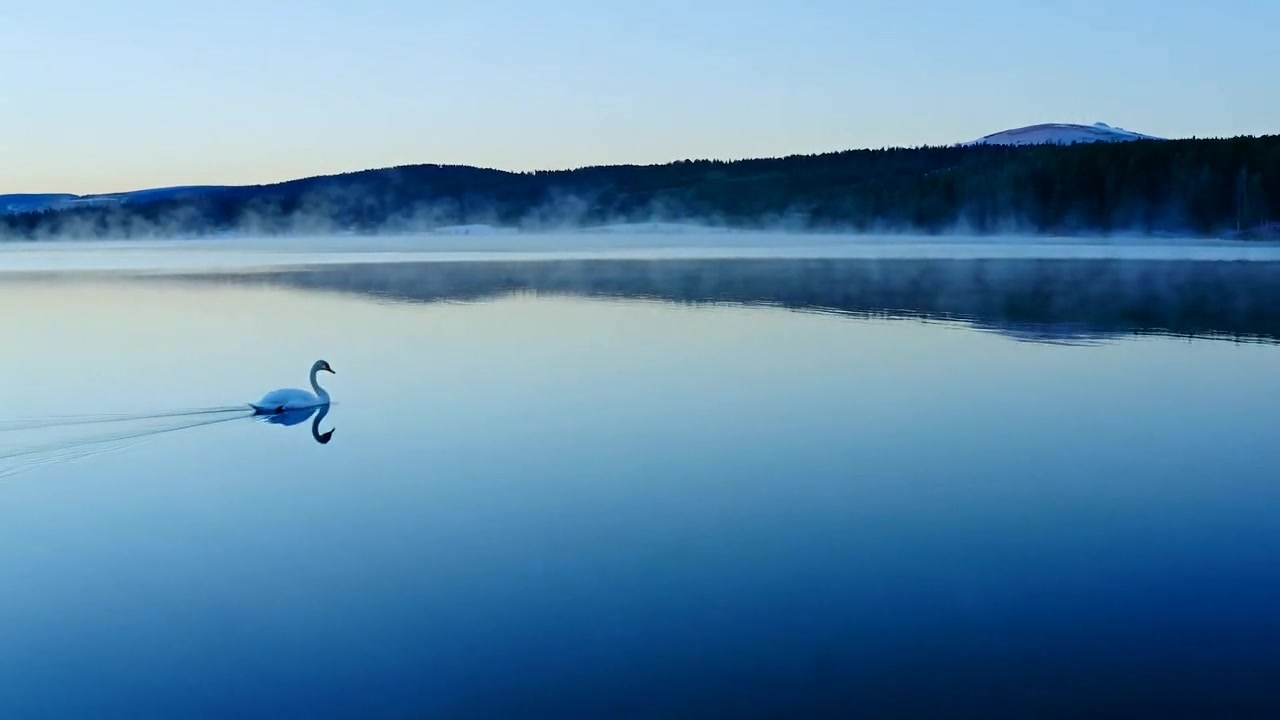}} & \raisebox{-.5\height}{\includegraphics[width=.138\textwidth,keepaspectratio]{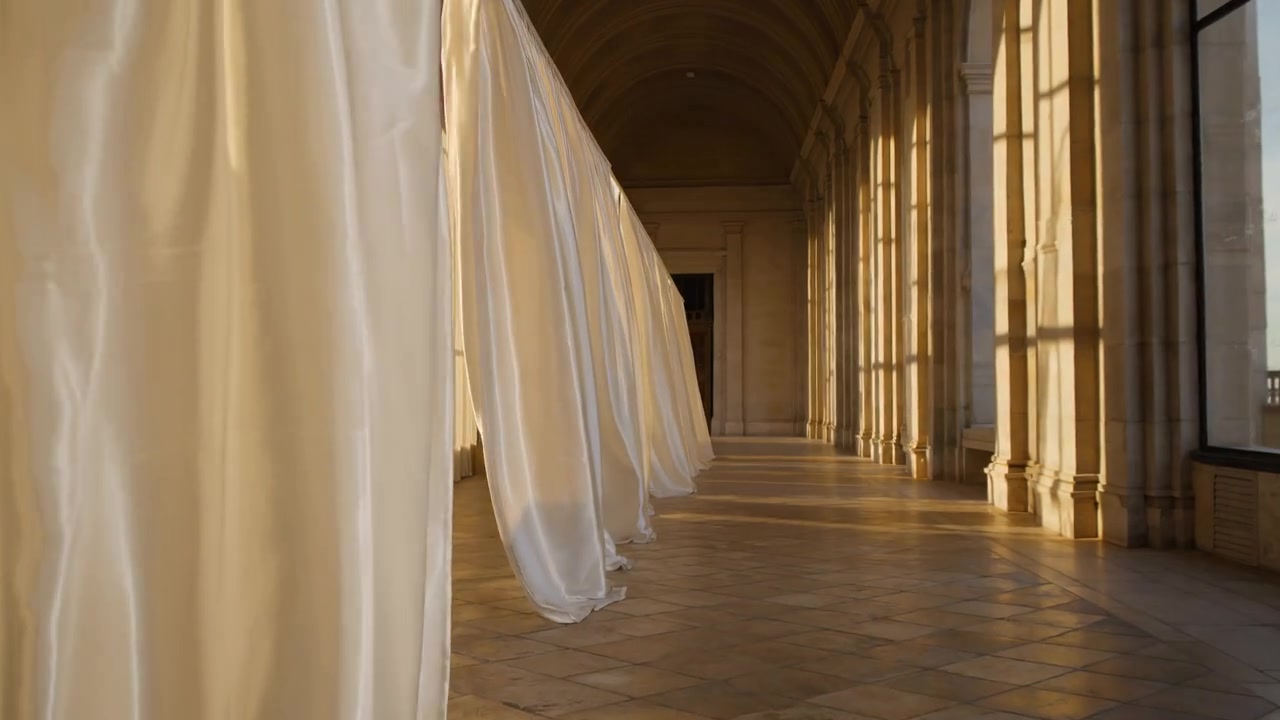}} & \raisebox{-.5\height}{\includegraphics[width=.138\textwidth,keepaspectratio]{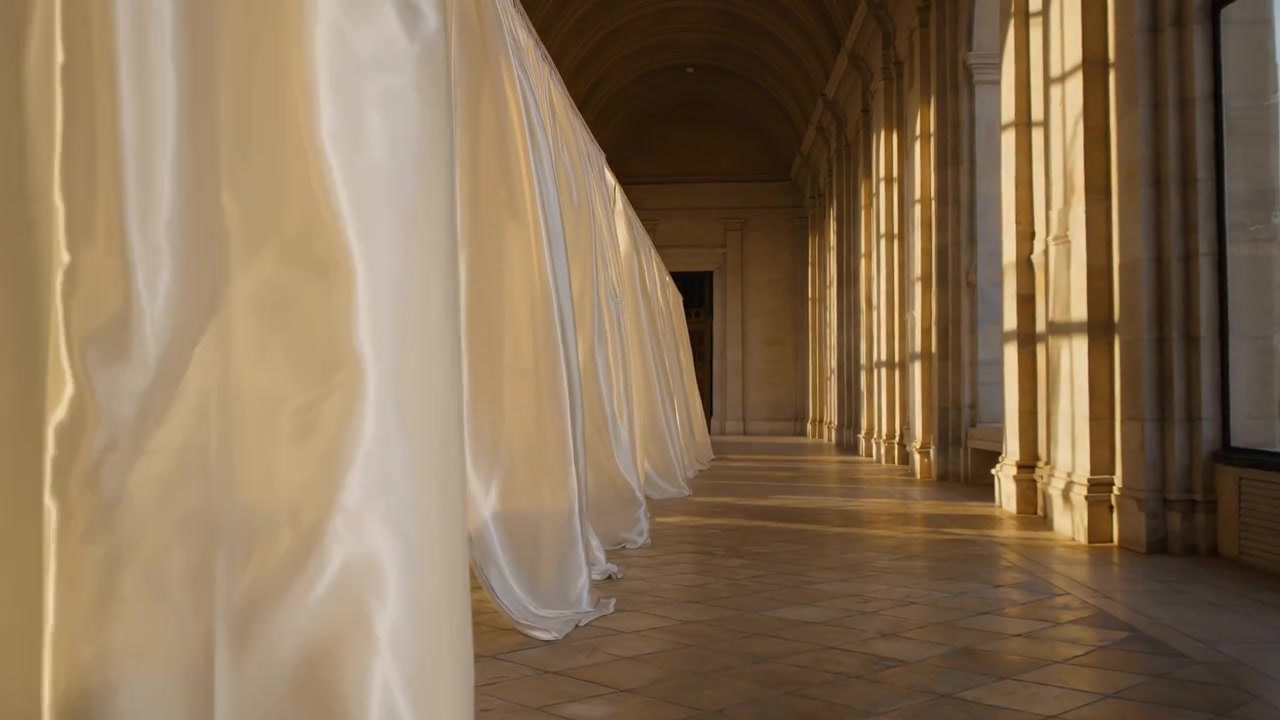}} & \raisebox{-.5\height}{\includegraphics[width=.138\textwidth,keepaspectratio]{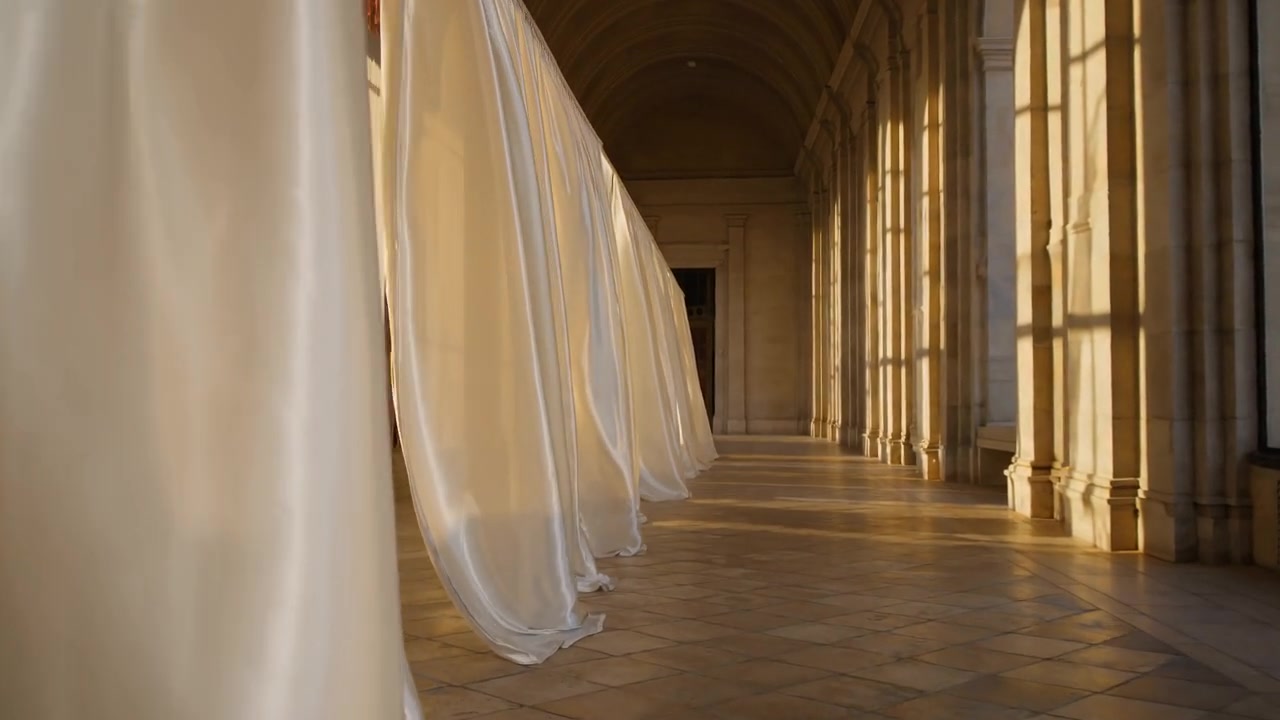}} \\
\end{tabular}%
}
\caption{Representative methods for text-to-video generation using their official configurations, with Wan2.1 T2V 14B as the shared backbone.}
\label{fig:sparse-t2v-qualitative}
\end{figure*}

%% file: tables/framework_examples_algorithm.tex
\begin{table*}[!t]
  \centering
  \caption{Representative methods described by the $4+1+3$ framework: algorithmic dimensions. Underlined text denotes a framework component, followed by its corresponding framework choices. A dash indicates no separate entry.}
  \label{tab:framework-examples-algorithm}
  \begingroup
  \newcolumntype{E}{>{\RaggedRight\arraybackslash}X}
  \newcommand{\examplecomponent}[1]{\underline{#1}}
  \newcommand{\examplenone}{--}
  \newcommand{\algorithmgroup}[1]{%
    \rowcolor{black!12}\multicolumn{6}{@{}l@{}}{\hspace{1mm}\bfseries #1}\tabularnewline
  }
  \setlength{\tabcolsep}{2.0pt}
  \renewcommand{\arraystretch}{1.12}
  \hyphenpenalty=10000
  \exhyphenpenalty=10000
  \emergencystretch=.6em
  \fontsize{6.15}{7.3}\selectfont
  \begin{tabularx}{\textwidth}{@{}>{\RaggedRight\arraybackslash}p{2.55cm}*{5}{E}@{}}
    \rowcolor{black}
    \color{white}\bfseries Method &
    \color{white}\bfseries Source &
    \color{white}\bfseries Evidence Construction &
    \color{white}\bfseries Granularity &
    \color{white}\bfseries Support Specification &
    \color{white}\bfseries Auxiliary Mechanisms \\

    \algorithmgroup{Static: prior- and template-guided support}
    \textbf{Sparse Transformer}~\cite{p190410509} & \examplecomponent{Structural Context}: priors & \examplenone & \examplecomponent{Token Group}: blocks & \examplecomponent{Structural Templates}: window, dilated & \examplenone \\
    \textbf{VideoGPT}~\cite{p210410157} & \examplecomponent{Structural Context}: priors & \examplenone & \examplecomponent{Token Group}: blocks & \examplecomponent{Structural Templates}: window, dilated & \examplenone \\

    \algorithmgroup{Static: offline-calibrated support}
    \textbf{DiTFastAttn}~\cite{p240608552} & \examplecomponent{Structural Context}: layer/head IDs; \examplecomponent{Calibration Traces}: fidelity measurements & \examplecomponent{Statistical Scoring}: error estimates & \examplecomponent{Token Group}: regions & \examplecomponent{Structural Templates}: window; \examplecomponent{Template Search}: error & \examplecomponent{Reuse}: attention outputs; \examplecomponent{Correction}: residual sharing \\
    \textbf{Sparse-vDiT}~\cite{p250603065} & \examplecomponent{Structural Context}: layer/head IDs; \examplecomponent{Calibration Traces}: fidelity measurements & \examplecomponent{Statistical Scoring}: error estimates & \examplecomponent{Token Group}: blocks & \examplecomponent{Structural Templates}: diagonal, stripe; \examplecomponent{Template Search}: error & \examplecomponent{Correction}: dense fallback \\
    \textbf{CalibAtt}~\cite{p260305503} & \examplecomponent{Structural Context}: layer/head IDs; \examplecomponent{Calibration Traces}: attention traces & \examplecomponent{Pooling}: sum aggregation; \examplecomponent{Statistical Scoring}: frequency estimates & \examplecomponent{Token Group}: blocks & \examplecomponent{Ranked Selection}: top-$p$; \examplecomponent{Threshold}: fixed $\tau$ & \examplecomponent{Reuse}: attention outputs \\

    \algorithmgroup{Dynamic: content-dependent support}
    \textbf{DraftAttention}~\cite{p250514708} & \examplecomponent{Current State}: $Q/K$ & \examplecomponent{Pooling}: mean aggregation; \examplecomponent{Statistical Scoring}: similarity scores & \examplecomponent{Token Group}: regions & \examplecomponent{Ranked Selection}: top-$k$ & \examplenone \\
    \textbf{FAST-AR}~\cite{p260201801} & \examplecomponent{Current State}: $Q$; \examplecomponent{History}: past features; \examplecomponent{Conditioning}: text & \examplecomponent{Projection}: random projection; \examplecomponent{Hashing}: LSH & \examplecomponent{Token Pair}: query--key connection; \examplecomponent{Token Group}: buckets & \examplecomponent{Ranked Selection}: top-$k$; \examplecomponent{Membership}: bucket & \examplenone\\
    \textbf{DSV}~\cite{p250207590} & \examplecomponent{Current State}: hidden states & \examplecomponent{Projection}: low-rank projection & \examplecomponent{Token Group}: regions & \examplecomponent{Ranked Selection}: top-$k$ & \examplenone \\
    \textbf{Sparse VideoGen2}~\cite{p250518875} & \examplecomponent{Current State}: $Q/K$; \examplecomponent{History}: past statistics & \examplecomponent{Clustering}: $k$-means; \examplecomponent{Statistical Scoring}: similarity scores & \examplecomponent{Token Group}: clusters & \examplecomponent{Ranked Selection}: top-$p$ & \examplecomponent{Reuse}: construction intermediates \\
    \textbf{SpargeAttention}~\cite{p250218137} & \examplecomponent{Current State}: $Q/K$, softmax state & \examplecomponent{Pooling}: mean aggregation; \examplecomponent{Statistical Scoring}: similarity scores, softmax bounds & \examplecomponent{Token Group}: blocks & \examplecomponent{Ranked Selection}: top-$p$; \examplecomponent{Threshold}: fixed $\tau$ & \examplenone \\
    \textbf{MoGA}~\cite{p251018692} & \examplecomponent{Current State}: hidden states & \examplecomponent{Scoring Network}: gating heads & \examplecomponent{Token Group}: clusters & \examplecomponent{Membership}: route & \examplenone \\
    \textbf{Sol-Attn}~\cite{p260724027} & \examplecomponent{Current State}: $Q/K/V$ & \examplecomponent{Pooling}: mean aggregation; \examplecomponent{Statistical Scoring}: similarity scores & \examplecomponent{Token Group}: blocks & \examplecomponent{Threshold}: adaptive $\tau$ & \examplecomponent{Correction}: Taylor correction \\
    \textbf{HyperVAttention}~\cite{p260703012} & \examplecomponent{Structural Context}: priors; \examplecomponent{Current State}: $Q/K$; \examplecomponent{History}: past statistics & \examplecomponent{Clustering}: $k$-means; \examplecomponent{Statistical Scoring}: similarity scores & \examplecomponent{Token Group}: clusters & \examplecomponent{Ranked Selection}: top-$p$ & \examplecomponent{Reuse}: construction intermediates \\

    \algorithmgroup{Dynamic methods with prominent auxiliary mechanisms}
    \textbf{PISA}~\cite{p260201077} & \examplecomponent{Current State}: $Q/K/V$ & \examplecomponent{Pooling}: mean/sum aggregation; \examplecomponent{Statistical Scoring}: covariance correction & \examplecomponent{Token Group}: blocks & \examplecomponent{Ranked Selection}: top-$k$ & \examplecomponent{Correction}: Taylor correction \\
    \textbf{LIVEditor}~\cite{p260504569} & \examplecomponent{Current State}: $Q/K/V$ & \examplecomponent{Pooling}: mean aggregation; \examplecomponent{Statistical Scoring}: similarity scores, error estimates & \examplecomponent{Token Group}: blocks & \examplecomponent{Ranked Selection}: top-$k$ & \examplecomponent{Correction}: Taylor correction, dense fallback \\
    \textbf{HEART}~\cite{p260514513} & \examplecomponent{Structural Context}: layer/head IDs; \examplecomponent{Calibration Traces}: offline samples; \examplecomponent{Current State}: $Q/K$; \examplecomponent{History}: past supports & \examplecomponent{Pooling}: mean aggregation; \examplecomponent{Statistical Scoring}: error estimates & \examplecomponent{Token Group}: blocks, clusters & \examplecomponent{Ranked Selection}: top-$p$ & \examplecomponent{Reuse}: support \\
    \bottomrule
  \end{tabularx}
  \endgroup
\end{table*}

%% file: tables/framework_examples_system.tex
\begin{table*}[!t]
  \centering
  \caption{Representative system methods in the $4+1+3$ framework. Underlined text denotes a component, followed by its corresponding framework choices. A dash indicates no separate entry.}
  \label{tab:framework-examples-system}
  \begingroup
  \newcolumntype{E}{>{\RaggedRight\arraybackslash}X}
  \newcommand{\examplecomponent}[1]{\underline{#1}}
  \newcommand{\examplenone}{--}
  \newcommand{\systemgroup}[1]{%
    \rowcolor{black!12}\multicolumn{4}{@{}l@{}}{\hspace{1mm}\bfseries #1}\tabularnewline
  }
  \setlength{\tabcolsep}{2.6pt}
  \renewcommand{\arraystretch}{1.12}
  \hyphenpenalty=10000
  \exhyphenpenalty=10000
  \emergencystretch=.6em
  \fontsize{6.5}{7.7}\selectfont
  \begin{tabularx}{\textwidth}{@{}>{\RaggedRight\arraybackslash}p{3.0cm}*{3}{E}@{}}
    \rowcolor{black}
    \color{white}\bfseries Method &
    \color{white}\bfseries Data Organization &
    \color{white}\bfseries Execution &
    \color{white}\bfseries Dataflow \\

    \systemgroup{Static: content-independent support}
    \textbf{Sparse Transformer}~\cite{p190410509} & \examplecomponent{Layout}: permutation & \examplecomponent{Computation}: kernel optimization & \examplenone \\
    \textbf{STA}~\cite{p250204507} & \examplecomponent{Connectivity Encoding}: implicit indexing; \examplecomponent{Layout}: permutation & \examplecomponent{Partitioning}: tiling; \examplecomponent{Computation}: kernel optimization & \examplecomponent{Pipelining}: asynchronous transfer, warp specialization \\

    \textbf{PAROAttention}~\cite{p250616054} & \examplecomponent{Connectivity Encoding}: bitmap encoding; \examplecomponent{Layout}: permutation & \examplecomponent{Partitioning}: tiling; \examplecomponent{Computation}: mixed-precision computation & \examplecomponent{Pipelining}: prefetching, double buffering \\
    \textbf{Sparse-vDiT}~\cite{p250603065} & \examplenone & \examplecomponent{Computation}: kernel optimization; \examplecomponent{Partitioning}: batching & \examplenone \\
    \textbf{CalibAtt}~\cite{p260305503} & \examplecomponent{Connectivity Encoding}: interval encoding & \examplecomponent{Computation}: kernel optimization; \examplecomponent{Partitioning}: tiling & \examplenone \\

    \systemgroup{Dynamic: content-dependent support}
    \textbf{DraftAttention}~\cite{p250514708} & \examplecomponent{Layout}: permutation & \examplecomponent{Partitioning}: tiling & \examplenone \\
    \textbf{FAST-AR}~\cite{p260201801} & \examplenone & \examplecomponent{Computation}: mixed-precision computation & \examplenone \\
    \textbf{DSV}~\cite{p250207590} & \examplenone & \examplecomponent{Computation}: kernel optimization; \examplecomponent{Partitioning}: query grouping; \examplecomponent{Scheduling}: cost-based dispatch & \examplecomponent{Access}: selective communication; \examplecomponent{Residency}: on-chip retention, offloading \\
    \textbf{FG-Attn}~\cite{p250916518} & \examplenone & \examplecomponent{Partitioning}: query grouping; \examplecomponent{Scheduling}: cost-based dispatch, persistent scheduling & \examplecomponent{Access}: cooperative gather; \examplecomponent{Pipelining}: asynchronous transfer \\
    \textbf{Sparse VideoGen2}~\cite{p250518875} & \examplecomponent{Layout}: permutation & \examplecomponent{Computation}: kernel optimization & \examplenone \\
    \textbf{SpargeAttention}~\cite{p250218137} & \examplecomponent{Layout}: permutation & \examplecomponent{Computation}: kernel optimization & \examplenone \\
    \textbf{MoGA}~\cite{p251018692} & \examplecomponent{Layout}: packing, permutation & \examplecomponent{Computation}: operator optimization & \examplenone \\
    \textbf{Sol-Attn}~\cite{p260724027} & \examplenone & \examplecomponent{Computation}: kernel optimization & \examplecomponent{Residency}: on-chip retention \\
    \textbf{HyperVAttention}~\cite{p260703012} & \examplecomponent{Layout}: permutation & \examplecomponent{Computation}: kernel optimization; \examplecomponent{Partitioning}: query grouping & \examplenone \\

    \systemgroup{Dynamic methods with prominent auxiliary mechanisms}
    \textbf{PISA}~\cite{p260201077} & \examplecomponent{Layout}: packing & \examplecomponent{Computation}: kernel optimization & \examplecomponent{Access}: coalesced access \\
    \textbf{LIVEditor}~\cite{p260504569} & \examplecomponent{Layout}: packing, permutation & \examplecomponent{Computation}: kernel optimization; \examplecomponent{Partitioning}: query grouping & \examplenone \\
    \textbf{PASA}~\cite{p260412219} & \examplenone & \examplenone & \examplecomponent{Pipelining}: prefetching, warp specialization \\
    \bottomrule
  \end{tabularx}
  \endgroup
\end{table*}

%% file: sections/05_evaluation.tex
\section{Evaluation Benchmarks and Metrics}
\label{sec:evaluation}
Evaluating sparse attention for visual generation requires paired evidence: the method should preserve task-level
  generation quality, and the interactions removed by sparsification should translate into measurable acceleration.

\subsection{Generation Quality}
Table~\ref{tab:generation-evaluation} organizes metrics and benchmarks by their intended visual-generation task. Reference-based metrics such as PSNR, SSIM, and LPIPS are useful when a dense output or paired target is available; distributional and semantic metrics instead assess sample realism, diversity, or condition alignment. Benchmark suites extend these measurements with task-specific diagnostics for image and video generation, long-horizon consistency, editing, world models, 3D/4D content, and audio-driven synthesis. 

\input{tables/visual_generation_evaluation.tex}

Figures~\ref{fig:sparse-t2i-qualitative} and~\ref{fig:sparse-t2v-qualitative} show qualitative comparisons for image and video generation.

\subsection{Acceleration Efficiency}
The amount of attention removed and the acceleration ultimately observed are related but not interchangeable.
  Table~\ref{tab:acceleration-metrics} includes both analytical and empirical measures: sparsity and FLOPs indicate how
  much attention computation is removed in principle, whereas attention latency records the execution time of the
  resulting sparse operator on a specific hardware platform. End-to-end latency additionally includes support
  construction, index or layout transformations, and the portions of the generator that remain dense. Throughput,
  memory, communication, and scaling efficiency further describe deployment behavior, while first-frame and per-frame
  latency are especially relevant to autoregressive and streaming generation.

Capacity savings should be distinguished from latency reductions. DSV offloads large sparse-index tensors between forward and backward use to relieve GPU memory pressure~\cite{p250207590}; the resulting capacity benefit does not establish a speedup, and transfer costs must be included in timing. By contrast, eliminating intermediate writes or padded arithmetic directly reduces the work required to evaluate sparse attention.

\input{tables/sparse_attention_acceleration_metrics.tex}

%% file: tables/visual_generation_evaluation.tex

\begin{table}[!t]
\caption{Common Metrics and Benchmarks for Visual Generation}
\label{tab:generation-evaluation}
\centering
\tiny
\setlength{\tabcolsep}{2.3pt}
\renewcommand{\arraystretch}{1.02}
\begin{tabularx}{\columnwidth}{>{\raggedright\arraybackslash}p{2.0cm}>{\raggedright\arraybackslash}p{2.30cm}X}
\toprule
\multicolumn{3}{l}{\textbf{Evaluation metrics}} \\
Metric & Task & Target \\
\midrule
FID \cite{heusel2017ttur} & Image and frame-level video & Distance between real and generated feature distributions. \\
IS \cite{salimans2016improved} & Image and class-conditioned video & Classifier confidence and semantic diversity across generated samples. \\
PSNR & Image, video, editing, world model, 3D and 4D & Pixel-level fidelity to a paired reference or dense output. \\
SSIM \cite{wang2004ssim} & Image, video, editing, world model, 3D and 4D & Local luminance, contrast, and structural similarity. \\
LPIPS \cite{zhang2018lpips} & Image, video, editing, 3D and 4D & Learned perceptual distance between paired visual outputs. \\
CLIPScore \cite{hessel2021clipscore} & Text-to-image, text-to-video, and editing & Semantic alignment between generated content and the text condition. \\
FVD \cite{unterthiner2018fvd} & Video, long video, and audio-driven video & Spatiotemporal distribution fidelity over a generated video set. \\
Chamfer distance \cite{p260405182} & 3D and 4D & Geometric discrepancy between generated and reference point sets. \\
Sync-C \cite{p251002617} & Audio-driven video and co-speech & Confidence in speech-to-lip synchronization. \\
Sync-D \cite{p251002617} & Audio-driven video and co-speech & Distance between speech and lip-motion embeddings. \\
\midrule
\multicolumn{3}{l}{\textbf{Evaluation benchmarks}} \\
Benchmark & Task & Target \\
\midrule
MS-COCO \cite{lin2014coco} & Text-to-image & Distributional image quality and caption--image alignment. \\
GenEval \cite{ghosh2023geneval} & Text-to-image & Object presence, attributes, counting, position, and color binding. \\
T2I-CompBench \cite{huang2023t2icompbench} & Text-to-image & Compositional alignment involving attributes, relations, and complex prompts. \\
VBench \cite{huang2024vbench} & Text-to-video & Multidimensional visual, temporal, motion, and semantic video quality. \\
VBench++ \cite{huang2024vbenchpp} & Text-to-video and image-to-video & Multidimensional video quality across generation conditions. \\
EvalCrafter \cite{liu2024evalcrafter} & Text-to-video & Visual quality, content fidelity, motion quality, and text alignment. \\
T2V-CompBench \cite{sun2024t2vcompbench} & Text-to-video & Attributes, actions, interactions, spatial relations, and numeracy. \\
VBench-Long \cite{p260204789} & Long and autoregressive video & Long-horizon subject, motion, and temporal consistency. \\
ViewBench \cite{p260207854} & Video world model & View revisit fidelity, loop closure, and accumulated geometric drift. \\
EditVerse-\allowbreak Bench \cite{p260504569} & Video editing & Edit execution, source preservation, visual quality, and temporal coherence. \\
FiVE-Bench \cite{p260504569} & video editing & Object-level editing precision and preservation of unedited content. \\
GSO \cite{p260405182} & Object-centric 3D reconstruction & Novel-view appearance and reconstruction quality for individual objects. \\
TalkShow \cite{p251002617} & Audio-driven video & Portrait fidelity, body/hand motion, and speech synchronization. \\
\bottomrule
\end{tabularx}
\end{table}

%% file: tables/sparse_attention_acceleration_metrics.tex

\begin{table}[!t]
\caption{Metrics for Sparse-Attention Acceleration}
\label{tab:acceleration-metrics}
\centering
\tiny
\setlength{\tabcolsep}{3pt}
\renewcommand{\arraystretch}{1.03}
\begin{tabularx}{\columnwidth}{>{\raggedright\arraybackslash}p{2.75cm}X}
\toprule
Metric & Target \\
\midrule
Attention sparsity & Fraction of attention edges or blocks skipped. \\
FLOPs & Arithmetic cost of attention or the complete model. \\
Attention latency & Execution time of the sparse attention operator. \\
Attention speedup & Dense-to-sparse attention latency ratio under the same tensor shape. \\
End-to-end latency & Complete generation time including routing and non-attention modules. \\
End-to-end speedup & Dense-to-sparse complete-generation latency ratio under matched settings. \\
Throughput & Sustained samples, videos, frames, or tokens completed per second. \\
First-frame latency & Startup delay in autoregressive or streaming generation. \\
Per-frame latency & Steady-state frame-generation time after startup. \\
Peak GPU memory & Maximum device memory usage and feasible sequence, frame, or view scale. \\
Communication volume & Inter-device data transfer during distributed execution. \\
Scaling efficiency & Throughput or latency retention as the device count increases. \\
Selection overhead & Runtime cost of constructing and indexing sparse support. \\
Layout overhead & Runtime cost of tensor and index rearrangement for sparse execution. \\
\bottomrule
\end{tabularx}
\end{table}

%% file: sections/06_challenges.tex
\section{Challenges and Future Directions}
\label{sec:challenges}

\subsection{Reliable Support Under Nonstationary Generation}
Dynamic support faces a fundamental information constraint: the exact scores needed to identify important edges are the dense quantity being avoided. Selectors instead rely on proxies such as pooled Q/K statistics, sampled rows, cluster centroids, external conditions, or previously constructed masks, whose reliability varies across heads, layers, generation stages, and visual states. Fixed Top-$k$ budgets may misallocate computation, while reused masks become stale as the content changes \cite{p260812780,prapidcvpr2026,p260204789}. A promising direction combines complementary support sources: local, causal, or geometric edges provide a stable backbone, and an uncertainty-aware dynamic budget retrieves residual long-range interactions. Low-confidence queries can then receive more capacity, support drift can trigger refresh, and exceptional states can selectively fall back to dense attention. This design also balances the portability of training-free selectors against the co-adaptation and transfer requirements of trainable routers.

\subsection{From Local Approximation to Trajectory Stability}
Sparsification changes the hidden states and Q/K representations used by later attention calls, creating a feedback loop between representation error and subsequent support selection. Diffusion and flow models repeat this loop across layers and sampling steps, while autoregressive generators also insert the affected frame or chunk into future context. Light Forcing shows that an early-chunk error may remain visible even after later chunks return to dense attention \cite{p260204789}. Retained attention mass or small one-step error therefore cannot guarantee long-horizon identity, motion, geometry, or cross-modal consistency. Rectified SpaAttn, PISA, and SVG-EAR accordingly approximate or reallocate omitted contributions rather than simply setting them to zero \cite{p251119835,p260201077,p260308982}. Future methods should propagate local error estimates through the generation trajectory and use the remaining error budget to trigger refresh, densification, or residual compensation. Evaluation should likewise cover different rollout lengths and abrupt changes in motion, scene, or conditioning.

\subsection{Executable and Portable Dynamic Sparsity}

Removing attention edges is useful only when the resulting support can be executed efficiently. GPUs favor regular tiles and balanced workloads, whereas fine-grained dynamic masks introduce index construction, sorting, gather--scatter traffic, workload divergence, and many short kernel launches. The density that minimizes FLOPs may therefore fail to minimize latency, and the crossover between dense and sparse execution varies with tensor shape, block size, hardware generation, and the dense baseline. A useful next abstraction is an executable support representation shared by the selector, compiler, runtime, and distributed partitioner. Device-aware cost models could then choose a materialization layout, reorder blocks, balance work across devices, and switch among dense, static-sparse, and dynamic-sparse implementations at runtime. 

\subsection{Compositional Acceleration}

Sparse attention will increasingly operate alongside the techniques in Figure~\ref{fig:technique-comparison}, but their speedups do not multiply automatically and their errors can interact. Quantization may alter logits near a selection boundary, cache reuse can leave both activations and masks stale, and pruning or distillation may change the attention distribution on which a selector relies. QuantSparse, FPSAttention, and IndexCache respectively expose attention shift, joint sparse--quantized training, and the opportunity to amortize support construction through reuse \cite{p250923681,p250604648,p260312201}. Rather than stacking independently tuned modules, future designs should allocate shared error, latency, and memory budgets across numerical precision, support density, refresh frequency, and residual compensation. Matched factorial ablations are then needed to distinguish sparse-only, companion-only, and combined gains.

%% file: sections/07_conclusion.tex
\section{Conclusion}
This survey presents a 4 + 1 + 3 framework that links the algorithmic design of sparse attention for visual generation with its system implementation. Using this framework, we review how static and dynamic methods construct sparse support and how reuse and correction address repeated computation and approximation error. We relate these choices to the way sparse attention is implemented, tracing the progression toward joint design of sparse support and its execution. The survey also summarizes existing evaluation practices and outlines open research challenges.